\documentclass[10pt,a4paper]{article}

\usepackage[margin=1.5in]{geometry}
\usepackage{subcaption}

\usepackage{pratik}
\usepackage[color=gray!30,textsize=tiny]{todonotes}

\renewcommand{\etal}{\textit{et al.}}
\newcommand{\micron}{{${\mu}m$}}

\newcommand{\dpp}[2]{\frac{\partial #1}{\partial #2}}

\newcommand{\braintablefig}[2]{%
\begin{minipage}[t]{0.49\linewidth}
  \centering\vspace{0pt}
  \includegraphics[width=\linewidth]{#1}
  \end{minipage}
  \begin{minipage}[t]{0.49\linewidth}
  \centering\vspace{0pt}
  \includegraphics[width=\linewidth]{#2}
  \end{minipage}
}

\usepackage{xcolor}

\usepackage{algorithm}
\usepackage{algpseudocode}

\begin{document}

\title{FireANTs: Adaptive Riemannian Optimization for Multi-Scale Diffeomorphic Matching}

\author[a,d]{Rohit Jena}
\author[a,b,*]{Pratik Chaudhari}
\author[a,c,d,*]{James C. Gee}
\affil[a]{\normalsize Computer and Information Science, University of Pennsylvania}
\affil[b]{\normalsize Electrical and Systems Engineering, University of Pennsylvania}
\affil[c]{\normalsize Radiology, Perelman School of Medicine, University of Pennsylvania}
\affil[d]{\normalsize Penn Image Computing and Science Laboratory, University of Pennsylvania}
\affil[*]{\normalsize Corresponding Authors: \href{pratikac@upenn.edu}{pratikac@upenn.edu}, \href{gee@upenn.edu}{gee@upenn.edu}}
\clearpage

\maketitle

\begin{abstract}
The paper proposes FireANTs, a multi-scale Adaptive Riemannian Optimization algorithm for dense diffeomorphic image matching.
Existing state-of-the-art methods for diffeomorphic image matching are slow due to inefficient implementations and slow convergence due to the ill-conditioned nature of the optimization problem.
Deep learning methods offer fast inference but require extensive training time, substantial inference memory, and fail to generalize across long-tailed distributions or diverse image modalities, necessitating costly retraining.
We address these challenges by proposing a training-free, GPU-accelerated multi-scale Adaptive Riemannian Optimization algorithm for fast and accurate dense diffeomorphic image matching.
FireANTs runs about 2.5$\times$ faster than ANTs on a CPU, and upto 1200$\times$ faster on a GPU.
On the GPU, FireANTs performs competitively with deep learning methods on inference runtime while consuming upto 10$\times$ less memory.
FireANTs shows remarkable robustness to a wide variety of matching problems across modalities, species, and organs without any domain-specific training or tuning.
Our framework allows hyperparameter grid search studies with significantly less resources and time compared to traditional and deep learning registration algorithms alike.

\end{abstract}

\section{Introduction}
\label{sec:intro}

The ability to identify and map corresponding elements across diverse datasets or perceptual inputs -- known as \textit{correspondence matching} -- is fundamental to interpreting and interacting with the world.
Correspondence matching between images is one of the longstanding fundamental problem in computer vision.
Influential computer vision researcher Takeo Kanade famously once said that the three fundamental problems of computer vision are: ``Correspondence, correspondence, correspondence''~\cite{Wang-thesis}.
Indeed, correspondence matching is fundamental and ubiquitous across various disciplines, manifesting in many forms including but not limited to stereo matching ~\cite{hamid2022stereo}, structure from motion ~\cite{carrivick2016structure,smith2016structure}, template matching ~\cite{brunelli2009template}, motion tracking ~\cite{zhou2008human,leon2021review}, shape correspondence ~\cite{van2011survey}, semantic correspondence ~\cite{lee2019sfnet}, point cloud matching ~\cite{pomerleau2015review}, optical flow ~\cite{fortun2015optical}, and deformable image matching ~\cite{sotiras2013deformable}.
Solving these problems addresses the desiderata for a wide range of applications in computer vision, robotics, medical imaging, remote sensing, photogrammetry, geological and ecological sciences, cognitive sciences, human-computer interaction, self-driving among many other fields.

Correspondence matching is broadly divided into two categories: sparse and dense matching.
Most sparse matching problems like stereo matching, structure from motion, and template matching involve finding a \textit{sparse set} of \textit{salient features} across images followed by matching them.
In such cases, the transformation between images, surfaces, or point clouds is typically also parameterized with a small number of parameters, e.g., an affine transform, homography or a fundamental matrix.
These methods are often robust to noise, occlusions, and salient features can be detected and matched efficiently via analytical closed forms.
In contrast, dense matching is much harder because the entire image is considered for matching and cannot be reduced to a sparse set of salient features, and the transformation between images is typically parameterized with a large number of parameters, e.g., a dense deformation field.
Moreover, dense matching is sensitive to local noise, and cannot be solved efficiently via analytical closed forms -- necessitating iterative optimization methods~\cite{ants,avants2008symmetric,avants2004geodesic,beg2005computing,ashburner2007fast,vercauteren2007diffeomorphic}.
Due to the dense and high-dimensional nature, these methods are often plagued with ill-posedness~\cite{mang2017semi,sotiras2013deformable,clarenz2006computational}, difficulty in optimization, inefficient implementations, and lack of scalability to high-resolution data.

In this work, we focus on dense deformable correspondence matching, which is the non-linear and local (hence deformable) alignment of two or more images into a common coordinate system.
Dense deformable correspondence matching is a fundamental problem in computer vision~\cite{bai2022deep}, medical imaging~\cite{zhu2019applications,bai2020population,murphy2011evaluation}, microscopy~\cite{yoo2017ssemnet,hand2009automated}, and remote sensing. Here, we focus on applications in biomedical and biological imaging.
In the biomedical and biological sciences, deformable correspondence matching is also referred to as deformable registration.
Within dense deformations, \textit{diffeomorphisms} are of special interest as a family of deformations that are invertible transformations such that both the transform and its inverse are differentiable.
This allows us to accurately model the correspondence between images while ensuring that the topological structure of the anatomy is preserved, i.e. no tearing or folding of the anatomy is introduced.

We address and tackle two fundamental problems in dense correspondence matching: ill-conditioning and scalability.
The ill-conditioning arises due to the high-dimensional and heterogeneous nature of the dense matching optimization objective, that can be mitigated by adaptive optimization methods.
Although standard adaptive optimization methods~\cite{rmsprop,kingma2014adam} are shown to work in fixed Euclidean spaces, it is not obvious how to extend this formulation to the non-Euclidean space of diffeomorphisms.
Fortunately, diffeomorphisms admit many interesting mathematical properties like being embedded in a Riemannanian manifold, having a Lie Group structure, and local geodesic formulations that can be exploited for adaptive optimization.
We present a \textit{novel} and \textit{mathematically rigorous} framework for adaptive optimization of diffeomorphic matching~\cref{sec:diffgroup}.
This is done by exploiting the group structure of diffeomorphisms to define a custom gradient descent algorithm, followed by adaptive optimization on this space.
Second, we observe that most existing state-of-the-art methods are prohibitively slow for high-resolution data, which limits their applicability to rigorous hyperparameter studies, large-scale data, or high-resolution alignment at mesoscopic or microscopic resolutions.

Our novel operational contributions lead to an algorithm that is around $2-7\times$ faster than state-of-the-art optimization toolkits on CPU, and upto three orders of magnitude faster on GPU.
Compared to deep learning methods, our framework demonstrates better generalization to unseen novel modalities and species while being competitive in terms of runtime and utilizing upto $10\times$ less memory.
This quantum leap in speedup and scalability also allows us to perform hyperparameter grid search studies with significantly less resources and time (\cref{fig:hyperparam,fig:runtime}) compared to traditional and deep learning registration algorithms, and tackle novel application areas like sub-micron expansion microscopy (\cref{fig:rnrexm}), and high-resolution atlas generation in under 25 minutes (\cref{fig:atlasfig}).
We package our contributions into a software toolkit called \textbf{FireANTs}, which is an open-source state-of-the-art toolkit for dense deformable correspondence matching.
Our framework can allow a practitioner to perform interactive dense matching that is useful for aligning complex, multimodal, multi-channel or high-resolution images~\cite{wang2020allen,xia2019spatial} or to provide guidance to compensate for missing data (e.g., in microscopy imaging), wherein a typical dense image matching algorithm would fail to run interactively.

\section{Results}
\label{sec:results}
FireANTs represents the next generation of frameworks superseding the widely established and successful adoption of the ANTs ecosystem spanning the gamut of biomedical and life sciences research.
We evaluate our method on fourteen datasets spanning more than 15000 image pairs, three organs (brain, lung, abdomen), seven modalities (T1w MRI, T2$^*$w MRI, CT, expansion microscopy, LSFM, fMOST, 9.4T MRI), six species (human, ovine, zebrafish, mouse, rat, non-human primates), and show remarkable generalization across all datasets without any domain-specific enhancements.
Since our method is a methodological improvement over ANTs, we evaluate FireANTs against established benchmarks where ANTs is one of the top performing methods, among other winning methods for the respective challenges.

FireANTs demonstrates remarkable runtime efficiency compared to ANTs on both CPU and GPU, while also outperforming most deep learning methods at inference runtime and consuming up to a tenth of the GPU memory, setting a new standard for runtime and memory efficiency.
This unprecedented efficiency allows a multitude of novel capabilities including registering 50$\times$ larger volumes in minutes on a single GPU, faster amortized runtimes over large batches enabling scalable registration of large datasets, efficient hyperparameter grid search studies, and high-resolution atlas building in under 25 minutes.

\subsection{Experiment Setup}
\label{sec:expsetup}
We briefly describe the significance, existing state-of-the-art and challenges associated with the chosen benchmarks to demonstrate the efficacy of FireANTs.
More details about the datasets and evaluation metrics are outlined in ~\cref{app:datasets}.

\paragraph{In-vivo brain mapping challenges ~\cite{klein2009evaluation,oasisdataset}} %
Klein \etal~\cite{klein2009evaluation} in their landmark paper reported an extensive evaluation of fourteen state-of-the-art registration algorithms on four neuroimaging datasets. 
The four neuroimaging datasets (IBSR18, CUMC12, MGH10, LPBA40) comprise different whole-brain labelling protocols, eight different evaluation measures and three independent analysis methods of over 2000 brain volume pairs. 
The Learn2Reg~\cite{hering2022learn2reg} version of the OASIS dataset~\cite{oasisdataset} is another large scale dataset with 414 subjects for inter-subject brain MRI registration, routinely used as a training dataset for deep learning algorithms~\cite{balakrishnan2019voxelmorph,lku,tian2024unigradicon}.
Evaluating on these challenges is therefore imperative to establish FireANTs as an effective, versatile and robust algorithm for neuroimaging applications.
In total, we compare with state-of-the-art baselines on over \textit{2500 brain volume pairs}, with varying number of labeled anatomical regions and resolutions.

\begin{figure}[p]%
    \centering
    \begin{minipage}{0.95\linewidth}
        \includegraphics[width=\linewidth]{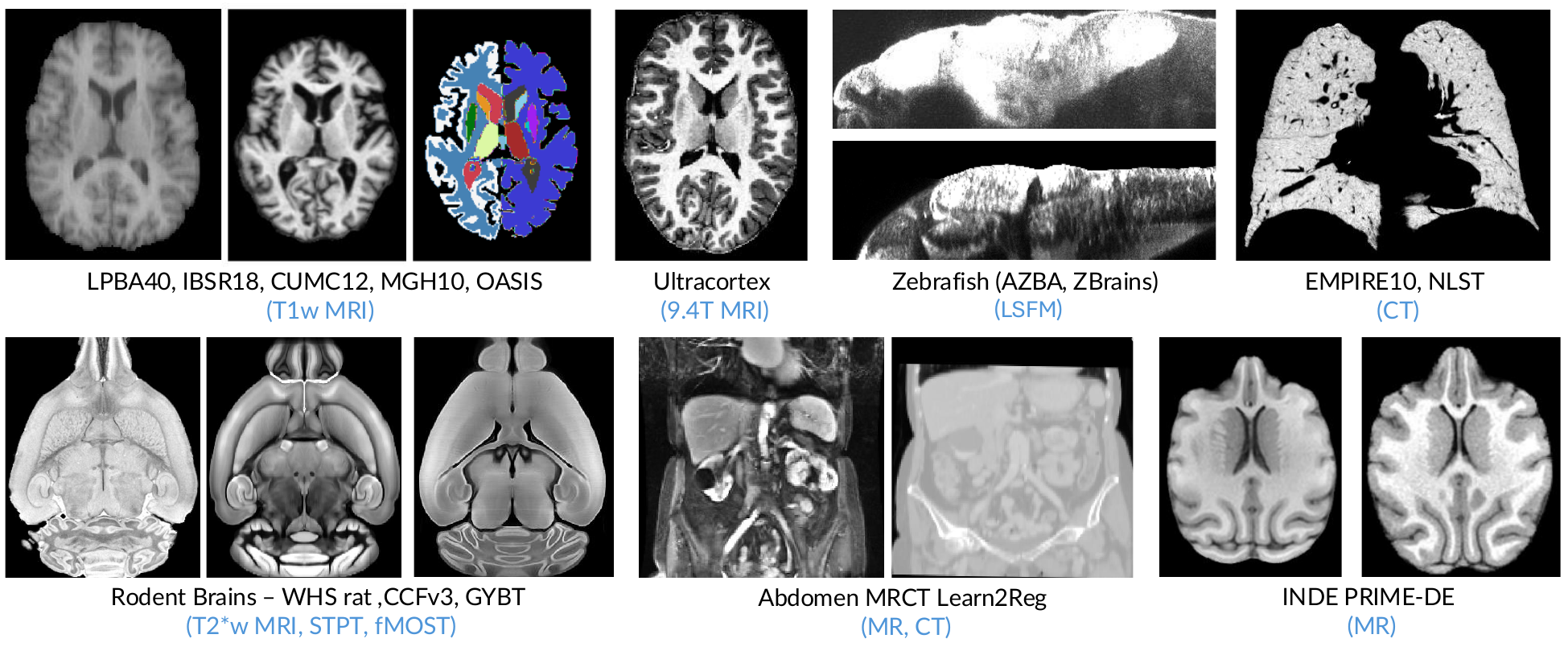}
        \subcaption{\footnotesize Overview of the datasets used in the paper}
    \end{minipage}
    \begin{minipage}{0.37\linewidth}
        \centering
        \includegraphics[width=\linewidth]{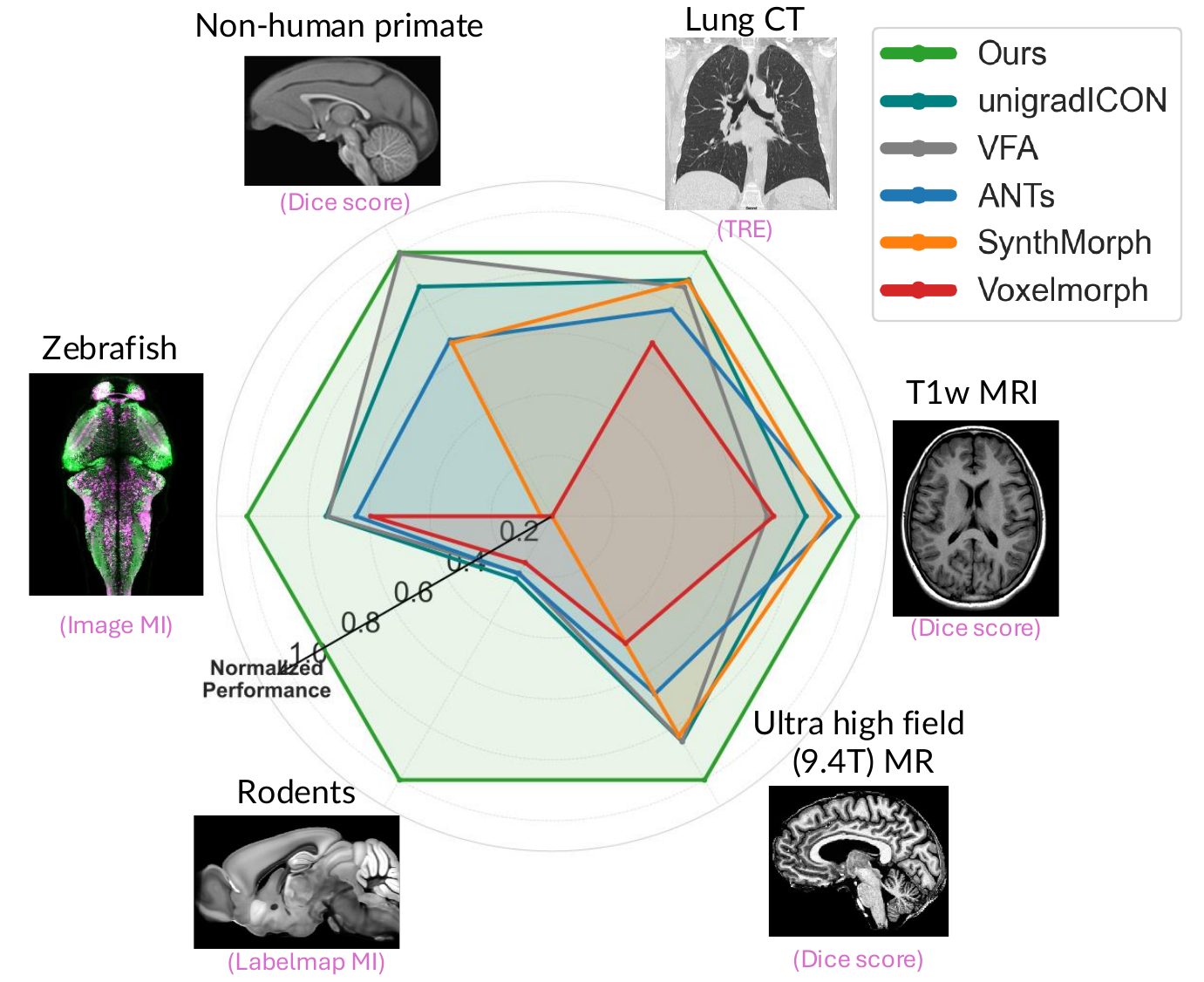}
        \label{fig:radarchart}
        \subcaption{\footnotesize Normalized performance of state-of-the-art registration algorithms across a wide range of datasets and benchmarks. FireANTs achieves asymptotically best normalized performance across various datasets and evaluation criteria.}
    \end{minipage}\hfill
    \begin{minipage}{0.6\linewidth}
        \centering
        \includegraphics[width=0.32\linewidth]{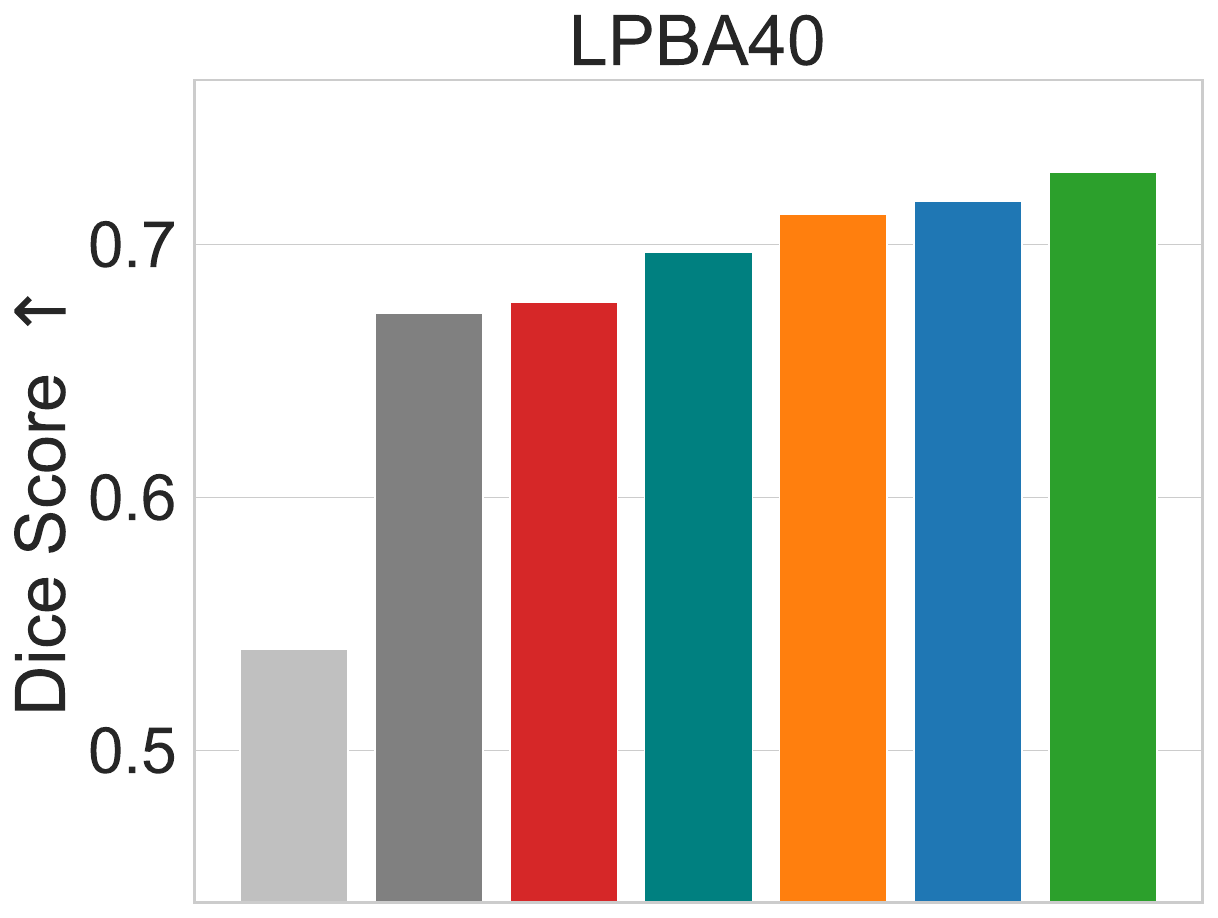}
        \includegraphics[width=0.32\linewidth]{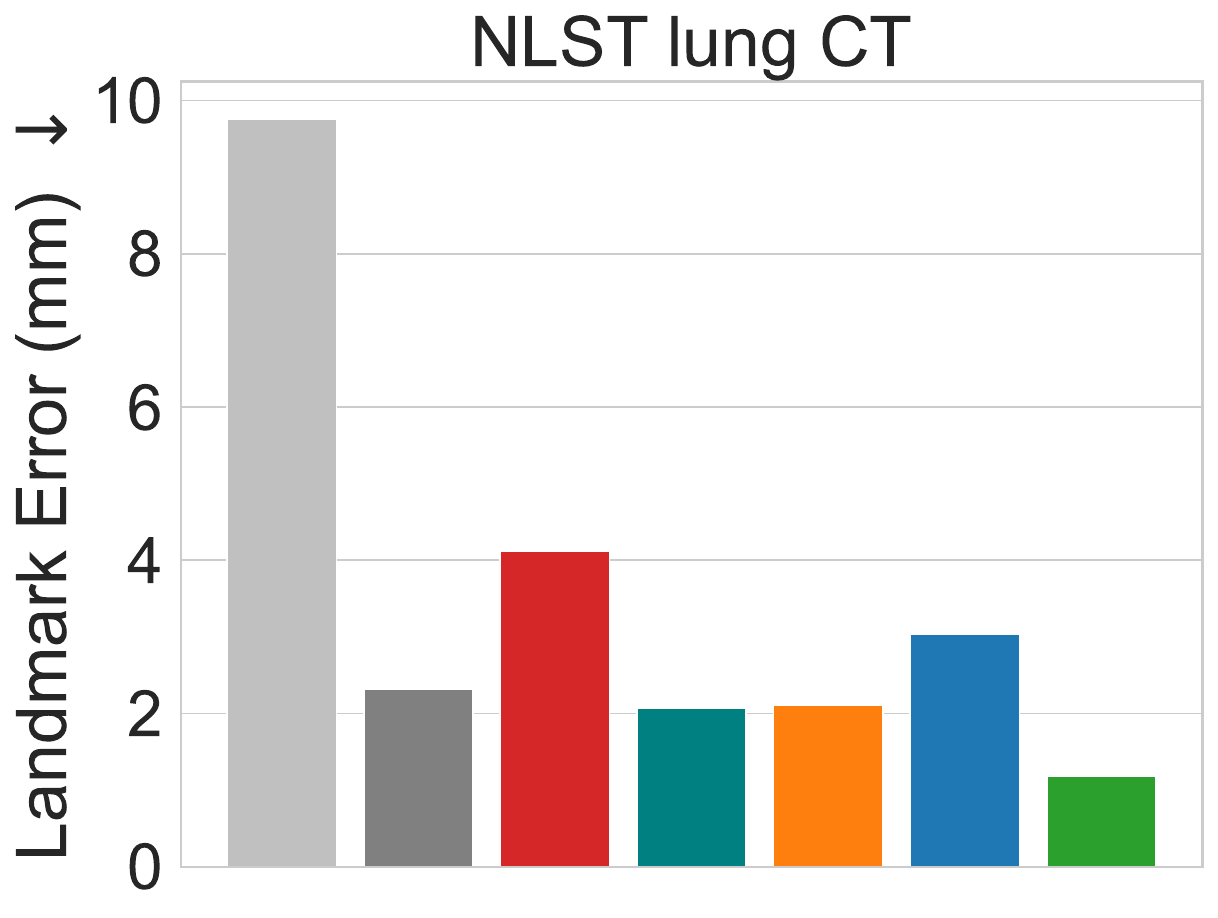}
        \includegraphics[width=0.32\linewidth]{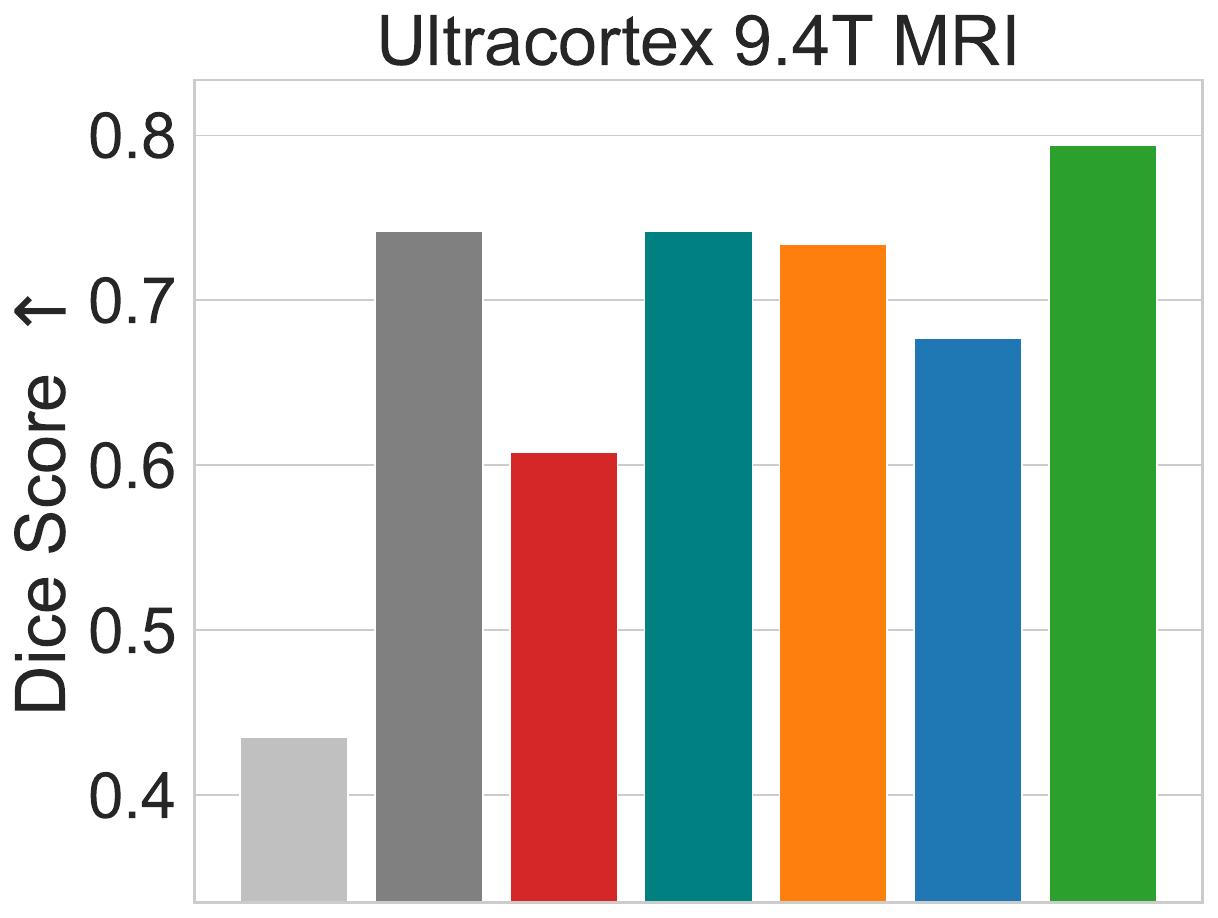}\\[0.5ex]
        \includegraphics[width=0.32\linewidth]{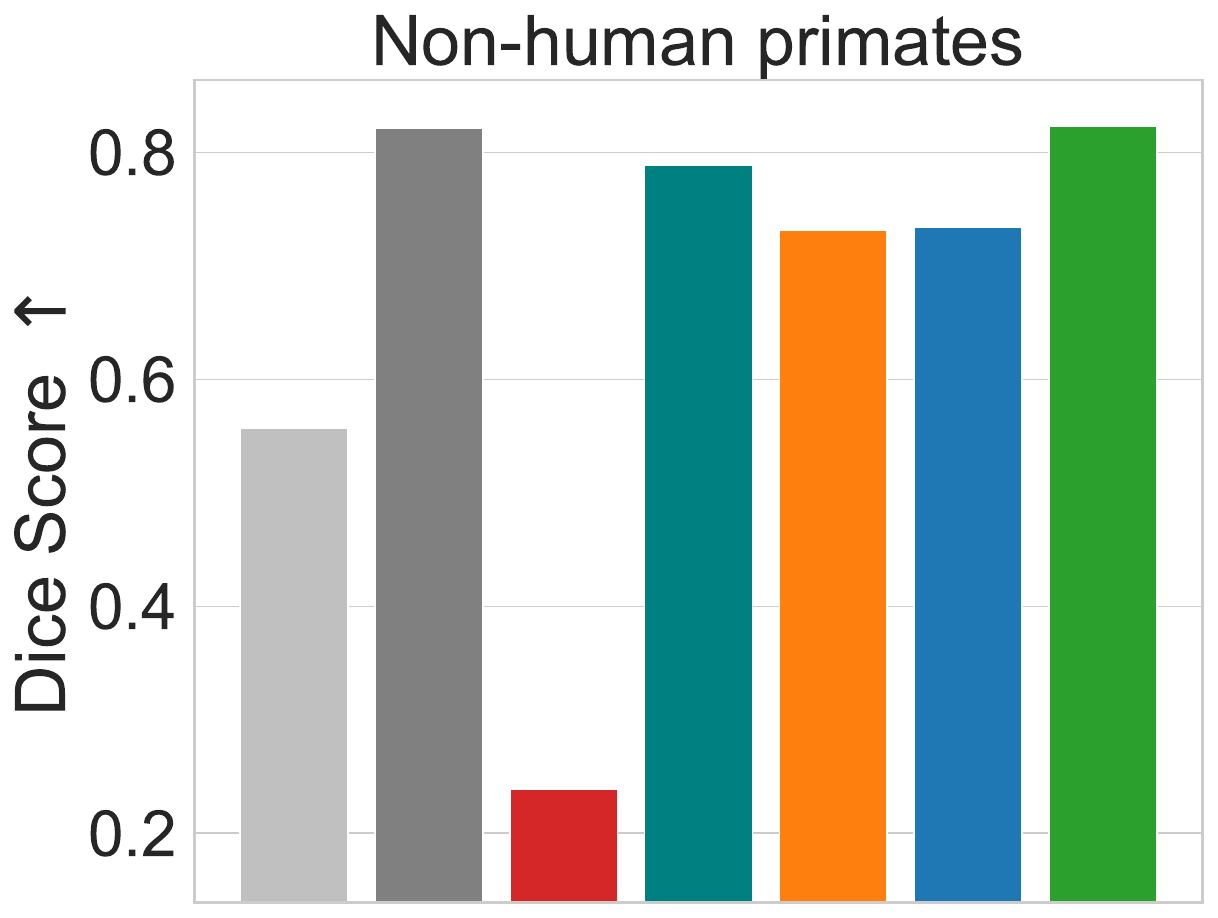}
        \includegraphics[width=0.32\linewidth]{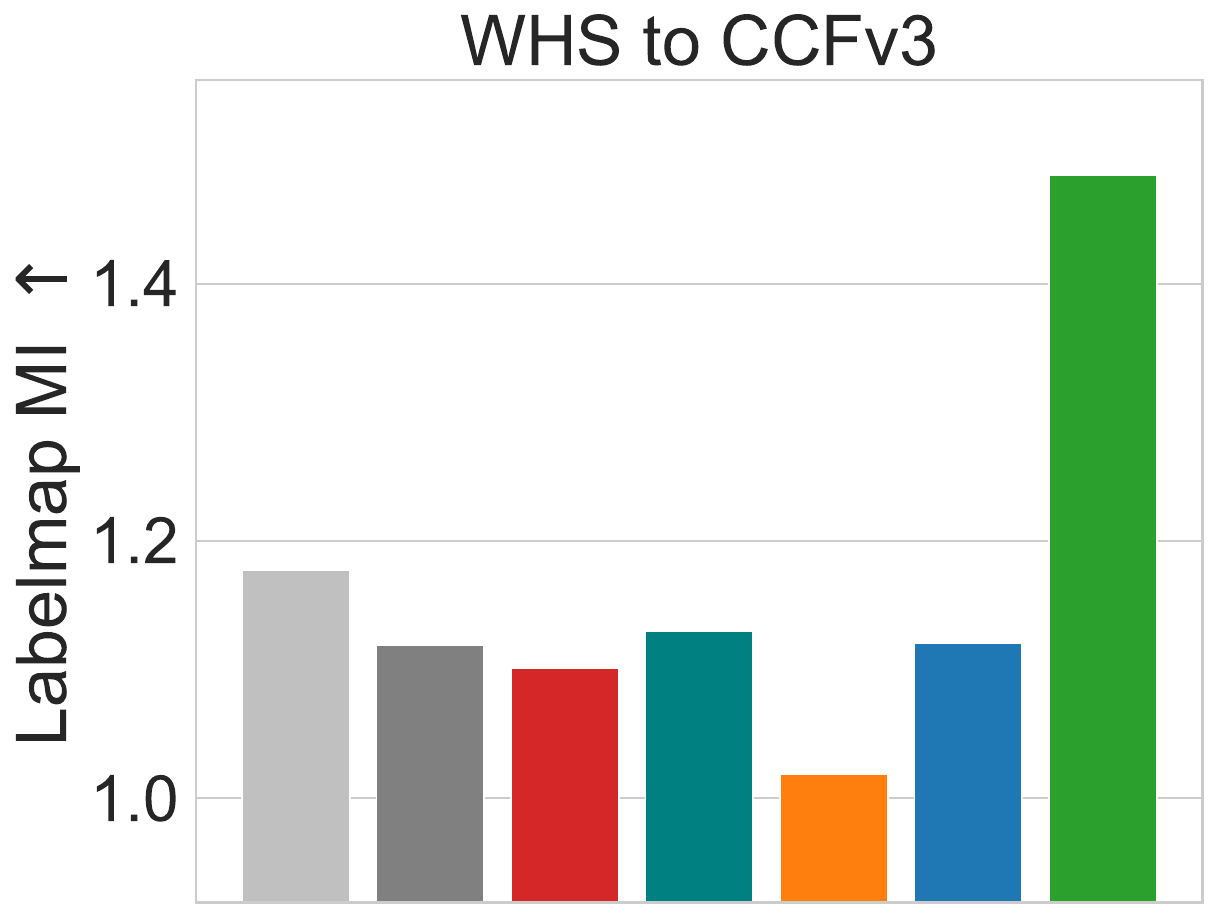}
        \includegraphics[width=0.32\linewidth]{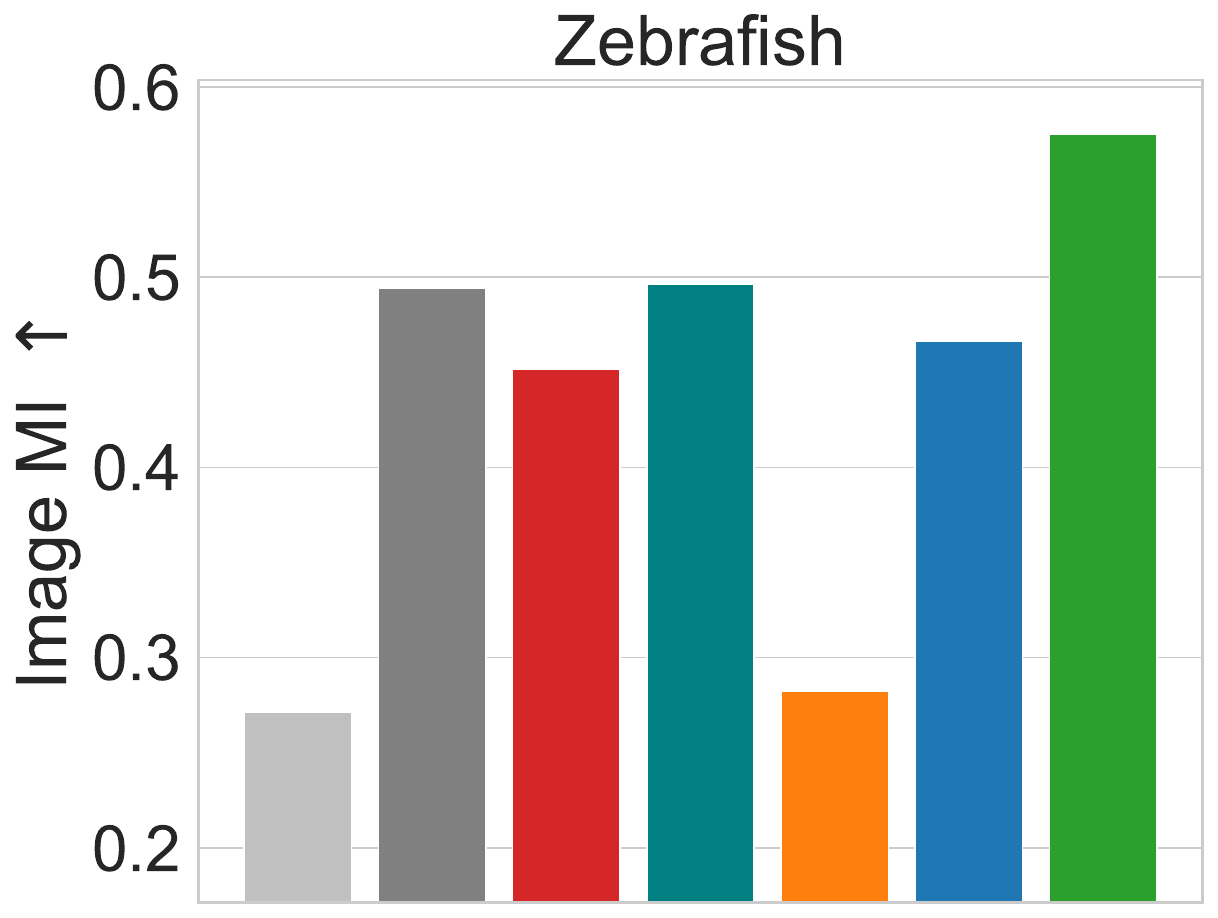}
        \subcaption{\footnotesize Raw performance (measured by Dice Score, Mutual Information of Labelmap and Intensity, Landmark Distance) of state-of-the-art registration algorithms across various datasets.
        In all datasets except NLST, higher scores are better.
        Colorbars are shown in \cref{fig:radarchart} with lightgray denoting zero displacement (baseline).
        }
    \end{minipage}

    \caption{
        \textbf{FireANTs can generalize to a large variety of modalities and datasets}:
        Registration quality is validated by measuring either the labelmap overlap, Mutual Information between aligned labelmap for different labelmaps across datasets, or anatomical landmark distance between the fixed and warped coordinate frames.
        We consider two community standard challenges where ANTs was the winner, two analogous contemporary challenges to enable broader comparison with deep learning methods, and five other scenarios spanning a broad set of challenges.
        Across six datasets spanning a spectrum of anatomical systems, species, and modalities, FireANTs achieves the best performance across all evaluation criteria, showcasing its generalization capabilities.
        }
    \label{fig:intro}
\end{figure}

\paragraph{PRIMatE Data Exchange (PRIME-DE) \cite{primate}}
The overarching goal of PRIMatE Data Exchange (PRIME-DE) is to create an open science resource for the neuroimaging community to facilitate the mapping of the non-human primate connectome.
The dataset features a familiar modality and anatomy (T1w MRI brain) but different structural organization (non-human primate).
This presents a challenge to compare the generalization capabilities of domain-agnostic or foundational registration algorithms with FireANTs.

\paragraph{Ultracortex \cite{ultracortex}}
The Ultracortex dataset hosts a unique collection of ultra-high field (9.4 Tesla) MRI data of the human brain.
This challenge provides a complementary problem to PRIME-DE - familiar anatomy and structural organization (human brain) but different modality (9.4T MRI) and resolution (sub-millimeters).

\paragraph{Waxholm Rat Brain and Allen CCFv3 mouse brain datasets \cite{kleven2023waxholm,wang2020allen}}
The datasets feature high-resolution atlases of the rat and mouse brain with different modalities (T2$^*$w MRI and STPT) respectively. 
The motivation for using these datasets is to provide a benchmark for \textit{cross-species, multimodal} registration.
This addresses the growing need to map neuroanatomy across species~\cite{mezias2024establishing,beauchamp2022whole}, which is central to revealing the core evolutionary computational motifs and unique adaptations to handle specific ecological and behavioral demands.

\paragraph{Lung CT mapping challenges ~\cite{murphy2011evaluation,nlst}}
Pulmonary registration has significant clinical applications, including aligning breath-hold scans for visual comparison, modeling lung expansion, and tracking disease progression. 
Murphy et al. introduced the EMPIRE10 challenge~\cite{murphy2011evaluation} to facilitate the evaluation of CT lung registration algorithms, including inspiration-expiration, breath-hold over time, 4D, ovine, contrast-noncontrast, and artificially warped scans.
EMPIRE10 provides only scan pairs and binary lung masks, withholding fissures and landmarks for \textit{private} evaluation. 
The scans vary in spatial and physical resolution, necessitating a registration algorithm agnostic to anisotropy in both voxel and physical space. 
The National Lung Screening Trial (NLST)~\cite{nlst} subset curated by Learn2Reg challenge is another widely used community-standard dataset. 
It consists of 210 intra-subject lung pairs, with low-dose helical CT scans with limited field of view and high-dose scans with full field of view, supplemented with more than a thousand keypoints per subject pair.
This challenge provides a benchmark for comparison of methods beyond the neuroanatomical domain.

\paragraph{RnR ExM Mouse Isocortex Dataset ~\cite{rnrexm}}
The RnR-ExM challenge evaluates the ability to perform non linear deformable registration on ultra-high-resolution images.
Out of the three species (mouse brain, C. elegans, zebrafish), the mouse isocortex dataset is the only dataset with non-trivial non-linear deformations. %
Registration of high-resolution sub-micron volumes is imperative to creating and understanding the comprehensive cell atlas of the mammalian brain at scale.
The voxel size of each image volume is $2048\times2048\times81$ and the voxel spacing is 0.1625$\mu$m $\times$ 0.1625$\mu$m $\times$ 0.4$\mu$m.
These volume sizes are about two orders of magnitude larger compared to existing biomedical datasets, representing a significant challenge in quick and scalable registration.

\paragraph{AZBA and ZBrain datasets \cite{azba,zbrain}}
The ZBrain atlas is an anatomical and functional reference constructed from high-resolution confocal microscopy images of larval zebrafish (6 days post fertilization) expressing nuclear and cytoplasmic fluorescent markers.
The AZBA atlas, in contrast, represents the adult zebrafish brain at cellular resolution.
Registration of these templates enables a 
powerful cross-developmental comparison between the larval and adult zebrafish brains. 
Conceptually, this tsak establishes spatial correspondence between larval and adult brain regions, providing a foundation for developmental neuroanatomy.
We use this dataset to conduct preliminary experiments to access the generalization capabilities of registration algorithms to cross-developmental data on an unseen species and modality.

\paragraph{BICCN Mouse Dataset}
The high-throughput and high-resolution fluorescence micro-optical sectioning tomography (fMOST) platform~\cite{zheng2013visualization,fmost1} was used to image 55 mouse brains containing gene-defined neuron populations.
The brains are imaged at a resolution of $0.35\times0.35\times1.0$\micron$^3$.
The dataset is used to generate a $25$\micron-resolution atlas of the mouse brain in under 25 minutes.
This unprecedented scale, enabled by FireANTs, will advance multimodal integration, standardize cross-species comparisons, and drive scalable, reproducible neuroscience research highly pertinent to large-scale collaborative efforts such as BICCN and BICAN.

\paragraph{Learn2Reg Abdomen MRCT registration \cite{hering2022learn2reg}}
The dataset features intra-patient multimodal abdominal MRI and CT registration (122 scans in total) for diagnostic and follow-up.
We use this dataset as a testbed to ablate the effect of Jacobian-free optimization on abdominal MRCT registration.

\subsection{Results on generalization to long-tail of modalities}
Generalization to unseen modalities, species, resolutions, and anatomical organization is a 
central requirement for accessible and scalable registration algorithms.
Model-free optimization algorithms generalize well on clinical datasets, but the increased heterogenity on large-scale datasets presents a significant challenge in terms of convergence and runtime.
Deep Learning methods typically do not generalize well beyond the data distribution seen during training \cite{balakrishnan2019voxelmorph}, although domain-agnostic or foundational methods \cite{hoffmann2021synthmorph,tian2024unigradicon} have shown promising results.
Other methods \cite{vfa} claim generalization due to architectural design encoding inductive biases for the task.
Therefore, we compare FireANTs against ANTs, VoxelMorph as a DL baseline, and SynthMorph, unigradICON, and VFA as other methods that claim generalization to unseen data.

\cref{fig:intro} shows the performance on six datasets - LPBA40, NLST, Ultracortex, PRIME-DE, Zebrafish (ZBrain and AZBA), and Rodents (Waxholm and CCFv3) encompassing four evaluation criteria (Anatomical Label overlap, Landmark Distance, Mutual information of registered Intensity and Labelmap volumes).
The normalized performance is obtained by rescaling the performance of the baseline to 0 and the best performing method to 1.
The radar chart shows the generalization of FireANTs across all datasets, and individual plots show unnormalized performance on each dataset.
SynthMorph, unigradICON, and VFA perform at par with ANTs on the brain datasets (Ultracortex, PRIME-DE), but severely underperform on the Zebrafish and Rodent datasets.
Surprisingly, VFA also underperforms compared to other methods on the LPBA40 dataset, showing a potential weakness in registering highly parcelled anatomical regions.
FireANTs achieves the best performance on \textit{all} datasets, and performs significantly better on the multimodal cross-species registration task.
This establishes FireANTs as a general-purpose registration algorithm that can be used across a wide range of modalities, species, and resolutions.

\subsection{Results on state-of-the-art biomedical benchmarks}

FireANTs proposes an algorithmic improvement over the state-of-the-art ANTs toolkit.
As such, we compare FireANTs with ANTs on community standard benchmarks where ANTs is established as one of the top performing methods.
This include Klein \etal~\cite{klein2009evaluation} neuroimaging challenge, EMPIRE10~\cite{murphy2011evaluation} pulmonary challenge.
We also add comparisons on two contemporary benchmarks for neuroanatomy (OASIS) and pulmonary (NLST) datasets from the Learn2Reg~\cite{hering2022learn2reg} challenge, which is of high relevance to the neuroimaging and pulmonary communities.

\paragraph{In-vivo brain MRI mapping}
We compare FireANTs with two state-of-the-art optimization algorithms on Klein \etal~\cite{klein2009evaluation}: ANTs - which won the original challenge, and Symmetric Log Demons~\cite{vercauteren2007diffeomorphic}, and two widely used deep learning algorithms: VoxelMorph~\cite{balakrishnan2019voxelmorph} and SynthMorph~\cite{hoffmann2021synthmorph} using their provided pretrained models.
Since no methods utilize label maps, we run registration on all 414 image pairs prescribed in the dataset.
Results for the brain datasets are shown in ~\cref{fig:meanvalklein}, ~\cref{fig:regionwise-dice-brain}, and ~\cref{fig:oasis}. 
Our algorithm outperforms all baselines on four out of five datasets, with an improvement in \textit{all} metrics evaluating the volume overlap of the fixed and warped label maps.
The improvements are consistent across varying parcellations and relative sizes of anatomical label maps. %
In the IBSR18 and CUMC12 datasets, the median target overlap of our method is better than the third-quantile of ANTs. %
~\cref{fig:regionwise-dice-brain} also highlights the improvement in label overlap per labeled brain region across all datasets. 
For deep learning methods, a noticable performance drop is observed when the anisotropic volumes are fed into the network, which is undesirable as the trained model is essentially `locked' to a single physical resolution - which limits the generalizability of the model to various modalities with different physical resolutions.
For Demons, ANTs, and FireANTs (Ours), we do not perform any additional normalization or resampling.
On the OASIS dataset, all methods perform at par with each other with no significant differences.
SynthMorph is more robust to the domain gap than VoxelMorph due to its training strategy with synthetic images, but still underperforms optimization baselines when their recommended hyperparameters are chosen. %

\begin{figure}[p]
\centering    
\begin{minipage}{\linewidth}
    \subcaption{
        Following the evaluation setup of Klein {\etal} paper, we validate registration performance using the average volume overlap of all anatomical label maps between the fixed and warped label maps.
        We consider ANTs (the winner of the challenge), and Diffeomorphic Demons as state-of-the-art optimization algorithms, and Voxelmorph and Synthmorph as state-of-the-art unsupervised deep learning baselines.
        Evaluation is shown for five metrics with $\uparrow$ denoting a higher score is better, and $\downarrow$ signifying a lower score is better.
        For deep learning baselines, appropriate preprocessing (intensity normalization, alignment, and resampling to 1mm isotropic) is performed to ensure a fair comparison, whereas no such preprocessing is required for optimization methods, including FireANTs.
        FireANTs shows significant gains in performance that are consistent across all four datasets, with the median overlap scores outperforming the third quartile of all other methods for IBSR18 and CUMC12 datasets.
        Comparison of overlap metrics by specific anatomical regions are in \cref{fig:regionwise-dice-brain}.
        For the overlap aggregation mentioned in~\cite{klein2009evaluation}, results are shown in~\cref{fig:braintable_klein}.
    }
    \label{fig:meanvalklein}
    \begin{minipage}{\linewidth}
    \includegraphics[width=0.48\linewidth]{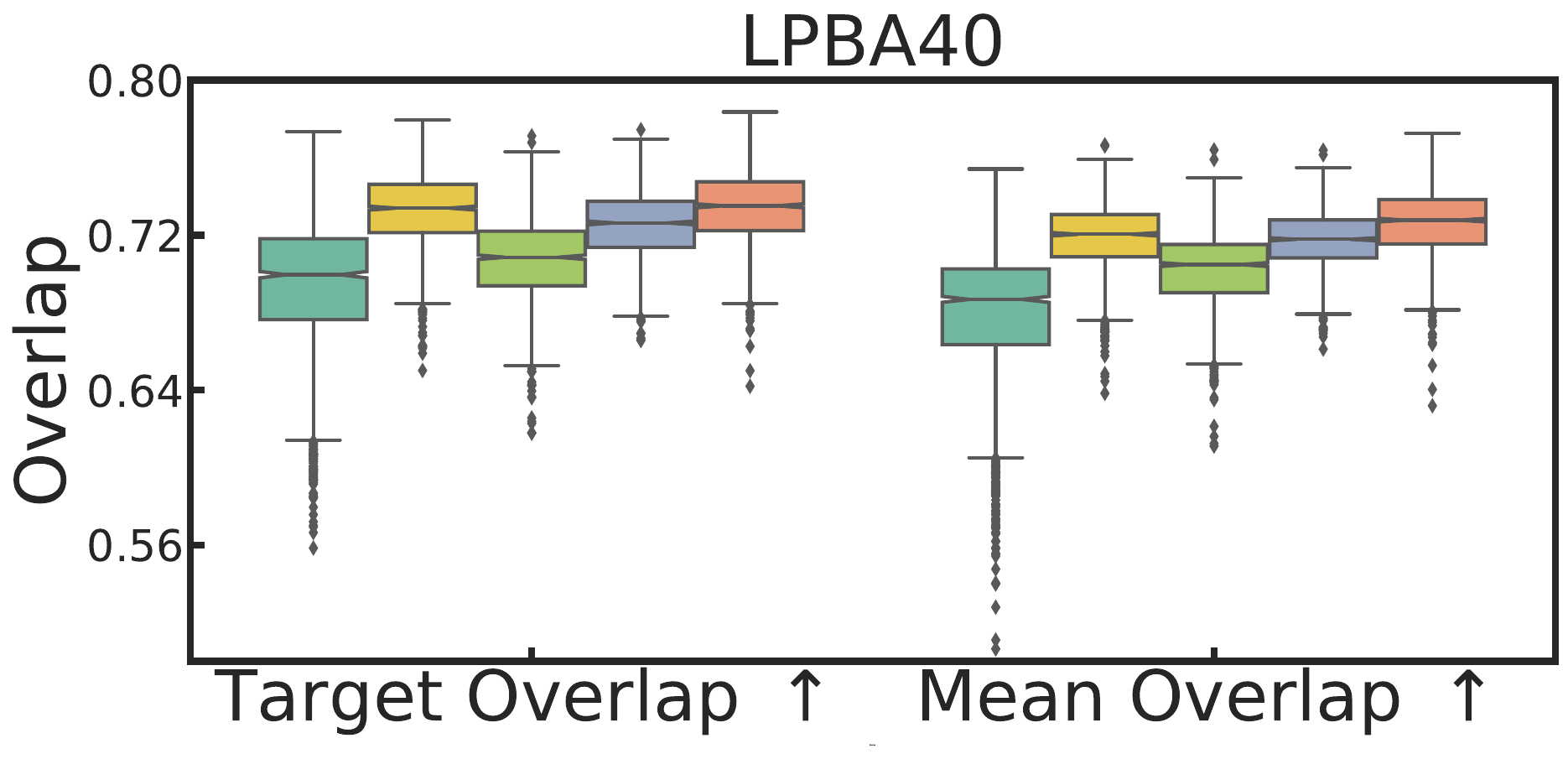}
    \includegraphics[width=0.48\linewidth]{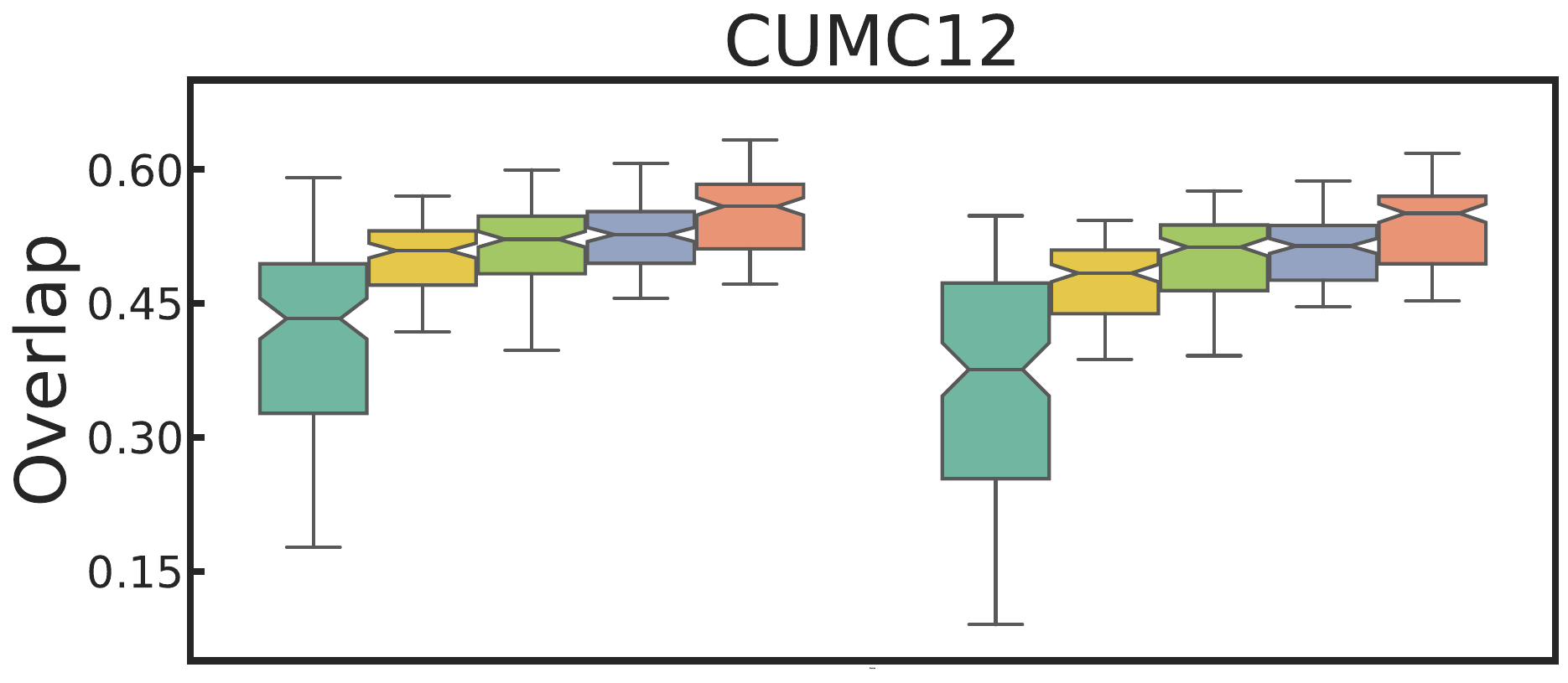} 
    \end{minipage}
    \begin{minipage}{\linewidth}
    \includegraphics[width=0.48\linewidth]{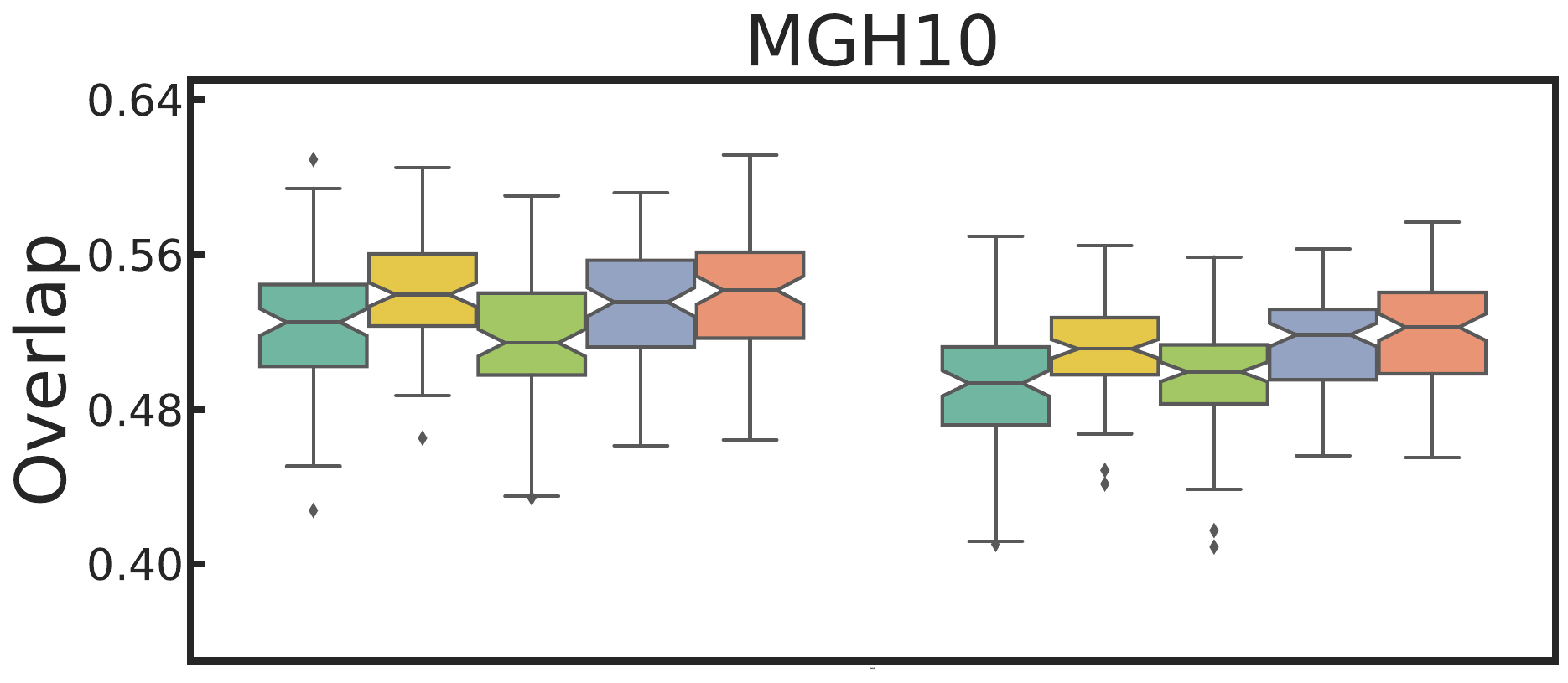}
    \includegraphics[width=0.48\linewidth]{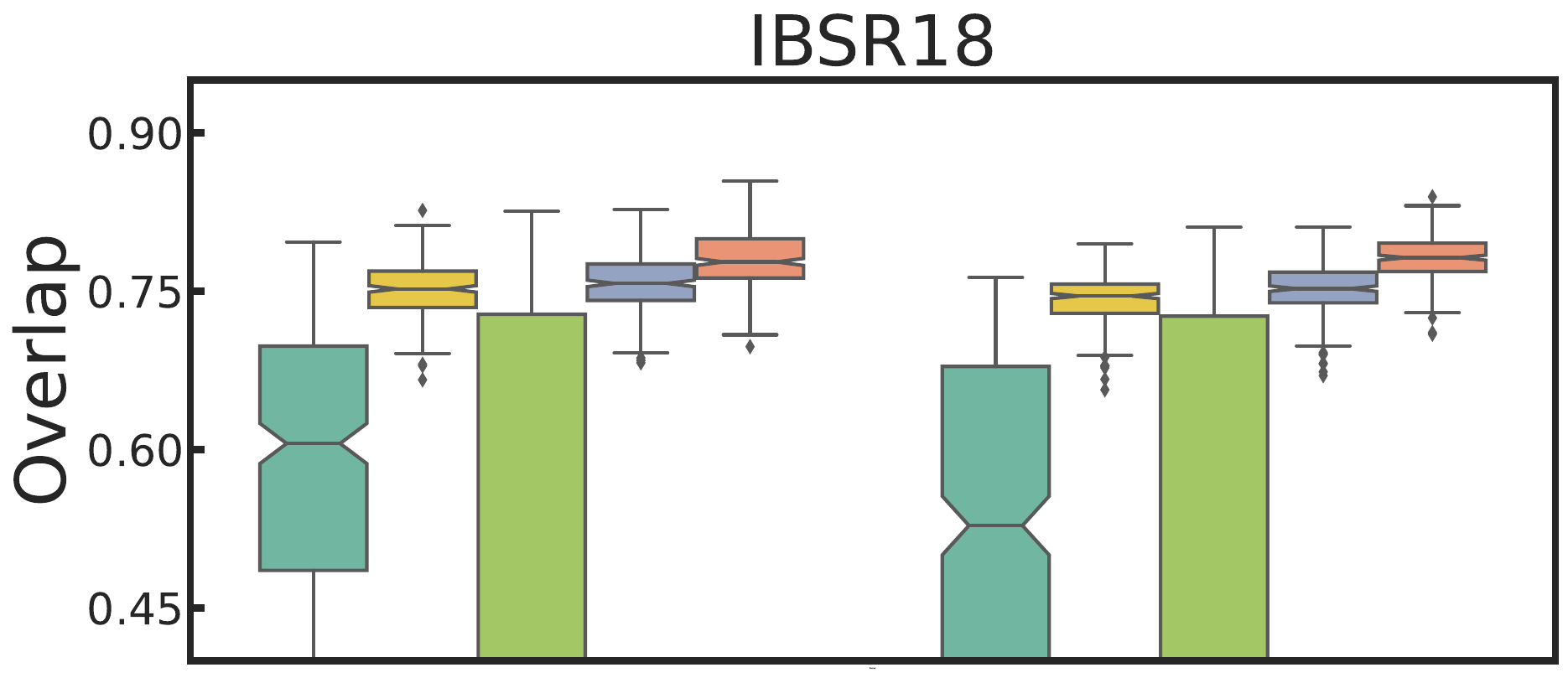}
    \end{minipage}
\end{minipage}

\begin{minipage}{0.55\linewidth}
    \subcaption{\footnotesize FireANTs achieves substantially lower inter-quartile range of fissure errors, defined as the percentage of marked pixels that are registered to points on the opposite side of the fissure boundary.}
    \label{fig:fissureempire}
    \includegraphics[width=\linewidth]{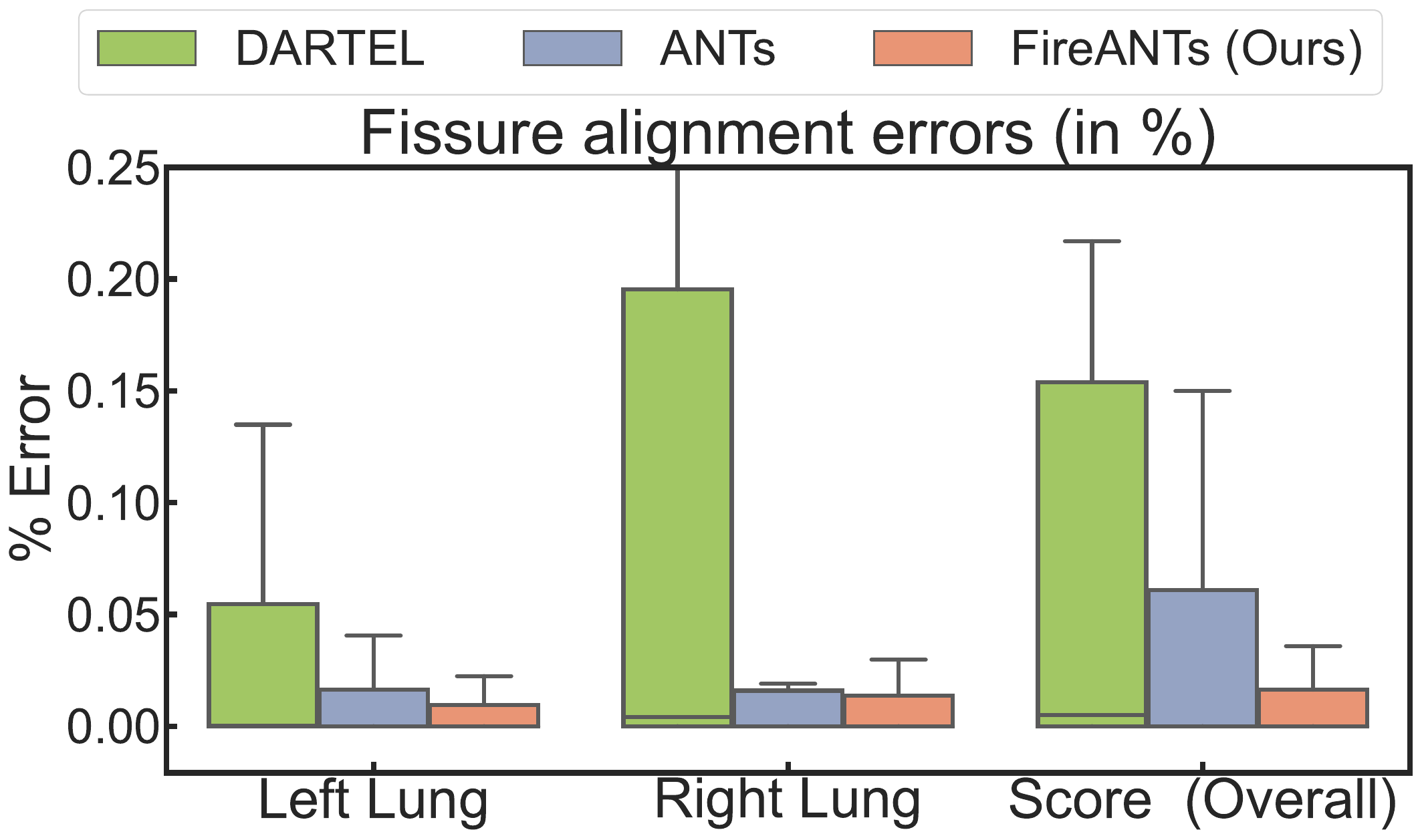}
\end{minipage}
\vspace*{-13pt}
\begin{minipage}{0.44\linewidth}
\centering
\resizebox{\linewidth}{!}{%
\def\arraystretch{1.6}
\begin{tabular}{lccc}
    \toprule
     \textbf{\% Error} & \textbf{DARTEL} & \textbf{ANTs} & \textbf{Ours} \\
     \midrule
     Left Lung & 3.9983 & 0.0069 & \textbf{0.0000}  \\
     Lower Lung & 2.7514 & 0.0177 & \textbf{0.0000} \\
     Right Lung & 2.4930 & 0.0107 & \textbf{0.0000} \\
     Upper Lung & 5.2037 & \textbf{0.0000} & \textbf{0.0000} \\
     Score (Overall) & 3.0681 & 0.0088 & \textbf{0.0000} \\
     \bottomrule
\end{tabular}
}
\subcaption{ %
\footnotesize Singularity errors are defined as fraction of voxels that define a non-invertible deformation. Singularity quantifies the percentage of implausible deformations.
FireANTs achieves \textit{zero percent} singularity errors. 
}
\label{fig:singularempire}
\end{minipage}
\end{figure}

\begin{figure}[p]\ContinuedFloat
\begin{minipage}{\linewidth}
    \centering
    \includegraphics[width=0.95\linewidth]{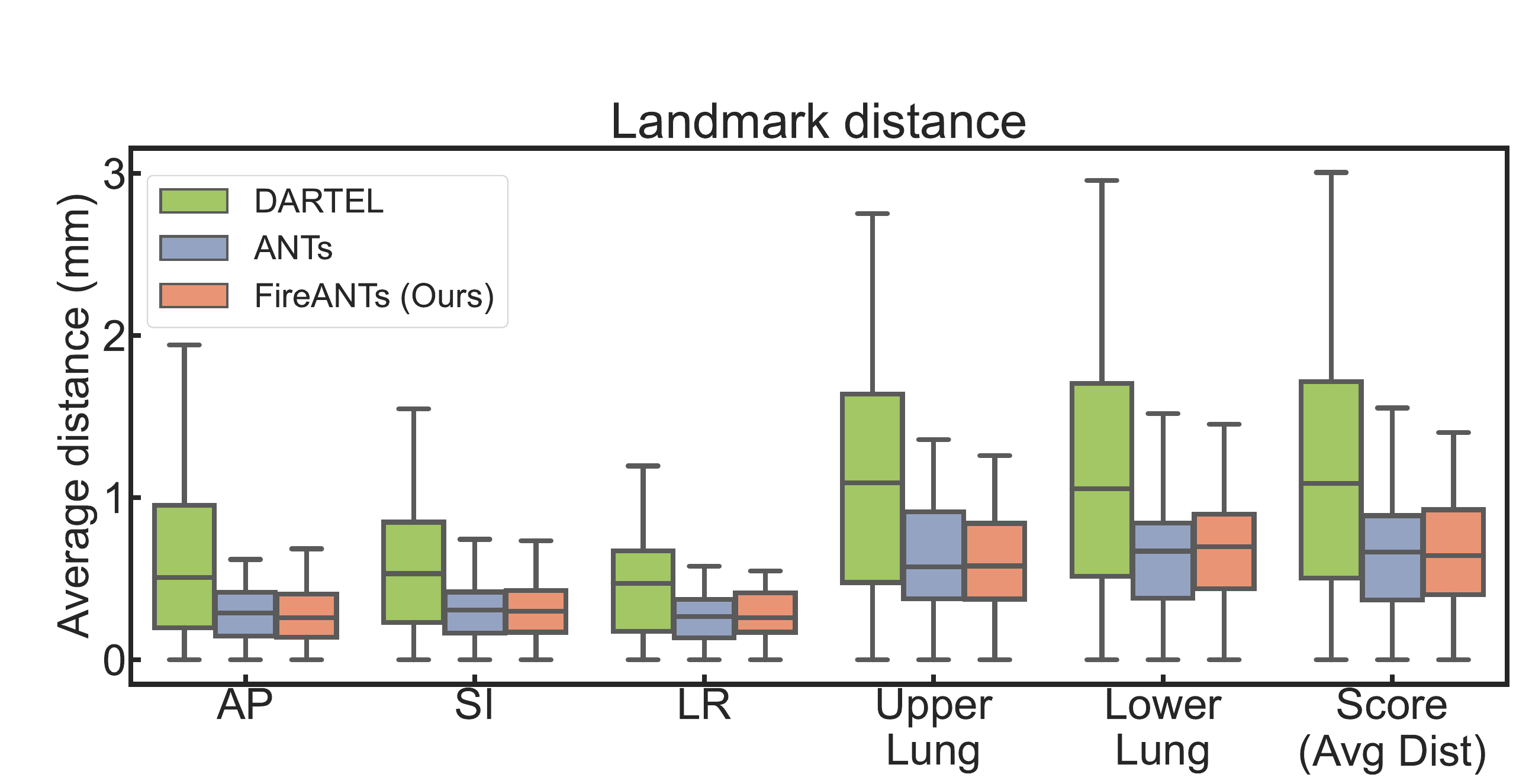}
    \vspace*{-5pt}
    \subcaption{ %
    Landmark distance is the Euclidean distance between well-dispersed landmark points between the fixed and warped images.
    FireANTs has a lower median and narrower interquartile range than baselines on five out of six subregions.
    }
    \label{fig:landmarkempire}
    \end{minipage}
\begin{minipage}{0.58\linewidth}
\centering
\resizebox{\linewidth}{!}{%
\begin{tabular}{lrrr}
\toprule
\textbf{Method} &  \textbf{Left Lung} &  \textbf{Right Lung} &  \textbf{Score } \\
& & & \textbf{(\% Error Overall)} \\
\midrule
FireANTs (Ours) &             \cellcolor{best_color}0.0185 &              \cellcolor{better_color}0.0254 &                    \cellcolor{best_color} 0.0227 \\
MRF Correspondence Fields  &             0.0824 &            \cellcolor{best_color}  0.0211 &                    \cellcolor{better_color}0.0485 \\
ANTs &             \cellcolor{better_color}0.0249 &              0.1016 &                    \cellcolor{good_color}0.0747 \\
Dense Displacement Sampling   &             \cellcolor{good_color}0.0578 &              0.0919 &                    0.0826 \\
ANTs + BSpline &             0.0821 &              0.0848 &                    0.0861 \\
DISCO &             0.1256 &              \cellcolor{good_color}0.0499 &                    0.0882 \\
VIRNet &             0.0834 &              0.0934 &                    0.0890 \\
Feature-constrained nonlinear registration &             0.1210 &              0.0758 &                    0.1032 \\
Explicit Boundary Alignment   &             0.1063 &              0.1246 &                    0.1209 \\
MetaReg    &             0.1049 &              0.2224 &                    0.1791 \\
\bottomrule
\end{tabular}
}
\subcaption{Fissure alignment error on top 10 algorithms in the challenge sorted by fissure alignment error, averaged on all scan pairs.
FireANTs outperforms a wide array of baselines, including direct optimization (ANTs, ANTs+BSpline), neural networks (VIRNet), and explicit correlation volumes (MRF, Disco).
}
\label{fig:fissureallmethods}
\end{minipage}
\begin{minipage}{0.4\linewidth}
    \resizebox{\linewidth}{!}{%
    \begin{tabular}{lc}
        \toprule
        \multicolumn{2}{c}{\textbf{Validation metrics on NLST}} \\ \hline
        \textbf{Method} & \textbf{TRE30 (in mm)} \\ \hline
        Zero displacement (Baseline) & 9.76 \\
        VoxelMorph~\cite{balakrishnan2019voxelmorph} & 4.12 \\
        Im2Grid~\cite{im2grid} & 3.05 \\
        ANTs & 3.04 \\ 
        Vector-Field Attention~\cite{vfa} & 2.31 \\
        RWC-Net~\cite{rwcnet} & 2.11 \\
        unigradICON~\cite{tian2024unigradicon} & 2.07 \\
        unigradICON + instance optimization & 1.77 \\
        FireANTs (Ours) & \textbf{1.18} \\ \bottomrule
    \end{tabular}
    }
    \subcaption{Robust landmark distance (TRE30) comparison of state-of-the-art algorithms highlights the effective performance of FireANTs on the NLST dataset, outperforming a plethora of state-of-the-art optimization and deep learning baselines.}
    \label{fig:nlst}
\end{minipage}
\caption{\footnotesize \textbf{FireANTs demonstrates state-of-the-art performance on community-standard neuroimaging and pulmonary  challenges}:
\textbf{(a) EMPIRE10}: Lung fissure plates are an important anatomical landmark demarcating lobes within the lung. Fissure alignment errors (in \%) denote the percentage of locations near the lung fissure plates that are on the wrong side of the fissure post-registration. Over all 30 scan pairs, our method performs 5$\times$ better than ANTs.
\textbf{(b) EMPIRE10}: Singularity errors defined as percentage of voxels that have a non-diffeomorphic deformation, a proxy for physically implausible deformations.
In the DARTEL baseline, singularities can be introduced for larger deformations due to numerical approximations of the integration. 
Even for ANTs, the solutions (deformations) returned are not entirely diffeomorphic.
Our method shows much better fissure and landmark alignment (\cref{fig:lungtable}\textbf{(a,c)}, ~\cref{fig:lung-images-1},~\cref{fig:lung-images-2}) with guaranteed diffeomorphic transforms.
\textbf{(c) EMPIRE10}: Landmark distance in mm for selected landmarks. Across different lung subregions, our method shows results at least at par with ANTs, with slightly better average and median results across all regions.
\textbf{(d) EMPIRE10}: Shows the top 10 algorithms for average fissure alignment error in \% in the EMPIRE10 challenge. Error metrics are taken from the evaluation server. 
Other methods perform well on one lung (MRF for right, ANTs for left) but comparatively poorly on the other lung, compared to our method showing both accurate and robustness to both the left and right lung.
\colorbox{best_color}{\hspace{0.15cm} \vphantom{X}} = First, \colorbox{better_color}{\hspace{0.15cm} \vphantom{X}} = Second, \colorbox{good_color}{\hspace{0.15cm} \vphantom{X}} = Third best result.
\textbf{(e) NLST}: Landmark distance in mm for provided landmarks. Our method outperforms a variety of state-of-the-art optimization and deep learning algorithms.
}
\label{fig:lungtable}
\end{figure}

\paragraph{Lung CT mapping challenges} 
The EMPIRE10 lung dataset\cite{murphy2011evaluation} consists of volumes that are about 10$\times$ larger than the brain dataset, thereby presenting a scaling challenge for deformable registration algorithms. 
Evaluation is done with privately withheld labels; we use the provided results from the leaderboard to compare with other methods.
We evaluate three criteria: (1) fissure alignment errors (\%)—the fraction of misaligned fissure voxels (\cref{fig:fissureempire,fig:fissureallmethods}), (2) landmark distance in mm (\cref{fig:landmarkempire}), and (3) singularity errors—the fraction of non-diffeomorphic voxels (\cref{fig:singularempire}). ~\cref{fig:lungtable} highlights the impact of representation choice in modeling diffeomorphisms. DARTEL, using an exponential map, performs significantly worse than ANTs across all metrics by three orders of magnitude. In contrast, our method reduces fissure alignment error by 5$\times$ compared to ANTs and outperforms it in 5 out of 6 landmark subregions.
While all methods theoretically ensure diffeomorphism, SVF-based approaches introduce singularity errors due to non-adaptive scaling-and-squaring.
We discuss the numerical limitations of SVF-based approaches in ~\cref{sec:limitations-stationary-velocity-fields}.
ANTs also introduces some singularities, whereas our method computes numerically perfect diffeomorphic transforms. 
Finally,~\cref{fig:fissureallmethods} compares fissure alignment errors among EMPIRE10 submissions, showing FireANTs achieves the lowest landmark errors and the fastest runtime among the top 10 methods, setting new benchmarks in computational efficiency and accuracy.
\textbf{NLST}:
For the NLST dataset\cite{nlst}, we compare with representative state-of-the-art optimization and deep-learning baselines.
We use the evaluation criteria provided by the challenge, and measure results on the Robust Target Registration Error (TRE30) in millimeters between the registered keypoints.
Results in \cref{fig:nlst} show that FireANTs outperforms all baselines on the NLST dataset, with improvements of upto 51.6\% in robust target registration error (TRE30) of provided keypoints compared to state-of-the-art deep learning benchmarks including Im2Grid, Vector-Field Attention, RWC-Net, and a 50.8\% improvement in TRE30 over foundation models like unigradICON.
This demonstrates the broad applicability of FireANTs beyond neuroimaging applications.

\subsection{Evaluation on high-resolution mouse isocortex registration}
Expansion Microscopy (ExM) is an emerging super-resolution fluorescence imaging technique that enables 3D nanoscale visualization of cellular and molecular structures~\cite{chen2015expansion}. 
While ExM provides rich structural data, its large-scale images remain challenging for existing registration algorithms due to repetitive textures, highly non-linear hydrogel deformations, imaging noise, and size constraints. 
The Robust Non-rigid Registration Challenge for Expansion Microscopy (RnR-ExM)~\cite{rnrexm} offers a benchmark dataset, where we focus on registering mouse isocortex images, characterized by hydrogel-induced deformations and staining intensity loss. 
Each volume ($2048\times2048\times81$ voxels) has a voxel spacing of {0.1625\micron$\times$0.1625\micron$\times$0.4\micron} and is 40.5 times larger than brain imaging datasets. 
Current state-of-the-art methods either register small independent chunks~\cite{bigstream}, losing inter-chunk information, or process highly downsampled images~\cite{lku}, significantly reducing resolution (by 64$\times$ in-plane).

\begin{figure}[p]
    \centering
    \begin{minipage}{\linewidth}
    \centering
    \subcaption{Snapshot of the RnR-ExM leaderboard}
    \includegraphics[width=0.9\linewidth]{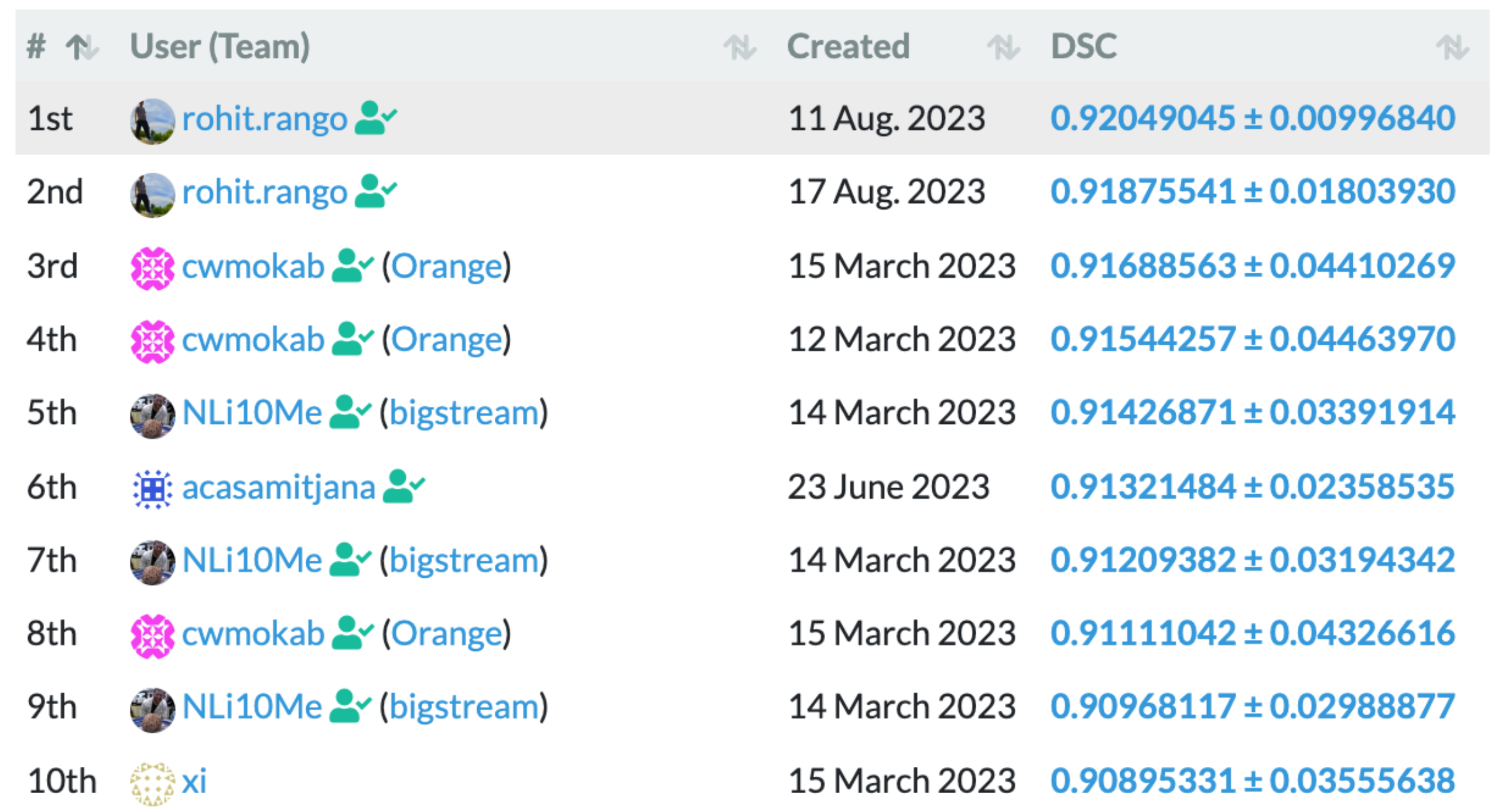}
    \end{minipage}
    \begin{minipage}{\linewidth}
    \subcaption{Qualitative comparison of registration of Bigstream and FireANTs}
    \includegraphics[width=\linewidth]{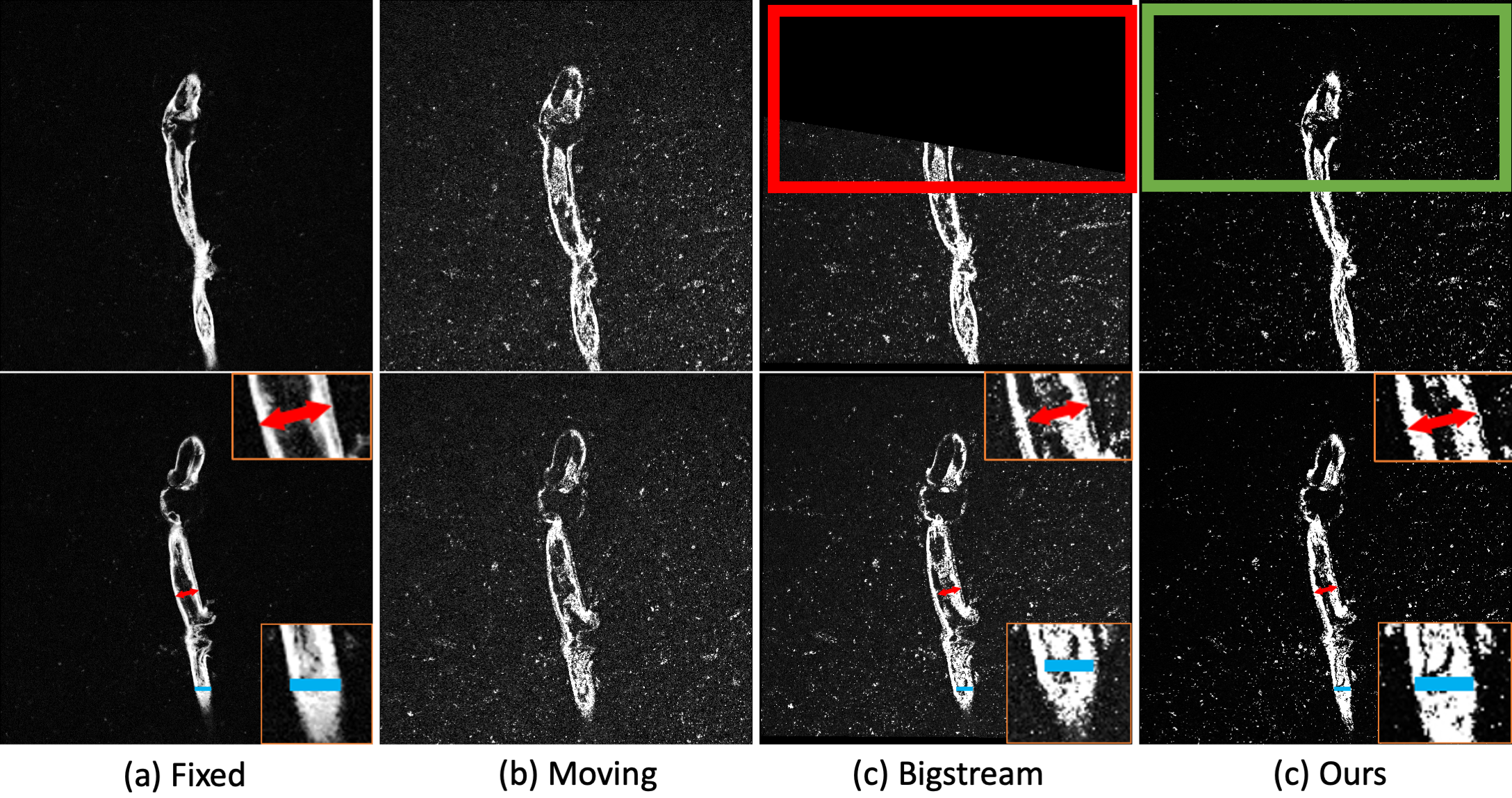}
    \end{minipage}
    \caption{\footnotesize \textbf{FireANTs secures first rank in the RnR-ExM mouse dataset}:
    \textbf{(a)}: As of \today, our method ranks first in the mouse brain registration task, which is the only task in the challenge requiring deformable registration.
    Our top two successful submissions secure the first and second position, with a 0.361 improvement in Dice score compared to the 3rd ranked submission, which is 
    0.261 better than the 5th ranked submission (bigstream).
    Note that among the top 10 submissions, our method has the lowest standard deviation (4.42$\times$ lower than the second best submission) showing the robustness of our model across different microscopy volumes.
    \textbf{(b)} shows a qualitative comparison of FireANTs with Bigstream ~\cite{bigstream}, the other top leading method in the challenge.
    The moving image volumes have substantially more noise than the fixed image volumes, making intensity-based registration difficult.
    The non-rigid deformation dynamics of the hydrogel are clearly visible, as the moving volume has a thicker boundary than the fixed volume.
    Bigstream does not capture these dynamics very well -- this is illustrated by comparing the thickness of the cortex at various points (zoomed {\textcolor{orange}{orange}} crops in bottom row), where Bigstream does not deform the cortex enough to match the fixed image.
    FireANTs deforms and accurately depicts these morphological changes, which can be crucial for downstream morphometric studies.
    Moreover, the affine registration in Bigstream knocks the boundary slices out of the volume ({\textcolor{red}{red}} highlight in top row), leading to drop in registration performance.
    On contrary, our method's affine and deformable stages are more stable, leading to better registration and avoiding spurious out-of-bound artifacts at the boundary slices.
    }
    \label{fig:rnrexm}
\end{figure}

In contrast, FireANTs is able to register the volume at native resolution. 
We perform an affine registration followed by a diffeomorphic registration step. 
The entire method takes about 2-3 minutes on a single A6000 GPU.
As shown in ~\cref{fig:rnrexm}, our method secures the first place on the leaderboard, with a considerable improvement in the Dice score and a 4.42$\times$ reduction in the standard deviation of the Dice scores compared to the next best method.
~\cref{fig:rnrexm} also shows qualitative comparison of our method compared to Bigstream~\cite{bigstream}, the winner of the RnR-ExM challenge.
Bigstream performs only an affine registration, leading to inaccurate registration in one of three test volumes, leading to a lower average Dice score and higher variance.
Moreover, the affine registration leads to boundary in-plane slices being knocked out of the volume, leading to poor registration (\cref{fig:rnrexm}).
FireANTs preserves the boundary in-plane slices during its affine step, and subsequently performs an accurate diffeomorphic registration at submicron resolution leading to accurate registration with substantially lower variance.
This experiment demonstrates the versatility and applicability of FireANTs for high-resolution microscopy registration.

\subsection{Runtime and Memory Efficiency Analysis}
One of the critical bottlenecks for scalable registration with ANTs is the prohibitively large runtimes for single-threaded CPU registration \cite{balakrishnan2019voxelmorph}.
Deep learning methods aim to reduce the runtime by performing feedforward inference, but these methods in turn have steep memory requirements due to activation overheads~\cite{ultrascaleplaybook,reducingactivation}, making them infeasible for high-resolution registration.
FireANTs circumvents both these issues using a lightweight implementation on GPU.

\begin{figure}[h!]
\centering
\begin{minipage}{\linewidth}
\subcaption{Runtime analysis and comparison with ANTs on CPU and GPU on five brain datasets}
\centering
\resizebox{\linewidth}{!}{%
\begin{tabular}{lccccccc} \toprule
\multirow{2}{*}{\textbf{Dataset}} & \multirow{2}{*}{\textbf{ANTs}} & \multicolumn{3}{c}{\textbf{Ours (asymmetric)}} & \multicolumn{3}{c}{\textbf{Ours (symmetric)}} \\ 
 & & \textbf{CPU (s)} & \textbf{GPU (s)} & \textbf{Speedup} & \textbf{CPU (s)} & \textbf{GPU (s)} & \textbf{Speedup} \\ \hline
 IBSR18 & $324.39 \pm 16.67$ & $119.65 \pm 9.31$ & $1.06 \pm 0.04$ & $2.71 / 305.76$ & $399.64 \pm 2.58$ & $2.04 \pm 0.02$ & $0.81 / 159.30$ \\
 CUMC12 & $484.29 \pm 64.14$ & $83.85 \pm 4.08$ & $1.09 \pm 0.07$ & $5.78 / 442.43$ & $176.64 \pm 7.06$ & $2.09 \pm 0.03$ & $2.74 / 231.29$ \\
 MGH10 &  $329.95 \pm 119.37$ & $46.75 \pm 0.84$ & $1.13 \pm 0.06$ & $7.06 / 291.84$ & $98.17 \pm 1.76$ & $1.19 \pm 0.44$ & $3.36 / 278.03$ \\
 LPBA40 & $227.30 \pm 37.37$ & $153.76 \pm 10.04$ & $1.22 \pm 0.06$ & $1.48 / 186.61$ & $272.98 \pm 20.33$ & $2.30 \pm 0.12$ & $0.83 / 98.74$ \\ 
 OASIS & $131.06 \pm 9.01$ & $54.47 \pm 1.44$ & $0.75 \pm 0.01$ & $2.41 / 174.75$  & $94.92 \pm 1.018$  & $1.62 \pm 0.11$ & $1.38 / 80.90$\\
\bottomrule
\end{tabular}
}
\label{fig:timingbrain}
\end{minipage}
\begin{minipage}{\linewidth}
\subcaption{Runtime analysis and summary on EMPIRE10 dataset. ANTs and DARTEL take significantly longer to run due to the high-resolution nature of the CT lung dataset. FireANTs runs a \textit{minimum} of 320 times faster than ANTs, saving a substantial amount of time, at no loss (in fact, with substantial gains) in registration quality.}
\label{fig:timinglung}
\centering
\resizebox{0.9\linewidth}{!}{%
\begin{tabular}{lrrrrr} \toprule
    & \textbf{ANTs} & \textbf{DARTEL} & \textbf{Ours} & \textbf{Speedup (ANTs)} & \textbf{Speedup (DARTEL)} \\ \midrule
    Avg     & 6hr 14m & 7hr 16m & 0m 39s & 562.67 & 663.77 \\
    Min & 0h 55m & 1h 8m & 0m 9s & 320.74 & 315.23 \\
    Max & 12h 41m & 10h 11m & 1m 5s & 1231.27 & 796.51 \\
    \bottomrule
\end{tabular}
}
\end{minipage}

\begin{minipage}{\linewidth}
\centering
\subcaption{Tradeoff between accuracy, runtime, and memory requirements with state-of-the-art deep learning methods.}
\label{fig:deeplearning-efficiency}
\includegraphics[width=0.32\linewidth]{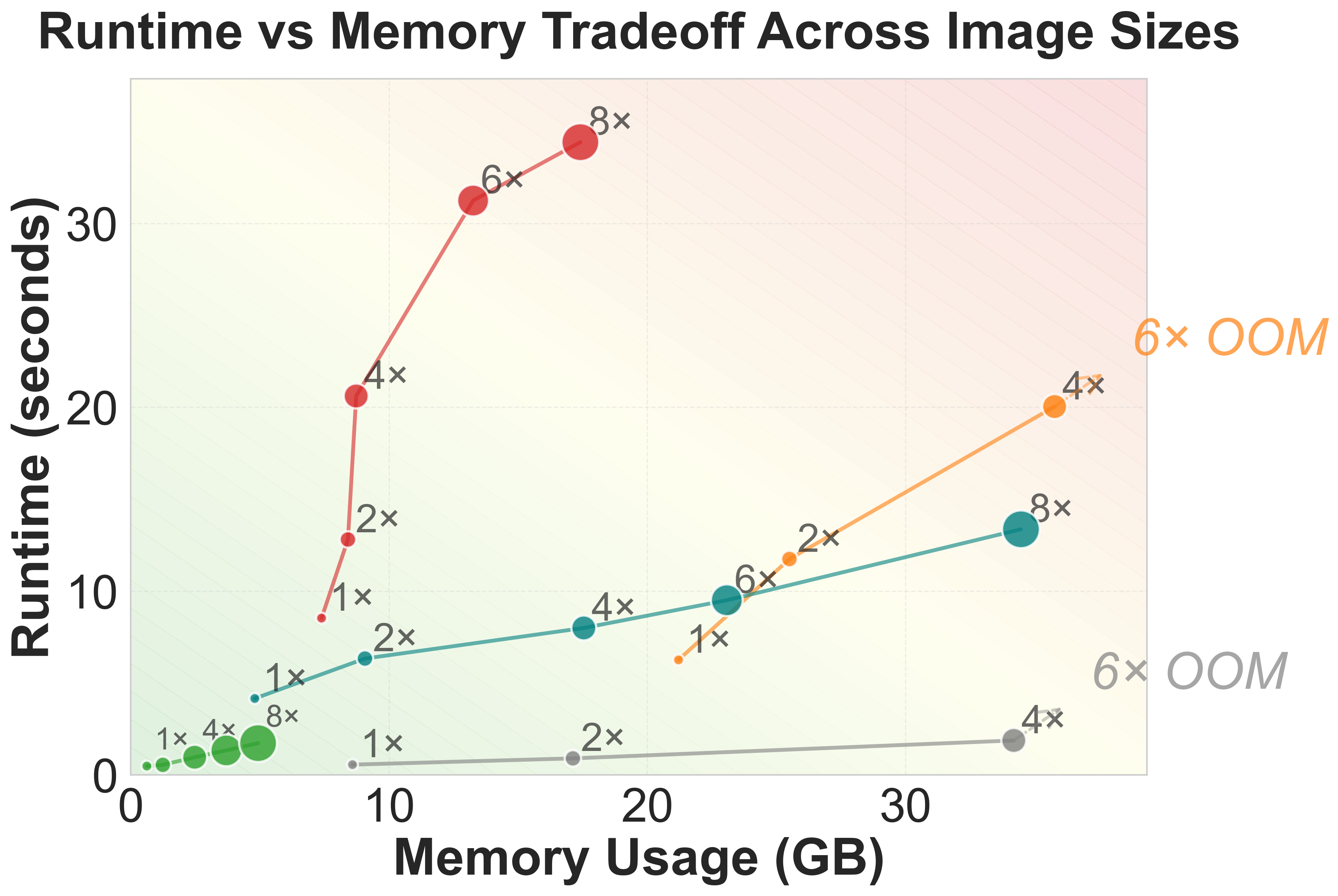}
\includegraphics[width=0.32\linewidth]{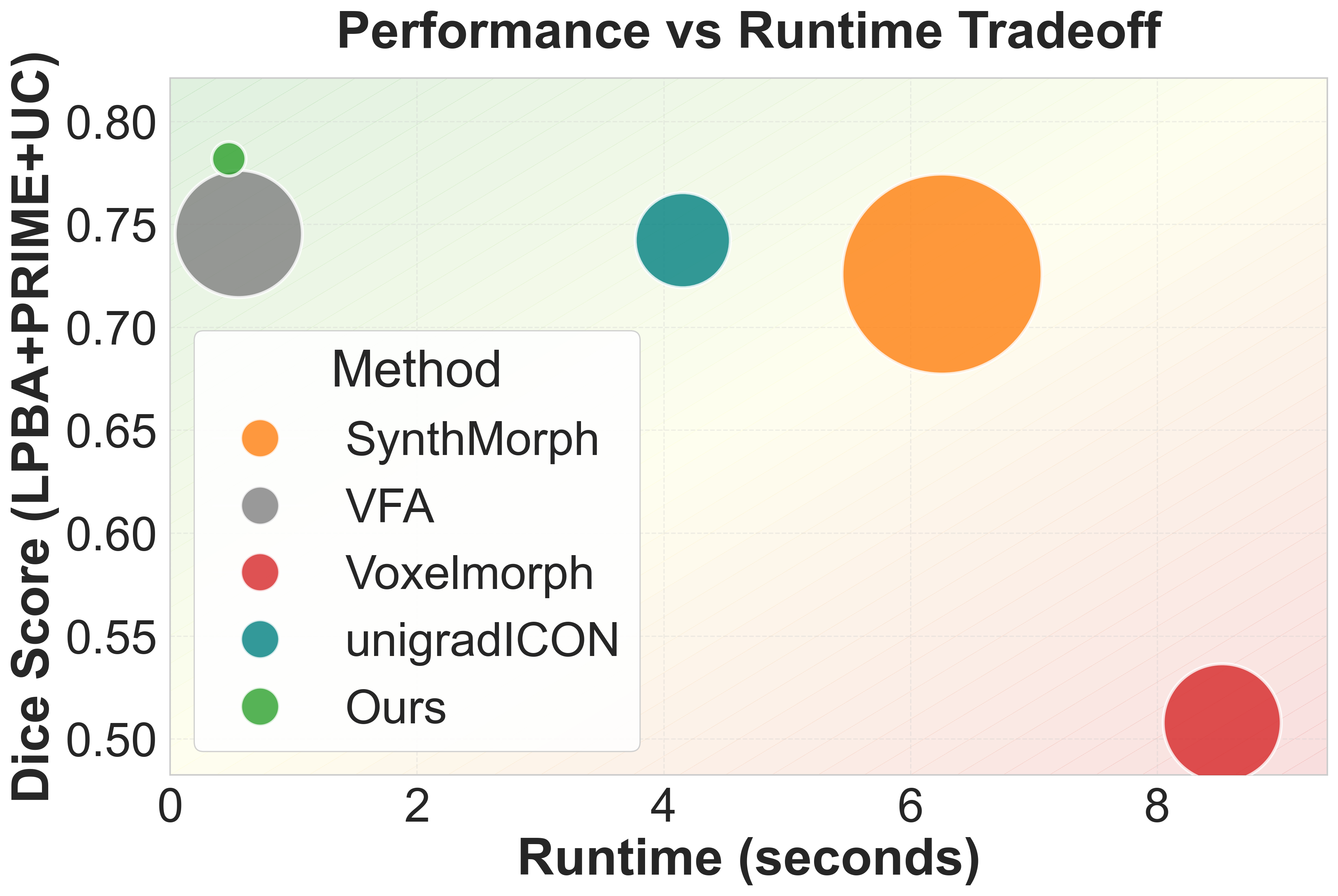}
\includegraphics[width=0.32\linewidth]{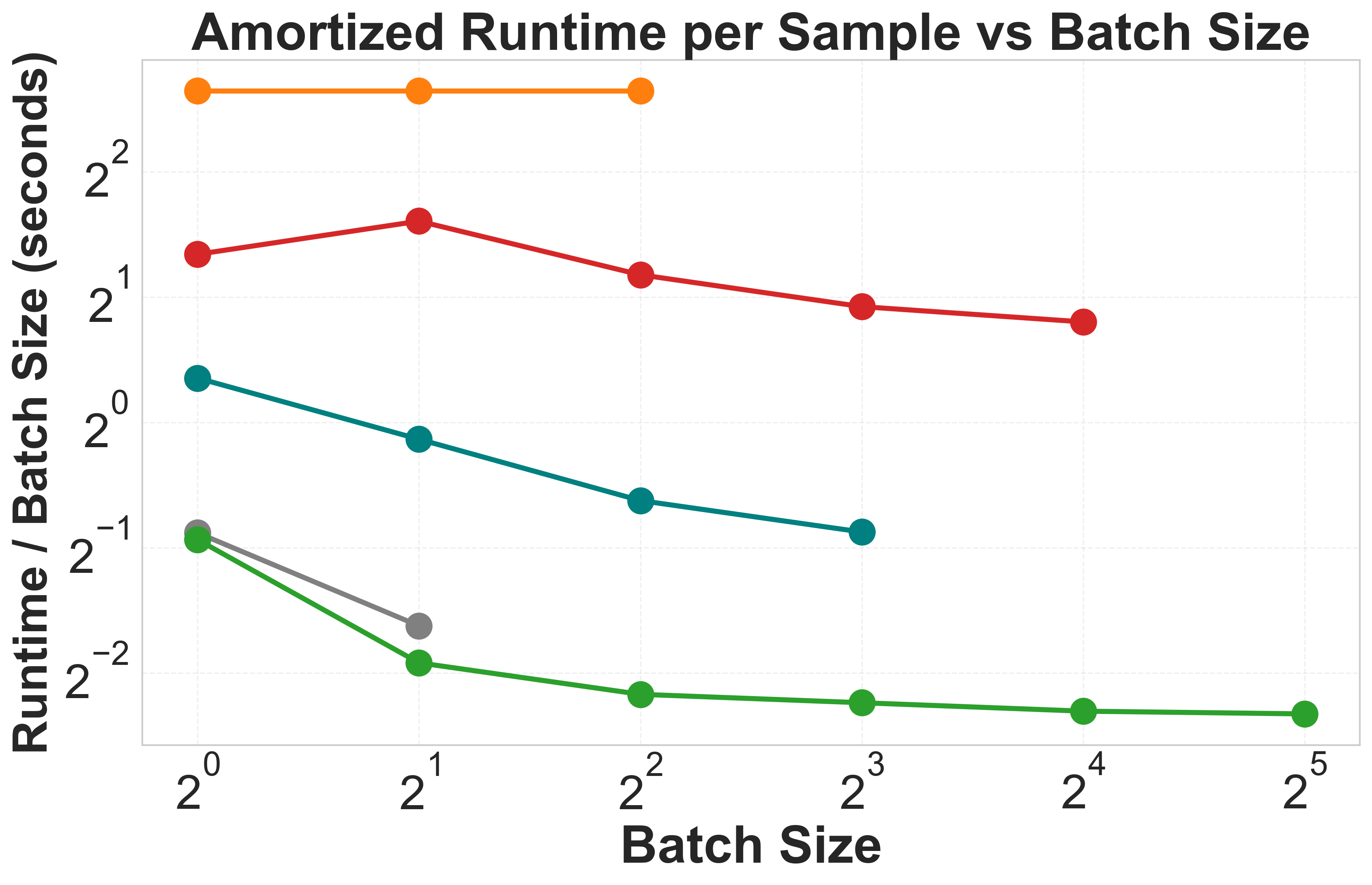}
\end{minipage}

\caption{\footnotesize \textbf{FireANTs facilitates quick and scalable registrations.} We compare the runtime of our implementation with the ANTs library.
\textbf{(a)} shows histogram of speedup (runtime of ANTs divided by runtime of our method) and statistics of runtimes (in seconds) for the four brain MRI datasets.
For all datasets, our implementation runs a \textit{minimum} of two orders of magnitudes faster, making it suitable for hyperparameter search algorithms, and larger datasets.
Table \textbf{(b)} shows the runtime of ANTs, DARTEL and our implementation on the EMPIRE10 challenge data. The first three colums show the actual runtime of the methods, followed by the speedup obtained by our method when compared to ANTs and DARTEL. Note that our method runs a \textit{minimum} of 320 times faster than ANTs, saving a substantial amount of time, at no loss in registration quality.
\textbf{(c)} shows the runtime and memory requirements of our method compared to deep learning methods. \textbf{Left} shows the runtime and memory requirements of our method compared to deep learning methods for increasing problem sizes. FireANTs is upto $10\times$ more memory efficient than SOTA deep learning methods, while performing faster than almost all of them at inference. \textbf{Middle} shows the plot of average performance over three brain datasets compared with average runtime, with the size of the bubble indicating average memory usage.
FireANTs performs \textit{better} while being faster and more memory efficient than all deep learning methods, indicating that a tradeoff is not necessary for good performance. \textbf{Right} shows that further gains in amortized runtime are possible by increasing the batch size at inference. FireANTs achieves less than 0.25 seconds per image pair and runs more than double the number of image pairs copmared to all other deep methods, showing unprecedented efficiency for high-throughput registration.
}
\label{fig:runtime}
\end{figure}

\paragraph{Runtime compared to ANTs}
We evaluate the efficiency of FireANTs compared to ANTs by running both algorithms on the CPU with 32 threads, with identical multi-scale optimization settings. 
Furthermore, we run FireANTs on the GPU with a batch size of 1 to avoid amortizing runtimes over larger batches.
The runtimes on all five brain datasets (IBSR18, CUMC12, MGH10, LPBA40, and OASIS) are shown in \cref{fig:timingbrain}.
FireANTs is upto $7\times$ faster than ANTs on CPU, and 442$\times$ faster on GPU.
The runtime improvement on the CPU can be attributed to faster convergence and better implementation of the optimization since both methods are run with identical multi-scale optimization settings and capped at the same CPU resources.
\cref{fig:timinglung} shows the runtime of FireANTs compared to ANTs on the EMPIRE10 dataset.
Since the runtime for ANTs and DARTEL are provided in the submission writeup without details on hardware or number of threads, it is not possible to reproduce the same results, and use their provided numbers as expected runtime for the dataset.  
However, FireANTs runs an average of 560$\times$ faster than ANTs on the GPU, which is at par with the runtime improvements on neuroimaging datasets.

\paragraph{Runtime and memory requirements compared to deep learning methods}
FireANTs is highly efficient compared to deep learning methods, both in terms of runtime and memory usage.
We highlight the efficiency using three experiments:
\begin{itemize}
    \item \textbf{Accuracy-Runtime tradeoff}: We compare the performance, runtime, and memory usage of FireANTs with SOTA deep learning methods averaged over three neuroimaging datasets (LPBA40, Ultracortex, PRIME-DE).
    \item \textbf{Runtime and Memory requirements with increasing problem sizes}: We take an OASIS MRI image pair and progressively upsample it by factors of $2\times, 4\times, 6\times, 8\times$ larger and measure the runtime and memory usage of each method.
    \item \textbf{Amortized runtime with increasing batch sizes}: Amortized runtime over larger batches can improve GPU utilization and hide kernel launch and activation cache loading overheads. 
    We perform batched inference using an OASIS MRI volume for increasing batch sizes, and plot the amortized runtime (i.e. runtime divided by batch size) as a function of the batch size, and keep increasing the batch size until we run out of memory.     
\end{itemize}

For all methods, we measure the runtime only for the registration call function, not including image loading, preprocessing and postprocessing steps, model loading, etc. 
We note that if these steps were to be included in the runtime comparison, the efficiency gains of FireANTs would be even more significant.

These results in \cref{fig:deeplearning-efficiency}
show that FireANTs is upto $10\times$ more memory efficient than SOTA deep learning methods, while performing faster than most of them at inference.
This is a result that challenges the common belief that deep learning methods are faster than iterative optimization methods at inference, similar to results shown in Hering \textit{et al.} \cite{hering2022learn2reg}.
Deep learning feedforward inference can be slow due to convolutions of large activations, skip connections requiring repeated memory accesses. 
In contrast, the iterative optimization updates in FireANTs are lightweight and can be run for more iterations. %
Since FireANTs does not generate any feature activations beyond that in the loss function, it is very memory efficient. 
Over three brain datasets, FireANTs achieves the best accuracy-runtime performance, showing that a tradeoff is not necessary for good performance.
Amortized inference can further improve efficiency - on the OASIS dataset, FireANTs can register a batch of 32 image pairs in less than 0.25 seconds per pair, .
This unprecedented efficiency paves the way for rapid prototyping, hyperparameter tuning, and scaling to high resolution datasets.
This sets a new standard for high-throughput image registration on GPUs. %

\begin{figure}[p]
    \centering
    \begin{minipage}{0.7\linewidth}
        \centering
        \subcaption{Grid search on LPBA40 dataset}
        \label{fig:lpba-gridsearch}
        \includegraphics[width=\linewidth]{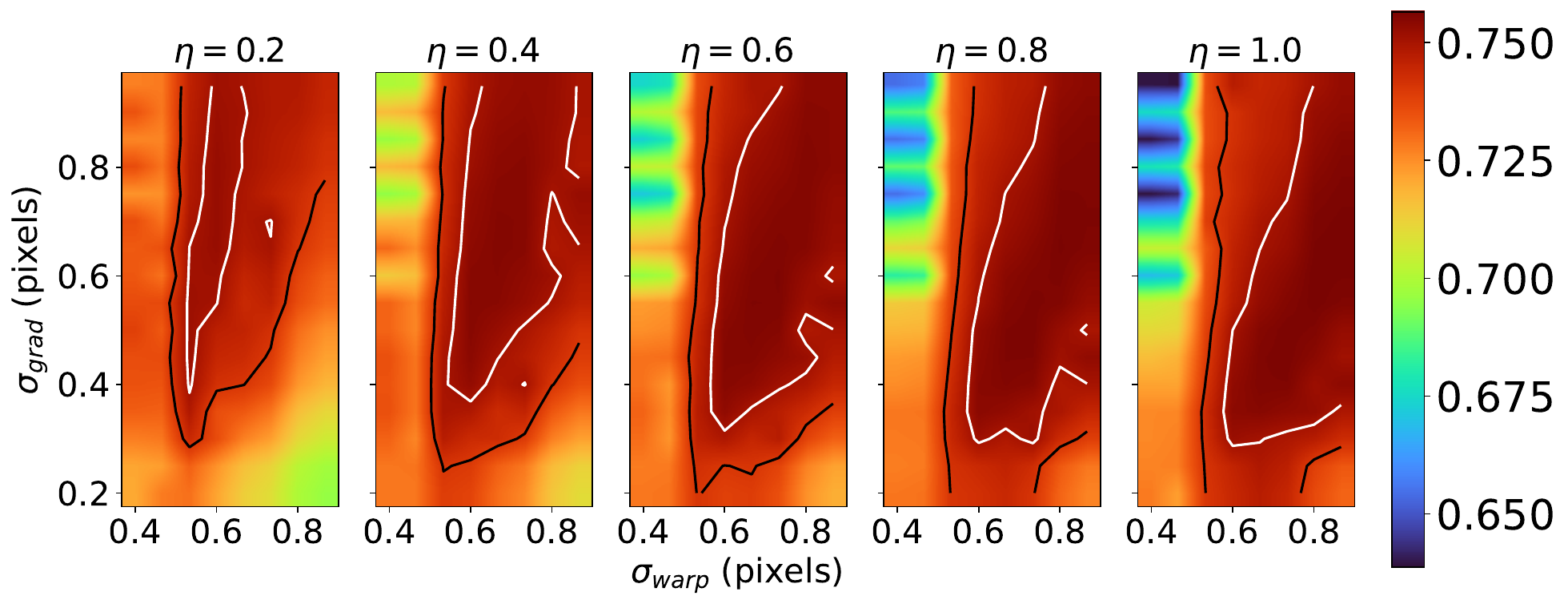}
    \end{minipage}
    \begin{minipage}{0.28\linewidth}
       \centering 
       \subcaption{Hyperparameter search efficiency on LPBA40 dataset}
       \label{fig:hyperparam-efficiency}
       \def\arraystretch{1.2}
       \resizebox{\linewidth}{!}{
       \begin{tabular}{lrr}
        \toprule
        \textbf{Method} & \textbf{GPU Hours} \\
        \hline
        ANTs & 7884.00 \\
        HyperMorph\cite{hypermorph} &  704.67  \\  %
        FireANTs & 161.60 \\
        \bottomrule
       \end{tabular}
       }
    \end{minipage}
    \begin{minipage}{0.53\linewidth}
        \centering
        \subcaption{Grid search on the EMPIRE10 dataset, with Dice score}
        \label{fig:empire-gridsearch}
        \includegraphics[width=\linewidth]{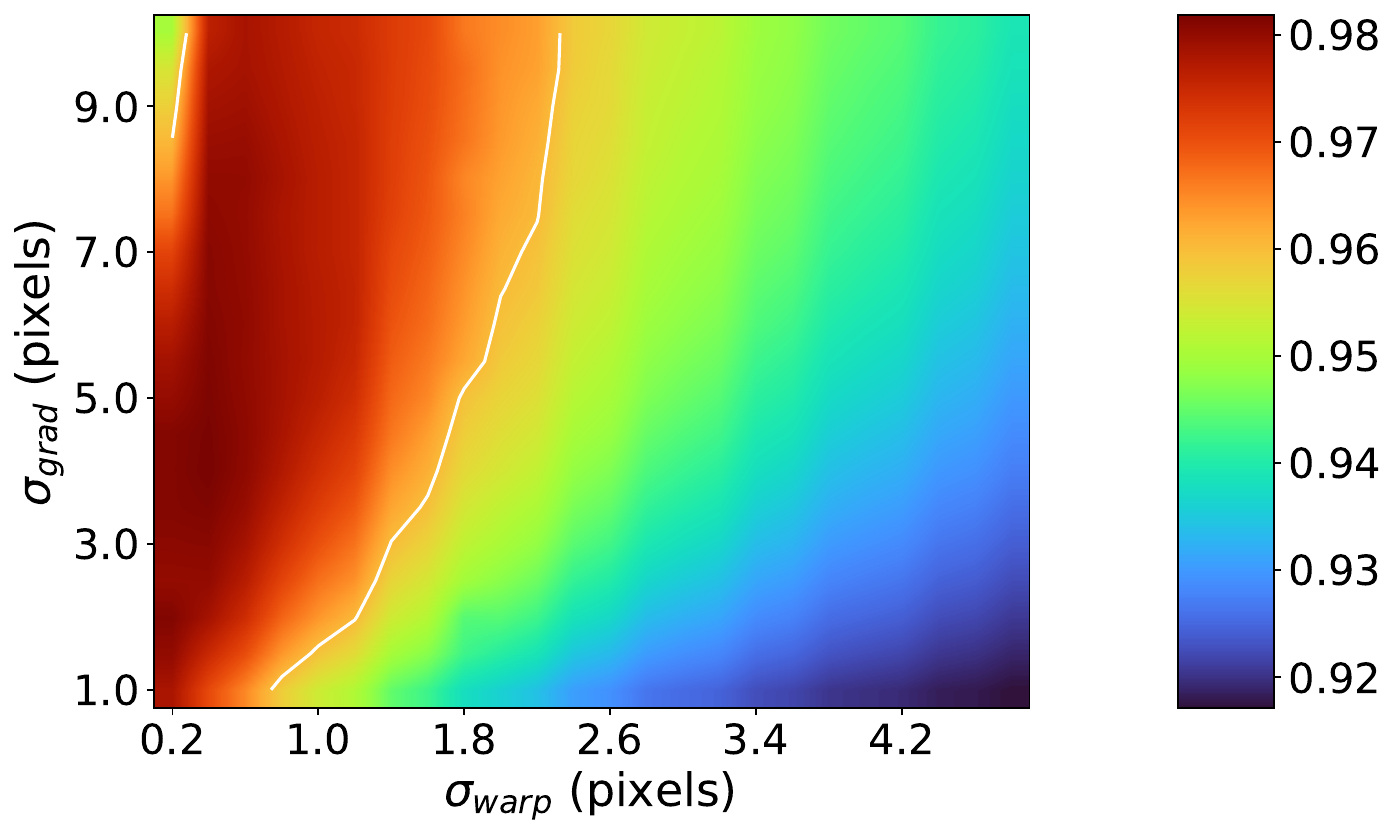}
    \end{minipage}
    \begin{minipage}{0.45\linewidth}
        \centering
        \vspace*{-25pt}
        \subcaption{Ablation on using Jacobian-free optimization on OASIS, NLST, and AbdomenMRCT-L2R datasets, spanning three organ types and modalities.}
        \label{fig:jac-ablation}
        \begin{subtable}[t]{\linewidth}
            \centering
            \resizebox{\linewidth}{!}{
            \def\arraystretch{1.2}
            \begin{tabular}{lrr}
                \toprule
                \textbf{OASIS} & \textbf{Dice Score $\uparrow$} & \textbf{Runtime (s) $\downarrow$} \\
                \hline
                \textbf{no Jacobian} & $0.783 \pm 0.032$ & $0.77 \pm 0.06$ \\
                \textbf{with Jacobian} & $0.784 \pm 0.032$ & $1.44 \pm 0.04$ \\
                \hline
                \textbf{NLST} & \textbf{TRE (mm) $\downarrow$} & \textbf{Runtime (s) $\downarrow$} \\
                \hline
                \textbf{no Jacobian} & $1.181 \pm 0.220$ & $1.80 \pm 0.12$ \\
                \textbf{with Jacobian} & $1.160 \pm 0.206$ & $4.95 \pm 0.49$ \\
                \hline
                \textbf{AbdomenMRCT-L2R} & \textbf{Dice Score $\uparrow$} & \textbf{Runtime (s) $\downarrow$} \\
                \hline
                \textbf{no Jacobian} & $0.762 \pm 0.196$ & $3.03 \pm 0.08$ \\
                \textbf{with Jacobian} & $0.764 \pm 0.195$ & $3.87 \pm 0.13$ \\
                \bottomrule
            \end{tabular}
            }
        \end{subtable}\hfill
    \end{minipage}

    \caption{\footnotesize \textbf{FireANTs facilitates feasibilility of extensive hyperparameter search in registration} The speed of FireANTs makes hyperparameter studies like these feasible, which ANTs would take years to complete.
    \textbf{(a)}: We perform a hyperparameter grid search on three hyperparameters of interest - smoothing kernel for the warp field ($\sigma_{\text{warp}}$) in pixels, smoothing kernel for the gradient of warp field ($\sigma_{\text{grad}}$) in pixels and learning rate $\eta$.
    The metric to optimize in this case is the target overlap.
    For the LPBA40 dataset, we perform a hyperparameter sweep over 640 configurations in 40 hours 
    with 8 A6000 GPUs.
    A corresponding hyperparameter sweep with 8 concurrent jobs with each job consuming 8 CPUs would take $\sim$3.6 years to complete. The white contour representing the level set for target overlap = 0.75, and the black contour representing the level set for target overlap of 0.74 show the robustness of our method to hyperparameters - performance is not brittle or sensitive to choice of hyperparameters. 
    \textbf{(b)}: Hyperparameter grid search on the EMPIRE10 dataset over $\sigma_{\text{warp}}$ and $\sigma_{\text{grad}}$ parameters (456 configurations), with a fixed learning rate of $\eta = 0.25$.
    The metric to optimize is the Dice score of the provided binary lung mask.
    This sweep takes about 12.37 hours on 8 GPUs, whereas a corresponding sweep would take 296 days for ANTs and 345 days for DARTEL (more in Fig.~\ref{fig:runtime}).
    The white contour corresponds to the level set for Dice score = 0.96, showing both a huge spectrum of parameters that achieve high Dice scores, and low sensitivity of the method to choice of hyperparameters.
    }
    \label{fig:hyperparam}
\end{figure}

\paragraph{FireANTs enables rapid  prototyping and hyperparameter tuning}
In optimization toolkits such as ANTs, correct choice of hyperparameters are key to high quality registration. 
Some of these hyperparameters are the window size for the similarity metric Cross-Correlation or bin size for Mutual Information. 
In our experience, the Gaussian smoothing kernel $\sigma_{\text{grad}},\sigma_{\text{warp}}$ for the gradient and the warp field are two of the most important parameters for accurate diffeomorphic registration.
The optimal values of these hyperparameters vary by image modality, intensity profile, noise and resolution.
Typically, these values are provided by some combination of expertise of domain experts and trial-and-error. 
However, non experts may not be able to adopt these parameters in different domains or novel acquisition settings.
Recently, techniques such as hyperparameter tuning have become popular, especially in deep learning \cite{hypermorph}.

In the case of registration, hyperparameter search can be performed by considering some form of label/landmark overlap measure between images in a validation set.
We demonstrate the stability and runtime efficiency of our method using two experiments : (1) Owing to the fast runtimes of our implementation, we show that hyperparameter tuning is now feasible for different datasets. 
The optimal set of hyperparameters is dependent on the dataset and image statistics, as shown in the LPBA40 and EMPIRE10 datasets;
(2) within a particular dataset, the sensitivity of our method around the optimal hyperparameters is very low, demonstrating the robustness and reliability of our method.
We choose the LPBA40 dataset among the 4 brain datasets due to its larger size (40$\times$39 = 1560 pairs).
We choose three parameters to search over : the learning rate ($\eta$), and the gaussian smoothing parameters $\sigma_{\text{warp}}, \sigma_{\text{grad}}$.
We use the Ray library (\texttt{https://docs.ray.io/}) to perform a hyperparameter tuning using grid search.
For the LPBA40 dataset, a grid search over three parameters (shown in ~\cref{fig:lpba-gridsearch}) takes about {40.4 hours} with 8 parallel jobs.
On the contrary, ANTs would require around {3.6 years} to complete the same grid search, with 8 threads allocated to each job and 8 parallel jobs. 
This makes hyperparameter search for an unknown modality computationally tractable.
A deep learning solution like HyperMorph~\cite{hypermorph} can perform amortized training over a predefined hyperparameter space, but still requires significant GPU hours for training and inference of 1560 pairs for each configuration to generate a plot like \cref{fig:lpba-gridsearch}.
\cref{fig:hyperparam-efficiency} shows the runtime and memory usage of FireANTs, ANTs, and HyperMorph on the LPBA40 dataset, showing that even a brute force grid search with FireANTs is about $4\times$ faster than state-of-the-art amortized hyperparameter learning. 

\paragraph{FireANTs is robust to a wide range of hyperparameters}
\cref{fig:lpba-gridsearch} shows a dense red region suggesting the final target overlap is not sensitive to the choice of hyperparameters.
Specifically, the maximum target overlap is 0.7565 and 58.4\% of these configurations have an average target overlap of $\ge$ 0.74.
This is demonstrated in \cref{fig:lpba-gridsearch} by the white contour line denoting the level set for target overlap = 0.75, and the black contour line denoting the level set for target overlap of 0.74.
The target overlap is quite insensitive to the learning rate ($\ge 0.4$) showing that our algorithm achieves fast convergence with a smaller learning rate.
On the EMPIRE10 dataset, we fix the learning rate and perform a similar hyperparameter search over two parameters, the Gaussian smoothing parameters $\sigma_{\text{warp}}, \sigma_{\text{grad}}$, shown in \cref{fig:empire-gridsearch}.
We use the average Dice score between the fixed and moving lung mask to choose the optimal hyperparameters.
FireANTs can perform a full grid search over 456 configurations on the EMPIRE10 dataset in 12.37 hours with 8 A6000 GPUs, while it takes SyN 10.031 days to run over a single configuration. %
Normalizing for 8 concurrent jobs and 456 configurations, it would take ANTs about 296 days, and DARTEL about 345 days.
This shows that our method and accompanying implementation can now make hyperparameter search for high-resolution 3D image registration studies feasible.

\begin{figure}[!ht]
    \centering    
    \begin{minipage}{\linewidth}
    \includegraphics[width=\linewidth]{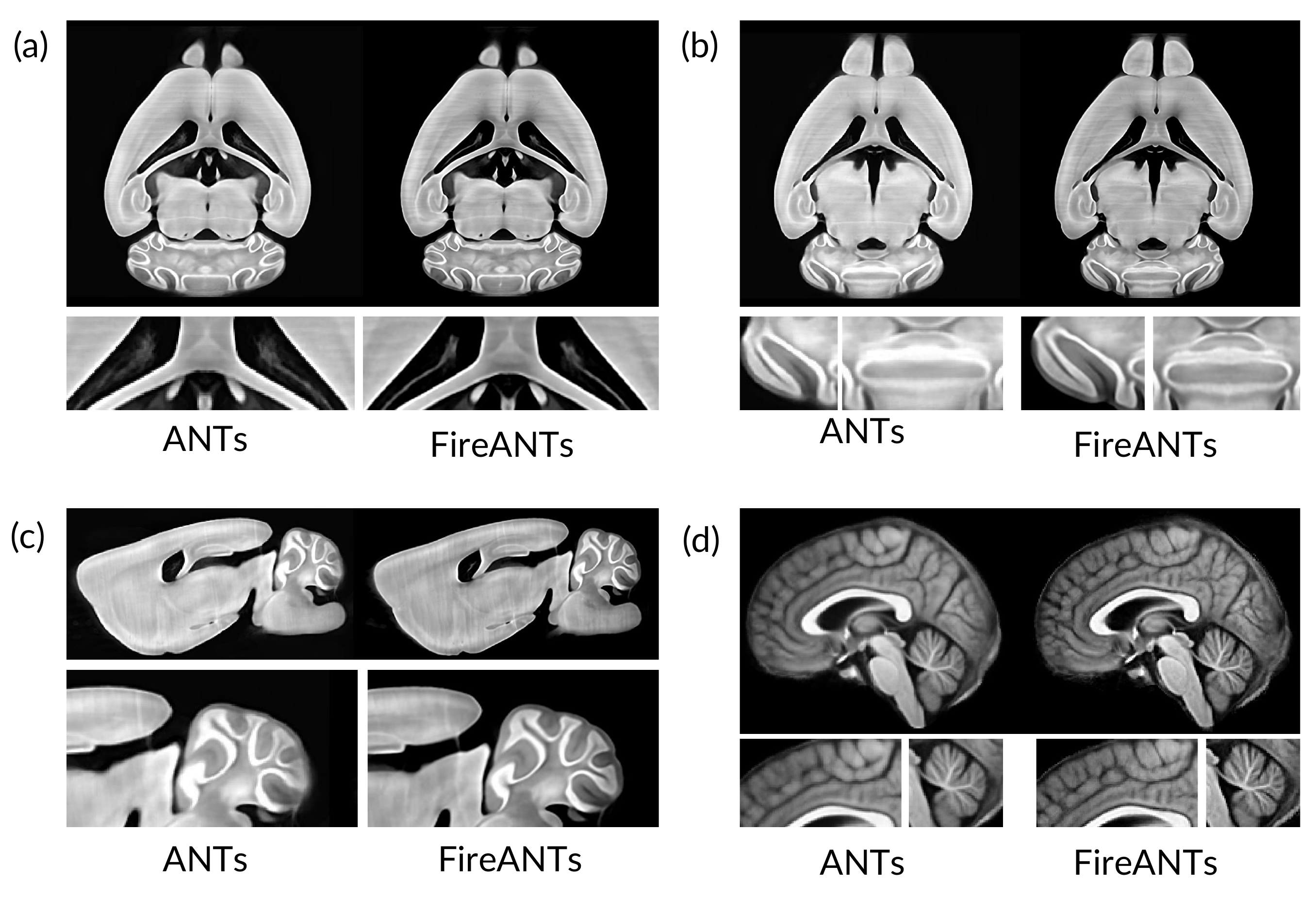}
    \end{minipage}
    \caption{ %
        Comparison of brain templates (atlases) constructed using ANTs (left) and FireANTs (right). \textbf{(a–c)} Coronal and sagittal sections of the 25{\micron} fMOST mouse brain template illustrate the improved structural fidelity of FireANTs. In the ANTs template, the internal regions of the lateral ventricles appear blurred (a), and the cerebellar architecture exhibits intensity bleeding (b, c), whereas FireANTs yields crisper delineation of these anatomical structures. \textbf{(d)} The in vivo human brain atlas further demonstrates the advantages of FireANTs, with sharper cortical folding and improved contrast and realistic intensity features in the cerebellum compared to ANTs. 
        FireANTs generates multiple high-fidelity templates while being 200-400 times faster than ANTs.
    }
    \label{fig:atlasfig}
    \label{fig:qualatlas}
\end{figure}

\paragraph{Runtime efficiency due to Jacobian-free optimization}
A key methodological approach proposed in this work is the Jacobian-free approximation of the gradient field for faster diffeomorphic optimization (\cref{sec:jac-free-eulerian}). 
A concern that may arise is the effect of this approximation on the accuracy of the registration, specifically due to the assumption that the Jacobian is positive-definite.
We ablate on the effect of this approximation on the accuracy of the registration on three datasets - OASIS, NLST, and AbdomenMRCT-L2R, spanning three organ types and modalities.
The heterogenity and variable biomechanics and deformation dynamics across multiple organ systems is a challenging testbed for measuring the effect of Jacobian-free optimization on registration accuracy.
\cref{fig:jac-ablation} shows that the Jacobian-free approximation does not significantly affect the accuracy of the registration, with a maximum Dice score difference of 0.002 and a maximum TRE difference of 0.021 mm.
However, avoiding computation of the Jacobian-augmented descent direction significantly reduces the runtime of the registration, making convergence upto 2.75$\times$ faster.
Turning on the Jacobian-augmented descent direction is easy in our implementation by only changing a flag during runtime, but we recommend turning it off for most applications.
In all our experiments except this ablation study, we do not compute the Jacobian-augmented descent direction.

\subsection{FireANTs facilitates scalable atlas generation}
Atlas generation is an important component of integrating large-scale imaging data -- including gene expression, connectivity patterns, and functional properties -— onto a common spatial coordinate system facilitating multimodal data alignment and comparison.
This requires atlas (or template) generation capabilities that scale with the unprecedented scale of acquired data.
In this section, we showcase the efficiency of atlas generation by reproducing the fMOST atlas proposed in the ANTsX ecosystem~\cite{tustison2024antsx} for the mouse brain.
We follow the steps outlined in the ANTsX Ecosystem~\cite{tustison2024antsx} for generating an fMOST atlas of the mouse brain, including preprocessing steps like downsampling to {$25$\micron} resolution, destriping, flipping along the sagittal plane for left-right symmetry, bias field correction, and affine preregistration to a common template.
Since no parcellations are available for the dataset, we qualitatively compare the atlases generated by ANTs and FireANTs, and compare their runtimes.
We also generate an in-vivo atlas for the OASIS dataset, to show scalability on smaller datasets and for quantitative evaluation.
\cref{fig:qualatlas} shows that the atlas generated by both methods is similar in terms of quality.
On a 64-thread core machine, ANTs takes \textit{141.5 hours} to generate the atlas with 6 epochs of template refinement.
With the identical number of iterations and configuration, FireANTs runs in \textit{22 minutes} with a distributed setup on an 8-GPU workstation, showing a significant improvement in runtime efficiency.
On a much lower-resolution OASIS dataset, ANTs takes \textit{2 hours and 16 minutes} to generate an atlas with 16 subjects, while FireANTs runs in \textit{32 seconds}.
To quantify atlas fidelity, we evaluate Dice Score overlap of image pairs after registering them to the atlas.
While pairwise Dice score overlap of subjects is $0.704 \pm 0.163$ with the ANTs template, the FireANTs template improves the Dice score to $0.722 \pm 0.161$.
This demonstrates that FireANTs can be used to generate high-fidelity atlases two orders of magnitude faster than ANTs at no loss in image quality, making it a powerful tool for large-scale atlas generation.

\subsection{Independent Evaluation}
Since the release of our code and documentation, FireANTs has been independently adopted by researchers in the field.
Few anectodal examples include the registration of high-resolution histology slides, and non-human primate data ~\cite{fireantsissue1,fireantsissue2}.
A compelling independent application and evaluation is performed by NextBrain~\cite{nextbrain}, a tool that utilizes FireANTs to perform Bayesian segmentation of in-vivo and ex-vivo brain MRI scans.
NextBrain uses FireANTs to register input brain MRI scans to an augmented template with a resolution of $200${\micron} and 333 regions of interest, providing a comprehensive structural analysis of the subject.
Quantitatively, incorporating FireANTs in the pipeline leads to no loss in performance as measured by Dice overlap.
The utilization of GPU-based toolkits including FireANTs reduces the runtime from 2-3 days / one week for $1mm$ in-vivo / $300${\micron} ex-vivo scans on a multi-core workstation, to \textit{less than 5 minutes} on a GPU.
Owing to the robustness and efficiency of FireANTs, it is set as the default registration method in NextBrain. %
Another independent evaluation is performed on the registration of high-resolution X-ray images~\cite{polypose}, where FireANTs establishes itself as a strong baseline, outperforming several baselines specialized for X-ray image registration.
This demonstrates the accessibility, scalability and efficiency of FireANTs in real-world applications.

\section{Discussion}
\label{sec:discussion}

We present FireANTs, a powerful and general-purpose multi-scale registration algorithm. 
Our method performs registration by generalizing the concept of first-order adaptive optimization schemes for optimizing parameters in a fixed Euclidean space, to multi-scale \textit{diffeomorphisms}. %
This generalization is highly non-trivial because diffeomorphisms are typically implemented as an image grid proportional to the size of the fixed image, and are optimized in a multi-scale manner to capture large deformations~\cite{gee1998elastic,ants,ashburner2007fast} leading to changing grid size throughout optimization.
Our method also avoids computationally expensive parallel transport and riemannian metric tensor computation steps for diffeomorphisms by solving an Eulerian descent that exploits the group structure to define descent directions from the identity transform.  %
FireANTs achieves consistent improvements in performance over state-of-the-art registration algorithms like ANTs, DARTEL, SynthMorph, VFA, unigradICON, and Bigstream.
This improvement is shown across 
fourteen datasets encompassing a broad spectrum of anatomical systems, contrast, image volume sizes, species, and modalities.
A key advantage of our method is that \textit{we do not tradeoff} any of accuracy, speed, or robustness for the others, thus being a powerful registration algorithm.

FireANTs generalizes to a long tail of out-of-distribution image modalities and datasets.
Although a substantial amount of research is focused on \textit{in-vivo} neuroimaging datasets and methods, a plethora of biomedical and life science application areas beyond in-vivo neuroimaging require a common coordinate frame for downstream quantitative analysis.
Building customized solutions from scratch for each application area requires significant domain expertise, resources, and time, and may only provide marginal additional value or insights on top of a general purpose registration algorithm.
On the contrary, FireANTs allows researchers to focus on the application-specific aspects of the problem by providing a powerful base framework for accurate and fast registration, that can be easily extended to the specific application domain using customized loss functions, regularization terms, preconditioning strategies, and learnable feature representations \cite{dio,anatomix}.
Beyond community standard registration \textit{in-vivo} neuroimaging~\cite{oasisdataset,klein2009evaluation} and pulmonary imaging~\cite{murphy2011evaluation} benchmarks that are widely regarded as a comprehensive evaluation of registration algorithms, FireANTs shows remarkable generalization on high-resolution microscopy imaging~\cite{rnrexm}, X-ray imaging~\cite{polypose}, non-human primate neuroimaging~\cite{primate}, sub-millimeter ultra-high field MRI scans~\cite{ultracortex}, multimodal rodent brains~\cite{wang2020allen,kleven2023waxholm}, and zebrafish~\cite{zbrain,azba}.
FireANTs achieves robust and accurate registration across a wide spectrum of anatomical systems, image scales, species, and modalities, providing a versatile foundation and modular framework that researchers can readily tailor to their specific application domains.

This study draws attention to an often overlooked source of performance variation - the choice of diffeomorphism representation (direct optimization versus exponential map), which motivated our adoption of direct optimization in FireANTs.
This improvement can be attributed to the representation - one can interpret direct optimization as integrating a set of \textit{time-dependent} velocity fields since the gradients change throughout optimization, allowing more flexibility in the space of diffeomorphisms it can represent, whereas SVF performs the integral of a \textit{time-independent} velocity field by design.
Using the SVF representation has three main drawbacks, which we mathematically formalize in~\cref{sec:limitations-stationary-velocity-fields}.
First, the scaling-and-squaring step is computationally expensive\cite{ashburner2007fast,mang2017lagrangian} - requiring 7-8 repeated warp compositions to obtain the diffeomorphism, and another 7-8 steps of compositions to compute the gradient, compared to only 1 warp composition for direct optimization.
Second, the sensitivity of the diffeomorphism depends on its magnitude, i.e. gradients with similar magnitudes will lead to larger updates in the diffeomorphism for larger velocity fields, which can lead to numerical instability for large deformations.
Third, the number of integration steps to ensure numerical diffeomorphisms depends on the Lipschitz constant of the velocity field, which requires adaptive Euler integration step sizes to ensure numerical diffeomorphisms.
Since most practitioners choose a fixed Euler integration step size regardless of the problem, choosing too few steps can lead to numerical inaccuracies while choosing too many steps can lead to high computational cost or numerical underflow in the base case.
Empirically we observe the exponential map representation (DARTEL) takes substantially longer to run than ANTs~\cref{fig:runtime}(a). %
In~\cref{fig:lungtable}, the results for shooting methods are substantially worse than for methods that optimize the transformation directly. We also observe this for the LPBA40 dataset in~\cref{fig:rgd-vs-exp}, where
the shooting method consistently underperformed over a wide range of hyperparameter choices. %
Modern deep learning methods have also moved away from SVF representations towards compositive (albeit non-diffeomorphic) updates \cite{vfa,sitreg}.

FireANTs is upto three orders of magnitude faster than ANTs on GPUs, and upto 10$\times$ memory efficient than deep learning methods while having highly competitive runtimes.
On the CPU, FireANTs is upto 7$\times$ faster than ANTs under identical computational budget, owing to an efficient implementation and faster convergence.
This efficiency allows for high-resolution registration of large microscopy datasets on a single GPU, grid search hyperparameter studies that are faster than amortized hyperparameter learning methods, amortized batched inference for large datasets, and high-resolution atlas generation in less than 25 minutes.
This unparalleled throughout will enable fast and accurate registration of high-resolution mesoscale and microscale imaging data that will play a paramount role in advancing our understanding of connectomics, neuroscience, cellular and molecular biology, genetics, pathology, among many other disciplines in the biomedical and biological sciences.
With breakthrough advances in high-resolution, high-throughput imaging techniques, it is imperative for registration algorithms to also scale with inordinate amounts of data.
In summary, FireANTs is a powerful and general-purpose multi-scale registration algorithm and sets a new state-of-the-art benchmark.
We propose to leverage the accurate, robust, and fast library to speed up registration workflows for ever-growing needs of performant and fast image registration in a spectrum of disciplines within the biomedical and biological sciences, wherein algorithms are bottlenecked by scalability.

\section{Methods}

\subsection{Preliminaries}

Given $d$-dimensional images $I: \Omega \to \reals^d$ and $I': \Omega \to \reals^d$ where the domain $\Omega$ is a compact subset of $\reals^2$ or $\reals^3$, image registration is formulated as an optimization problem to find a transformation $\varphi$ that warps $I'$ to $I$. The transformation can belong to an algebraic group, say $G$, whose elements $g \in G$ act on the image by transforming the domain as $(I \circ g) (x) = I(g(x))$ for all $x \in \Omega$. The registration problem solves for
\begin{equation}
    \varphi^* = \argmin_{\varphi \in G} L(\varphi) \doteq C(I, I' \circ \varphi) + R(\varphi)
    \label{eq:image-reg}
\end{equation}
where $C$ is a cost function, e.g., that matches the pixel intensities of the warped image with those of the fixed image, or local normalized cross-correlation or mutual information of image patches. There are many types of regularizers $R$ used in practice, e.g., total variation, elastic regularization~\cite{gee1993elastically}, enforcing the transformation to be invertible~\cite{christensen2001consistent}, or volume-preserving~\cite{haber2004numerical} using constraints on the determinant of the Jacobian of $\varphi$, etc. 
If, in addition to the pixel intensities, one also has access to label maps or different anatomical regions marked with correspondences across the two images, the cost $C$ can be modified to ensure that $\varphi$ transforms these label maps or landmarks appropriately.

\subsubsection{Properties of the considered Transformation Group}

A diffeomorphism is a smooth and invertible map with a corresponding differentiable inverse map~\cite{banyaga2013structure,leslie1967differential,younes2010shapes}.
We denote the set of all diffeomorphisms on $\Omega$ as $\text{Diff}(\Omega; \bR^d)$.
It is useful to note that unlike rigid or affine transforms that have a fixed number of parameters, diffeomorphisms require dense and variable parameterization, typically proportional to the size of the image.
When groups of transformations on continuous domains are endowed with a differentiable structure, they are called Lie groups. 
Diffeomorphisms are also examples of Riemannian manifolds, and are amenable to Riemannian optimization (see ~\cref{sec:rgd}).

In this work, we only consider a subgroup of diffeomorphisms.
Consider the set of continuously differentiable functions $u \in C^1_0(\Omega, \mathbb{R}^d)$ such that $u, J(u) = 0$ on $\partial \Omega$, where $J(u)$ is the Jacobian of $u$, such that $[J(u)(x)]_{ij} = \pder{u(x)_i}{x_j}$.
These functions can be extended to have $u \equiv 0$ outside $\Omega$.
Then, for a small enough $\epsilon > 0$, $x + \epsilon u(x)$ is a diffeomorphism (Proof in Proposition 8.6 in ~\cite{younes2010shapes}).
Although these diffeomorphisms are close to identity, diffeomorphisms with larger deviations from the identity can be constructed by composing these `small diffeomorphisms'.
Therefore, we study the subgroup of diffeomorphisms of the form 
\begin{equation}
\phi_n = (id + \epsilon_1 u_1) \circ \ldots \circ (id + \epsilon_n u_n)
\label{eq:flowdiff}
\end{equation}
where $u_i$s are defined as before.
We denote this subgroup as $G(\Omega, \mathbb{R}^d)$.
This subgroup retains the group structure with identity element $id$, the composition operation $\circ$ induced from $\text{Diff}(\Omega, \bR^d)$, and the inverse group element: $\phi_n^{(-1)} = (id + \epsilon_n u_n)^{(-1)} \circ \ldots \circ (id + \epsilon_1 u_1)^{(-1)}$ (as each individual $id + \epsilon_n u_n$ is shown to have an inverse~\cite{younes2010shapes}).
The elements of this subgroup can be thought of as diffeomorphisms arising from time-varying continuously differentiable flows.

However, the rate of convergence of these algorithms are contingent on the severity of ill-conditioning of ~\cref{eq:image-reg}. 
In subsequent sections, we first show the extent of ill-conditioning for diffeomorphic registration which subsequently warrants adaptive optimization over this subgroup of diffeomorphisms.

\subsection{Deformable Image Registration is a severely ill-conditioned problem}

The ill-conditioned nature of image registration represents a comparatively neglected domain of inquiry within the extant literature.
Recent works in the literature ~\cite{mang2017lagrangian,mang2017semi} only speculate the ill-conditioned nature of registration but do not quantify it. 
Computing the ill-conditioning requires us to analyze the 
Hessian of the registration cost function. 
This is infeasible in general due to the high dimensionality of the problem; the full Hessian of a MRI brain registration problem requires more than 15 petabytes of memory to store.
However, we consider a typical scenario of T1-weighted 3D MRI image registration with the L2 loss~\cite{balakrishnan2019voxelmorph,avants2004geodesic,beg2005computing}: i.e. $C(I_f, I_m, \varphi) = \sum_i (I_f(\mathbf{x}_i) - I_m(\varphi(\mathbf{x}_i)))^2$.
In this case, the gradient of $C$ w.r.t. $\varphi(\mathbf{x}_i)$ is $(I_m(\varphi(\mathbf{x}_i)) - I_f(\mathbf{x}_i))\nabla I_m(\varphi(\mathbf{x}_i))$, which does not depend on $\varphi(\mathbf{x}_j), j \ne i$.
Therefore, the full Hessian is simply a block-diagonal matrix containing pixelwise Hessians $H_i = \nabla^2_{\varphi(\mathbf{x}_i)}C$ with eigenvalues $\{\lambda_i; i=\{1,2,3\}\}$.
This makes the conditioning analysis tractable.
We calculate the per-pixel condition number, defined as $\kappa_i = |\lambda_i^{\text{max}}| / |\lambda_i^{\text{min}}|$; and investigate the relationship between the fraction of foreground pixels and $\kappa_i$ across multiple spatial resolutions of the images.
The study considers three downsampling factors: 1x (original resolution), 2x, and 4x, in accordance with existing multi-scale optimization techniques.
\cref{fig:conditioning} shows that across all resolutions, more than 60\% of foreground pixels have a condition number greater than 10.
To elucidate the impact of poor conditioning on optimization, we construct a simplified example of an ill-conditioned two-dimensional convex optimization problem, detailed in \cref{sec:twodim}.
Even with $\kappa=10$, convergence slows down drastically for an ill-conditioned convex optimization problem.
This indicates severe ill-conditioning of the registration problem, strongly motivating the need for first-order adaptive optimization.

\subsection{Adaptive Optimization for Diffeomorphisms}

We provide a brief overview of the mathematical frameworks employed to optimize parameters that reside on Riemannian manifolds like diffeomorphisms, followed by a novel algorithm that exploits the group action to define a gradient descent algorithm that eliminates computationally expensive steps.
This novel formulation of the `gradient descent' algorithm can then be formulated to incorporate adaptive algorithms such as Adam~\cite{kingma2014adam} to optimize diffeomorphisms.

\subsubsection{Euclidean gradient descent using the Lie algebra in shooting methods}
Each Lie group has a corresponding Lie algebra $\mathfrak g$ which is the tangent space at identity. This creates a locally one-to-one correspondence between elements of the group $g \in G$ and elements of its Lie algebra $v \in \mathfrak g$ given by the exponential map $\exp: \mathfrak g \to G;$
effectively to reach $g = \exp(v) \text{id}$ from identity $\text{id} \in G$, the exponential map dictates that the group element has to move along $v$ for unit time along the manifold. Exponential maps for many groups can be computed analytically, e.g., Rodrigues transformation for rotations, Jordan-Chevalley decomposition~\cite{chevalley1951theorie}, or the Cayley Hamilton theorem~\cite{mertzios1986generalized} for matrices.
For diffeomorphisms, the Lie algebra is the space of all smooth velocity fields $v: \Omega \to \reals^d$. There exist iterative methods to approximate the exponential map called the scaling-and-squaring approach~\cite{ashburner2007fast,balakrishnan2019voxelmorph} which uses the identity
\begin{align}
    \varphi = \exp(v) = \lim_{N \to \infty} \rbr{\text{id} + \frac{v}{N}}^N \label{eq:exp}
\end{align}
to define a recursion by choosing $N$ to be a large power of 2, i.e. $N = 2^M$ as
\begin{align}
    \varphi^{(1/2^M)} &= id + v/2^M \label{eq:ssbasecase} \\
    \varphi^{(1/2^k)} &= \varphi^{(1/2^{(k+1)})} \circ \varphi^{(1/2^{(k+1)})} \quad \quad \forall k \in \{0,1\ldots, M-1\}; \label{eq:scalingandsquaring}
\end{align}
This can be thought of as a special case of ~\cref{eq:flowdiff} with $n = 2^M, \epsilon = \frac{1}{n}$ and $u_1 = \ldots = u_n = v$.

By virtue of the exponential map, we can solve the registration problem of finding $\varphi \in G$ by directly optimizing over the Lie algebra $v$. This is because the Lie algebra is a vector space and we can perform, for example, standard Euclidean gradient descent for registration~\cite{moler2003nineteen,hall2013lie,hall2000elementary}. Such methods are called stationary velocity field or shooting methods.
At each iteration, one uses the exponential map to get the transformation $\varphi$ from the velocity field $v$, computes the gradient of the registration objective with respect to $\varphi$, pulls back this gradient into the tangent space where $v$ lies
\[
    \nabla_v L = \pder{\varphi}{v} \nabla_\varphi L
\]
and finally makes an update to $v$. Traditional methods like DARTEL~\cite{ashburner2007fast} implement this approach. This is also very commonly used by deep learning methods for registration~\cite{balakrishnan2019voxelmorph,krebs2019learning,Niethammer_2019_CVPR} due to its simplicity. 
Geodesic shooting methods are more sophisticated implementations of this approach where $\varphi$ is the solution of a time-dependent velocity that follows the geodesic equation; the geodesic is completely determined by the initial velocity $v_0 \in \mathfrak g$.

Adaptive optimization algorithms can be applied to the Lie algebra since it a Euclidean vector space $\mathfrak{g}$.
However, there are a number of challenges with this method.
First, this method requires computing the exponential map and its derivative, both of which need to be iteratively evaluated at each step of gradient descent.
This is evident in ~\cref{fig:lungtable} where direct optimization with ANTs runs much faster than the Lie-algebra counterpart.
Moreover, the exponential map is only \textit{locally} diffeomorphic, meaning it is suitable for modelling deviations close to the identity but not for large deformations -- this leads to less expressivity and poor performance.
In~\cref{fig:lungtable}, the greedy SyN method which employs direct optimization significantly outperforms the Lie algebra-based DARTEL. 
In~\cref{fig:rgd-vs-exp} we observed that across a large variety of hyper-parameters evaluated via grid search, direct optimization consistently led to better target overlap compared to its Lie algebra counterpart on the LPBA40 dataset.
Therefore, we do not consider this method in our work.

\subsubsection{Riemannian gradient descent}
\label{sec:rgd}
Solving the registration problem directly on the space of diffeomorphisms avoids repeated computations to and fro via the exponential map. The downside however is that one now has to explicitly account for the curvature and tangent spaces of the manifold. The updates for Riemannian gradient descent~\cite{manopt} at the $t^{\text{th}}$ iteration are
\beq{
    \aed{
        \varphi_{t+1} &= \exp_{\varphi_t} \rbr{- \eta\ \text{Proj}_{\varphi_t}(\nabla_\varphi L)}\\
        \text{where } \nabla_\varphi L &= \mathbf{g}^{-1}_{\varphi_t} \dpp{L}{\varphi},
    }
    \label{eq:rgd}
}
where one pulls back the Euclidean gradient $\dpp{L}{\varphi}$ onto the manifold using the inverse metric tensor $\mathbf{g}$ (which makes the gradient invariant to the parameterization of the manifold of diffeomorphisms) before projecting it to the tangent space using $\text{Proj}_{\varphi_t}$. Since the tangent space is a local first-order approximation of the manifold's surface, we can move along this descent direction by a step-size $\eta$ and compute the updated diffeomorphism $\varphi_{t+1}$, represented as the exponential map from $\varphi_t$ computed in the direction of $-\text{Proj}_{\varphi_t}(\nabla_\varphi L)$.

However, there are a few challenges in optimizing diffeomorphisms using Riemannian gradient descent.
First, adaptive optimization algorithms such as RMSProp~\cite{rmsprop}, Adagrad~\cite{duchi2011adaptive} and Adam~\cite{kingma2014adam} have become popular because they can handle poorly conditioned optimization problems in deep learning. 
Variants for optimization on low-dimensional Riemannian manifold exist~\cite{bonnabel2013stochastic,zhang2016riemannian,becigneul2018riemannian,geoopt2020kochurov}. 
In contrast to these manifolds, diffeomorphisms are a high-dimensional variable-sized group (e.g., the parameterization of the warp field scales with that of the image size). 
Therefore, operations like computing the Riemannian metric tensor, and parallel transport of the optimization state variables (momentum and curvature) are very computationally expensive.
For diffeomorphisms, computing the parallel transport requires solving a system of partial differential equations, which is computationally expensive.
For these reasons, we do not consider direct Riemannian optimization for diffeomorphisms in our work.

\subsection{Exploiting the group structure of diffeomorphisms}
\label{sec:diffgroup}

Diffeomorphisms are imbued with additional structure compared to a Riemannian manifold -- they are a Lie group as well.
Not all Riemannian manifolds are Lie groups - notable examples of non-Lie group Riemannanian manifolds include the sphere $\mathbb{S}^n$, fixed-rank matrices, and the Stiefel and Oblique manifolds~\cite{manopt}. 
The additional Lie group structure of $G(\Omega, \bR^d)$ allows us to exploit the group action to define a gradient descent algorithm that eliminates computationally expensive steps. 
In the following text, we provide a novel method to compute a descent direction in the group of diffeomorphisms that is computationally efficient and can be used with adaptive optimization algorithms.

\paragraph{Minimizing the Eulerian differential.} 
Consider a function $U: G \ra \bR$ that we aim to minimize.
Let $V$ be an admissible Hilbert space of vector fields on $\Omega$ embedded in $C^1_0(\Omega, \bR^d)$.
We define an \textit{Eulerian differential} in $V$ if there exists a linear form $\partial\bar{U} \in V^*$ such that for all $v \in V$:
\begin{equation}
    \left(\bar{\partial}{U}(\varphi)| v\right)_E = \partial_t U(\varphi \circ \varphi^v_{0t}) \Big|_{t=0} \label{eq:eulerian-differential}
\end{equation}
and $\varphi^v_{0t}(x) = \exp_{id}(tv)(x) = x + \int_{0}^{t} v(\varphi^v_{0s}(x)) ds$ is the flow of the vector field $v$ starting from the identity.
This definition of Eulerian differential is different from the one in ~\cite{younes2010shapes} to perform all updates ($v$) in the tangent space at identity and leverage Jacobian-free descent (see later).
The goal is to choose a suitable $v$ such that the directional change of the Eulerian differential along $v$ is negative, making $v$ a descent direction.
A more familiar rate of change of $U$ along a curve $v$ is given by the \textit{Gateaux derivative}: 
\begin{equation}
    \left(\pder{U}{\varphi} \Bigg| v\right)_G = \partial_t U(\varphi + tv) \Big|_{t=0}
\end{equation}
The Eulerian differential is closely related to the Gateaux derivative of $U$ at $\varphi$ as:
\[
    \left(\bar\partial{U}(\varphi)| v\right)_E = \left(\pder{U}{\varphi} \Bigg| J(\varphi) v\right)_G
\]
using chain rule. The right side is further expanded as:
\[
    \left(\bar\partial{U}(\varphi)| v\right)_E = \int_{\Omega} \left(\pder{U}{\varphi}(\varphi)(x)\right)^\top J(\varphi(x)) v(x) dx
\]
where $J(\varphi)(x) = J(\varphi(x))$ with slight abuse of notation. 
We introduce the Gateaux derivative and relate it to the Eulerian derivative because we typically have access to the Gateaux derivative using automatic differentiation tools like PyTorch, but to perform optimization on the group of diffeomorphisms, we need to compute the Eulerian differential.
Choosing $$v_d(x) = -J(\varphi(x))^\top\ \pder{U}{\varphi}(\varphi)(x)$$ gives us:
\[
    \left(\bar\partial{U}(\varphi)| v_d\right)_E = -\int_{\Omega} \Bigg\|J(\varphi)^\top \pder{U}{\varphi}(\varphi)(x)\Bigg\|^2 dx < 0
\]
This choice of $v_d(x)$ is therefore a descent direction for the Eulerian differential of $U$ at $\varphi$.
To perform gradient descent on the Eulerian differential at $\varphi$, we need to compute the descent direction $v_d$, perform the exponential map with a small learning rate $\eta_t$, and perform the update:
\[
    \varphi_{t+1} = \varphi_t \circ \exp_{id}(\eta_t v_d)
\]
For small enough $\eta_t$, the exponential map can be approximated with a retraction map (i.e. $\exp_{id}(\eta_t v_d) \approx id + \eta_t v_d$), which is quick to compute. \\

We quickly contextualize the key differences between Gateaux gradient descent and our proposed Eulerian descent. 
First, the steepest descent direction in Gateaux gradient descent is $-\pder{U}{\varphi}(\varphi)$, whereas it is $-J(\varphi)^\top \pder{U}{\varphi}(\varphi)$ in Eulerian descent.
Second, the update rule in Gateaux gradient descent is $\varphi_{t+1} = \varphi_t - \eta_t \pder{U}{\varphi}(\varphi)$, whereas it is $\varphi_{t+1} = \varphi_t \circ \exp_{id}(\eta_t v_d)$ in Eulerian descent.
These two differences capture the essence of performing optimization on the group of diffeomorphisms in contrast to optimizing on the (Euclidean) ambient space directly.

\paragraph{Adaptive optimization on diffeomorphisms.}
Note that for small enough $t$, the descent direction $v_d(x)$ can also be interpreted as a vector in the tangent space at identity, with $\varphi_{0t}^v = \exp_{id}(tv)$ since $\varphi_{00}^{v} = id$, and $\partial_t \varphi_{0t}^v |_{t=0} = v$.
Descent directions over gradient descent iterations $i$ denoted as $v_{d}^{(i)}$ all lie on the same vector space, i.e. the tangent space at identity.
Therefore, first order algorithms like Adam can be applied on the sequence of descent directions $v_{d}^{(i)}$ which now lie in the same vector space, without requiring computing the metric tensor, parallel transport or change of coordinates (charts) throughout the optimization process. 
This framework leveraging the group structure forms the core of our adaptive optimization algorithm for diffeomorphisms.
Our framework is therefore a significant advantage over Riemannian optimization methods which require parallel transport of the momentum and curvature vectors at each iteration.

\subsubsection{Jacobian-Free Eulerian Descent}
\label{sec:jac-free-eulerian}

We provided an obvious choice of descent direction $v_d(x)$ for the Eulerian differential of $U$ at $\varphi$.
The Gateaux derivative $\pder{U}{\varphi}$ is readily obtained using automatic differentiation tools like PyTorch.
However, the descent direction requires us to multiply this derivative with the Jacobian of the diffeomorphism $J(\varphi)$, which may be computationally expensive.
However, in most diffeomorphic image registration applications, the role of the diffeomorphism is warp the image by performing local translations, scaling and shearing without introducing large local rotations.
Mathematically, we consider the polar form of the Jacobian $J(\varphi)(x) = U(x)P(x)$ where $U(x)$ is a unitary matrix, and $P(x)$ is a positive definite matrix.
We assume that for most applications, $U(x) \approx I_{d\times d}$, making $J(\varphi)(x)$ positive definite.
With this assumption, we can choose the modified descent direction 
$$v_d'(x) = -\pder{U}{\varphi}(\varphi)(x)$$
and the Eulerian differential at $\varphi$ is
\[
    \left(\bar\partial{U}| v_d'\right)_E = -\int_{\Omega} \left(\pder{U}{\varphi}(\varphi)(x)\right)^\top J(\varphi(x)) \pder{U}{\varphi}(\varphi)(x) dx < 0
\]
since $v_d'(x)^\top J(\varphi(x)) v_d'(x) \ge 0$ for all $x \in \Omega$, owing to the (assumed) positive definiteness of $J(\varphi)(x)$.
For all experiments, Jacobian-free descent directions $v_d'(x)$ are used, and they provide faster runtime and with same accuracy.
Adaptive first-order optimization can now be performed on this modified sequence on descent directions $v_d'^{(i)}(x)$, saving significant computational and memory overhead by avoiding computation of $J(\varphi)$. \\

Note that this algorithm using the Eulerian differential is only possible due to the group structure of diffeomorphisms.
For an arbitrary Riemannanian manifold $\mathcal M$ and points $\varphi, \varphi_{0t}^v \in \mathcal{M}$, the operation $\varphi \circ \varphi_{0t}^v$ does not make sense.
The additional group structure of $G(\Omega, \bR^d)$ allows us to propose a novel Eulerian descent algorithm without performing Lie algebra optimization, or Riemannian gradient descent, both of which are computationally expensive for diffeomorphisms.

\subsubsection{Alternative formulations for Eulerian differential}
Our definition of \cref{eq:eulerian-differential} is different from the one in ~\cite{younes2010shapes} in two subtle but important ways.
First, we do not define the registration objective in terms of the group action or pullback image $\varphi.I = I(\varphi^{-1}(x))$, and instead define the objective in terms of the pushforward image $I(\varphi(x))$. 
This is to avoid computing and storing both $\varphi$ for autodifferentiation and $\varphi^{-1}$ for computing the objective, implementational simplicity, and consistency with more modern registration framework formulation.
FireANTs provides additional functionality to compute $\varphi^{-1}$ \textit{post hoc} using a multi-scale objective function similar to the image matching objective: $\varphi^{-1} = \argmin_{\psi \in G} \sum_{x \in \Omega} \|\psi(\varphi(x)) - x\|_2^2 + \|\varphi(\psi(x)) - x \|_2^2$.
This allows researchers to obtain an inverse transform on a \textit{post hoc} basis without computing $\varphi^{-1}$ \textit{during} optimization.
This subroutine is also used in the symmetric registration objective to compute the final transformation $\varphi = \varphi_M \circ \varphi_F^{-1}$.
The second difference is the definition of the Eulerian differential itself - note that we use the composition $U(\varphi \circ \varphi^v_{0t})$ in \cref{eq:eulerian-differential} instead of $U(\varphi^v_{0t} \circ \varphi)$.
Defining the Eulerian differential using the second formulation without using the group action $\varphi.I$ implies that we will compute the velocity field in the Langrangian frame, i.e. $V(y) = v(\varphi(x)) = -\frac{\partial U}{\partial \varphi}$.
To compute adaptive optimization updates and Lipschitz constant to scale the learning rate (see \cref{sec:limitations-stationary-velocity-fields}), we need to compute the corresponding velocity field in the Eulerian frame, i.e. $v(x) = V(\varphi^{-1}(y))$ which requires computing $\varphi^{-1}$.
Therefore, we choose a definition that avoids computing $\varphi^{-1}$ and allows inexpensive adaptive optimization updates.

\subsection{Interpolation strategies for multi-scale registration}
Classical approaches to deformable image registration is performed in a multi-scale manner.
Specifically, an image pyramid is constructed from the fixed and moving images by downsampling them at different scales, usually in increasing powers of two.
Optimization is performed at the coarsest scale first, and the resulting transformation at each level is used to initialize the optimization at the next finer scale.
Specifically, for the fixed image $I$ and the moving image $I'$ and $K$ levels, let the downsampled versions be $\{ I_k \}_{k=1}^{K}$ and $\{ I'_k \}_{k=1}^{K}$, where $k$ is the scale index from coarsest to finest.
At the $k$-th scale, the transformation $\varphi_k$ is optimized as
\[
    \varphi_k^* = \argmin_{\varphi_k \in G} L(I_k, I'_k \circ \varphi_k)
\]
where $\varphi_k$ is initialized as 
\[
    \varphi_k = \begin{cases}
        {id} & \text{if } k = 1\\
        \text{Upsample}(\varphi_{k-1}) & \text{otherwise}
    \end{cases}
\]
Unlike existing gradient descent based approaches, our Riemannian adaptive optimizer also contains state variables $m_k$ corresponding to the momentum and $\nu_k$ corresponding to the EMA of squared gradient, at the same scale as $\varphi_k$, which require upsampling as well.

Unlike upsampling images, upsampling warp fields and their corresponding optimizer state variables requires careful consideration of the interpolation strategy.
Bicubic interpolation is a commonly used strategy for upsampling images to preserve smoothness and avoid aliasing.
However, bicubic interpolation of the warp field can lead to overshooting, leading to introducing singularities in the upsampled displacement field when there existed none in the original displacement field.
In contrast, bilinear or trilinear interpolation does not lead to overshooting, and therefore diffeomorphism of the upsampled displacement is guaranteed, if the original displacement is diffeomorphic.
We demonstrate this using a simple 2D warp field in ~\cref{fig:rie-opt-tricks}(b).
On the left, we consider a warp field created by nonlinear shear forces.
This warp field does not contain any tears or folds - and is diffeomorphic.
We upsample this warp field using bicubic interpolation (top) and bilinear interpolation (bottom).
We also plot a heatmap of the negative of the determinant of the Jacobian of the upsampled warp, with a white contour representing the zero level set.
Qualitatively, bicubic interpolation introduces noticable folds in the warping field, leading to non-diffeomorphisms in the upsampled warp field.
The heatmap shows a significant portion of the upsampled warp field has a negative determinant, indicating non-invertibility.
On the other hand, bilinear interpolation looks blocky but preserves diffeomorphism everywhere, as also quantitatively verified by the absence of a zero level set in the heatmap. 

The complete algorithm is described in ~\cref{alg:fireants}.
\begin{figure}[H]
    \begin{minipage}{0.99\linewidth}
        \begin{algorithm}[H]
            \caption{FireANTs}
            \label{alg:fireants}
            \begin{algorithmic}[1]
                \State \textbf{Input:} Fixed image $I_f$, Moving image $I_m$ \\ Scales $[s_1, s_2, \ldots, s_n]$, Iterations $[T_1, T_2, \ldots T_n]$, $n$ scales \\ \texttt{optstate} optimizer state (for Adam, RMSProp, etc.) \\ \texttt{use\_jac} boolean specifying whether to use Jacobian in descent direction
                \State 
                \State Initialize $\varphi \leftarrow \mathbf{id}_{s_1}$.  \Comment{Initialize warp to identity at first scale}
                \State Initialize  ${l} \leftarrow 1$.   \Comment{Initialize current scale}
                \While{${l} \le n$}
                    \State Initialize $i \leftarrow 0$
                    \State Initialize $I_f^{l}, I_m^{l} \leftarrow \text{downsample}(I_f, s_{l}),  \text{downsample}(I_m, s_{l})$
                    \While{$i < T_{{l}}$}
                        \State $U_i \leftarrow C(I_f^{l}, I_m^{l} \circ \varphi^i) + R(\varphi)$
                        \State Compute $v_d'(x) \leftarrow \pder{U_i}{\varphi}(\varphi^{(i)})(x)$  \Comment{Jacobian-free Eulerian descent direction}
                        \If{\texttt{use\_jac}}
                        \State Compute $v_d'(x) \leftarrow J^\top(\varphi^{(i)}(x))v_d'(x)$ \Comment{Eulerian descent direction}
                        \EndIf
                        \State Update $(v_d'(x), \text{\texttt{optstate}}) \leftarrow \text{\texttt{optstate}}(v_d'(x)) $  \Comment{Apply and update optimizer state}
                        \State Update $\varphi^{(i+1)} \leftarrow \varphi^{(i)}\circ\exp_{id}(\epsilon_i v_d') \approx \varphi^{(i)}\circ(id + \epsilon_i v_d')$ 
                        \State $i \leftarrow i + 1$
                    \EndWhile
                    \If{$l < n$}
                        \State $\varphi \leftarrow \text{Upsample}(\varphi, s_{(l+1)})$ \Comment{Upsample warp to next scale using bilinear/trilinear interpolation}
                    \EndIf
                    \State $l \leftarrow l + 1$
                \EndWhile
            \end{algorithmic}
        \end{algorithm}
    \end{minipage}
    \caption{\textbf{Algorithm for FireANTs} \cref{alg:fireants} outlines the key steps in FireANTs - computing the Jacobian-free Eulerian descent direction which is simply the Gateaux derivative. If the boolean \texttt{use\_jac} is specified, then use the steepest Eulerian descent direction instead. This descent direction is then modified using any adaptive optimization algorithm denoted as \texttt{optstate}. The warp field is then updated using the exponential map or retraction map for small $\epsilon_i$. After optimization at a given scale, the warp field is upsampled using bilinear or trilinear interpolation to the next scale until optimization is complete for all steps.}
    \label{alg:pseudocode}
\end{figure}

\section{Data Availability}
The following datasets were used in the study, and are publicly available for free:

\begin{itemize}
    \item Klein {\etal} neuromapping challenge datasets (IBSR18, CUMC12, MGH10, LPBA40): \url{https://www.synapse.org/#!Synapse:syn3251018}
    \item OASIS, NLST, Abdomen MRCT registration datasets: \url{https://learn2reg.grand-challenge.org/Datasets/}
    \item EMPIRE10 lung dataset: \url{https://empire10.grand-challenge.org/Download/}
    \item RnR Expansion Microscopy mouse dataset: \url{https://rnr-exm.grand-challenge.org/data/}
    \item Fluorescence micro-optical sectioning tomography (fMOST) imaging for mouse brain data: https://knowledge.brain-map.org/data/K1YP17A0QIKJOMOAIS4 
    \item PRIME-DE Macaque dataset:  \url{https://fcon_1000.projects.nitrc.org/indi/indiPRIME.html}
    \item Ultracortex dataset: \url{https://openneuro.org/datasets/ds005216/versions/1.1.0/download}
    \item Waxholm Rat Brain dataset: \url{https://www.nitrc.org/frs/?group_id=1081#}
    \item Allen CCFv3 mouse brain dataset: \url{https://atlas.brain-map.org/atlas}
    \item AZBA Zebrafish dataset: \url{https://azba.wayne.edu/}
    \item ZBrains Zebrafish dataset: \url{https://zebrafishexplorer.zib.de/}
\end{itemize}

\section{Code Availability}
The code for FireANTs is publicly available at \url{https://github.com/rohitrango/fireants}.

\section{Sex and Gender reporting}
Sex and Gender was not considered in the study.

\section{Acknowledgements}
This work was supported by the National Institutes of Health (NIH) under grants RF1-MH124605, R01-HL133889, R01-EB031722, and U24-NS135568.

\section{Author Contributions}
R.J. conceptualized and designed the experiments, designed and implemented the framework, performed the experiments, analyzed and interpreted the results.
P.C. and J.C.G. provided advising and funding support.
All authors contributed to the writing, editing, and review of the manuscript, and approved the final version of the manuscript.

\section{Competing Interests}
The authors declare no competing interests.

\bibliographystyle{ieeetr}
\bibliography{references}

\clearpage

\appendix

\section{Related Work}
\label{app:relatedwork}
The field of image registration has evolved through several distinct phases, introducing progresively more expressive mathematical and computational frameworks.
Early work in computational anatomy established the foundational view of registration as the problem of mapping anatomical variability through smooth, invertible transformations between coordinate systems.
Pioneering contributions by Grenander and colleagues formalized anatomical shapes as elements of a deformable template space, leading to probabilistic formulations of anatomical variability and population analysis \cite{grenander1998computational}.
These works introduced the notion that anatomical differences could be quantified through the geometry of transformations rather than through intensity differences alone, thereby providing the mathematical basis for subsequent developments in diffeomorphic mapping.
Other works model diffeomorphisms as solutions of elastic matching \cite{gee1993elastically,gee1998elastic} and large deformation kinematics \cite{christensen1996deformable,christensen1997volumetric}.
To mitigate some of the limitations of these approaches, the Large Deformation Diffeomorphic Metric Mapping (LDDMM) framework \cite{beg2005computing} provided a rigorous variational and geometric formulation of registration. In LDDMM, deformations are modeled as flows of smooth velocity fields generating geodesics on the diffeomorphism group under a right-invariant metric.
Subsequent extensions introduced more efficient parameterizations and regularizers, including stationary velocity fields \cite{ashburner2007fast}, shooting formulations \cite{miller2006geodesic}, and Riemannian approaches to statistical analysis on the manifold of diffeomorphisms \cite{fletcher2004principal}.
A complementary approach to ensure smooth and anatomically plausible deformations is to supplement the image similarity objective with a regularization function that is applied on the displacement field.
Typical examples include minimizing the bending energy \cite{johnson2001landmark,rueckert2002nonrigid}, total variation \cite{vishnevskiy2016isotropic}, diffusion regularizer \cite{reithmeir2024learning,balakrishnan2019voxelmorph,wu2022nodeo} that aim to constrain the degrees of freedom of the deformation field.
Regularization can also be implicit - different deep network architectures induce different implicit regularizations \cite{ulyanov2018deep}, and choice of representation of the nonlinear transform (stationary velocity field, B-Spline, downsampled displacement fields) constrains the solution space, providing implicit regularization \cite{rueckert2002nonrigid,modat2012parametric}.
These advances in representation and regularization links registration to broader concepts in geometric mechanics and fluid dynamics, providing both theoretical interpretability and a pathway for optimization-based implementation.
Subsequent gradient-descent based optimization formulations \cite{avants2008symmetric,avants2004geodesic} led to the development of the Advanced Normalization Tools (ANTs)\cite{ants}, which is a state-of-the-art registration framework for medical image registration.
ANTs is used routinely across a broad range of biomedical and life sciences research workflows, and is the de-facto standard for image registration.
However, ANTs uses a simple gradient-descent based approach instead of utilizing powerful and efficient adaptive optimization algorithms, and is slow and not scalable due to a CPU-only implementation, and does not leverage GPUs to exploit the massively parallelizable nature of the problem.

More recently, the field has shifted toward model-based deep learning approaches that approximate the optimization solution with feedforward inference for inference speed.
Methods such as VoxelMorph \cite{balakrishnan2019voxelmorph} and subsequent unsupervised learning based variants \cite{lku,mok2020large,mok2021conditional} learn parametric mappings that approximate solutions to variational registration problems.
However, most methods cannot represent solutions to time-dependent velocity fields, therefore most deep learning methods rely on stationary velocity field representations to model Lie group of diffeomorphisms using its Lie algebra.
However, stationary velocity fields have a few computational and mathematical limitations, outlined in ~\cref{sec:limitations-stationary-velocity-fields}.
Another limitation of deep networks is the large activation memory overhead for inference, making them infeasible for high-resolution registration.
Other works \cite{zhang2015finite,wang2020deepflash,jia2023fourier} use approximations like bandlimited velocity fields to model efficient inference, at the cost of potentially losing some of the high-frequency details in the velocity field that is potentially useful for modelling cortical folding patterns, or large atrophy of ventricles in diseased patients.
Moreover, the memory overhead and scalability of these methods to large datasets is not well studied.
Another limitation of these methods is their drop in performance on out-of-distribution datasets \cite{magicormirage,jian2024mamba,liu2025unsupervised,hoffmann2021synthmorph,tian2024unigradicon}, that has motivated the need for domain-agnostic or foundational models \cite{hoffmann2021synthmorph,tian2024unigradicon}.
Other works \cite{vfa} also claim generalization due to architectural design choices that mimic registration specific operations.
In our experiment setup, we compare our method with these domain-agnostic methods to test the true long tail generalization capabilities of domain-agnostic registration algorithms.
We wish to preserve the full expressivity of time-depdendent velocity fields for modelling diffeomorphisms, and enable adaptive optimization on this space directly, while maintaining high computational efficiency and low memory overhead for inference.
This motivates revisiting the classical optimization-based method ANTs, and extending it to enable powerful adaptive optimization for improved accuracy and convergence while improving computational efficiency and scalability using a GPU implementation.

\section{Datasets and Evaluation Metrics}
\label{app:datasets}

We provide details about the datasets and evaluation metrics used in the paper.

\subsection{In-vivo brain MRI mapping challenges (Klein \etal and OASIS)}
\label{sec:kleindata}

\paragraph{Klein \etal neuromapping challenge}
Brain image data and their corresponding labels for 80 normal subjects were acquired from four different datasets.
The \textit{LPBA40} dataset contains 40 brain images and their labels to construct the LONI Probabilistic Brain Atlas (LPBA40).
All volumes were skull-stripped, and aligned to the MNI305 atlas ~\cite{evans19933d} using rigid-body transformation to correct for head tilt.
For all these subjects, 56 structures were manually labelled and bias-corrected using the BrainSuite software.
The \textit{IBSR18} dataset contains brain images acquired at different laboraties through the Internet Brain Segmentation Repository.
The T1-weighted images were rotated to be in Talairach alignment and bias-corrected.
Manual labelling is performed resulting in 84 labeled regions.
For the \textit{CUMC12} dataset, 12 subjects were scanned at Columbia University Medical Center on a 1.5T GE scanner.
Images were resliced, rotated, segmented and manually labeled, leading to 128 labeled regions.
Finally, the \textit{MGH10} dataset contains 10 subjects who were scanned at the MGH/MIT/HMS Athinoula A. Martinos Center using a 3T Siemens scanner.
The data is bias-corrected, affine-registered to the MNI152 template, and segmented.
Finally the images were manually labeled, leading to 74 labeled regions.
All datasets have a volume of $256\times256\times\{128,124\}$ voxels with varying amounts of anisotropic voxel spacing, ranging from $0.84\times0.84\times1.5$mm to $1\times1\times1.33$mm.
ANTs was one of the top performing methods for this challenge, performing well robustly across all four datasets.

\begin{figure}[!htpb]
    \centering
    \includegraphics[width=0.9\linewidth]{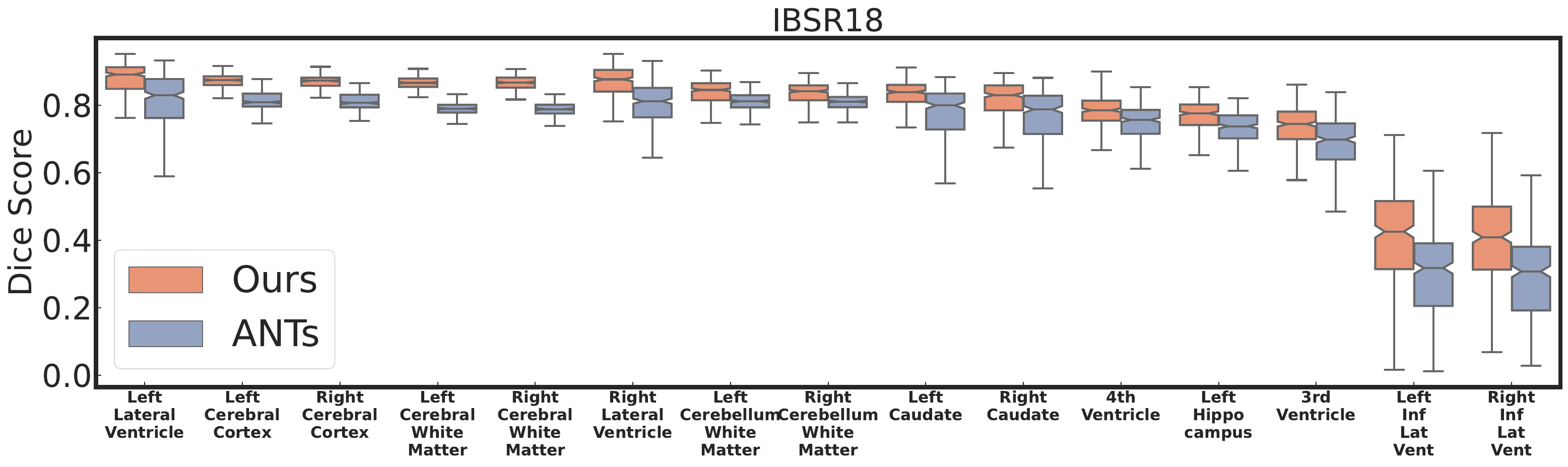}
    \includegraphics[width=0.9\linewidth]{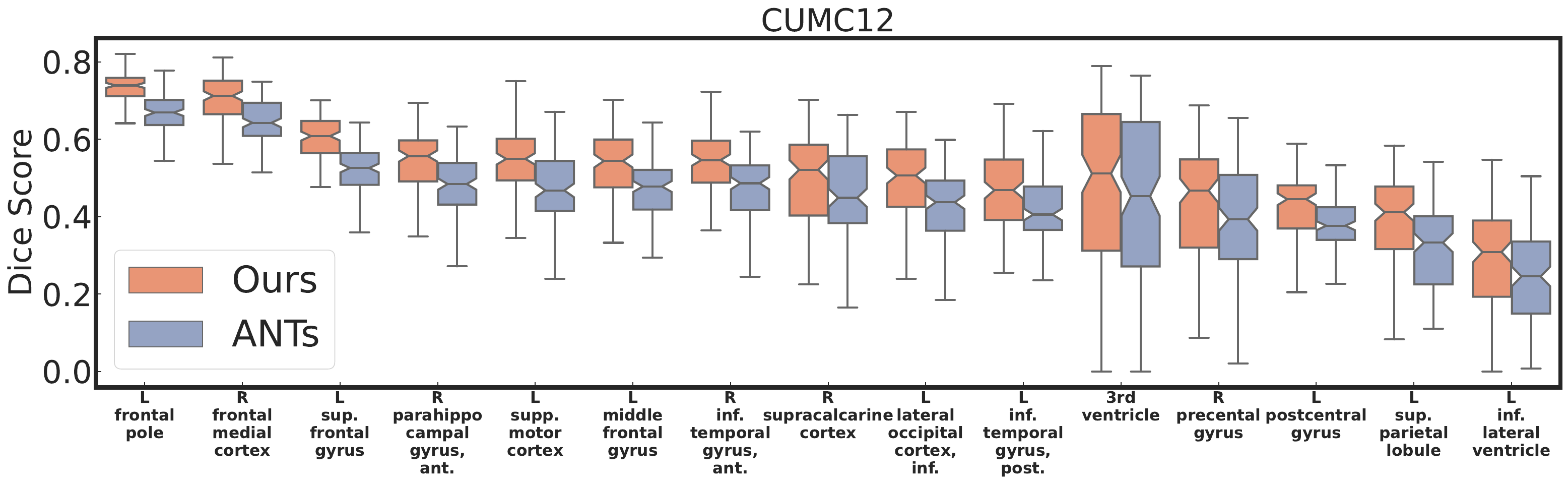}
    \includegraphics[width=0.9\linewidth]{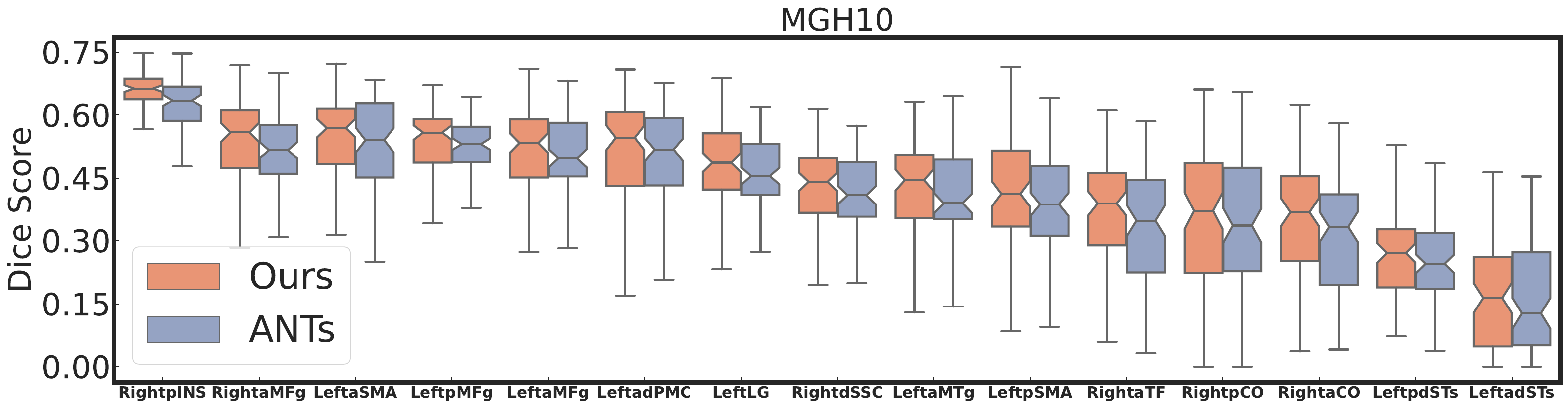}
    \includegraphics[width=0.9\linewidth]{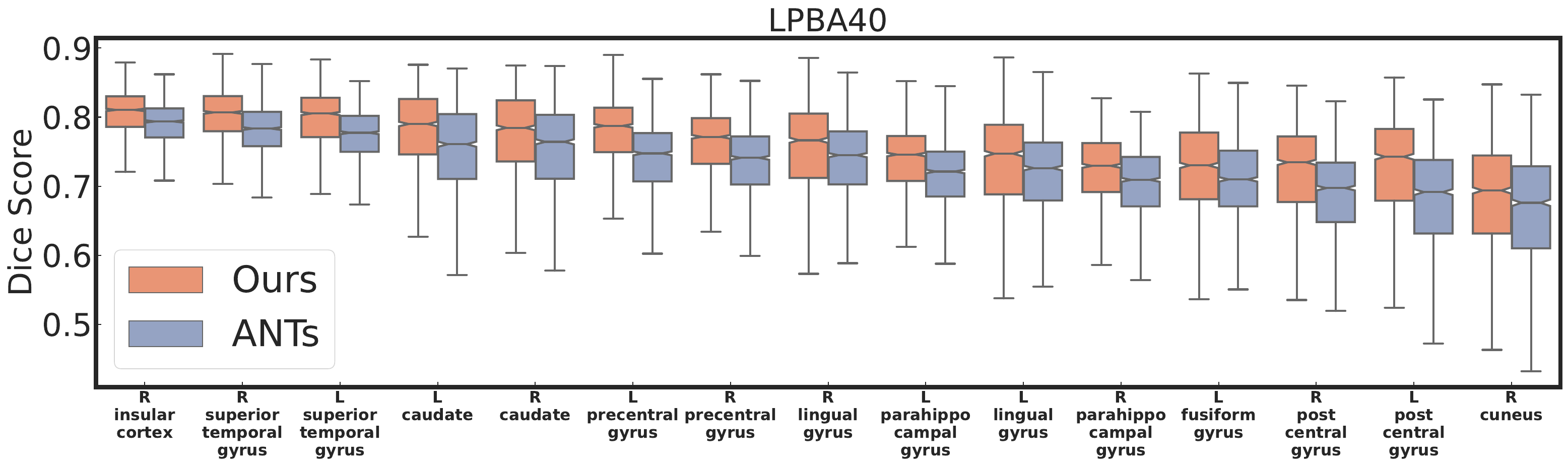}
    \caption{\textbf{Regionwise target overlap on the brain MRI datasets}: We further evaluate regionwise overlap scores by sampling 15 regions from each dataset, and comparing their distribution using our method and ANTs. Our method has a much higher median score, and better interquartile ranges across regions, demonstrating both accuracy and robustness.}
    \label{fig:regionwise-dice-brain}
\end{figure}

\begin{figure}[!htpb]
    \centering
    \includegraphics[width=0.8\linewidth]{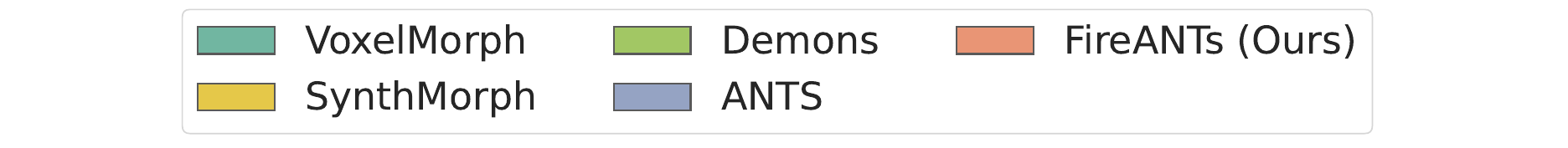}
    \braintablefig{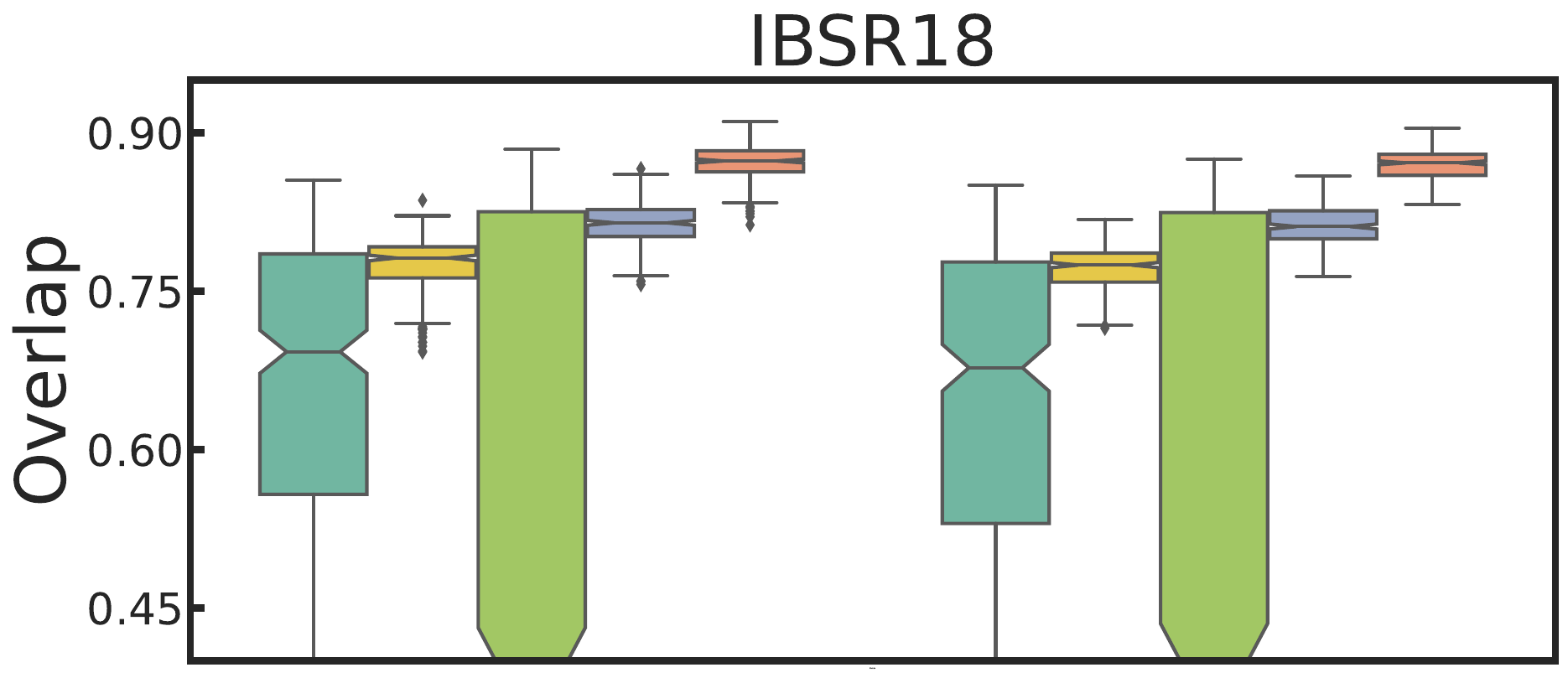}{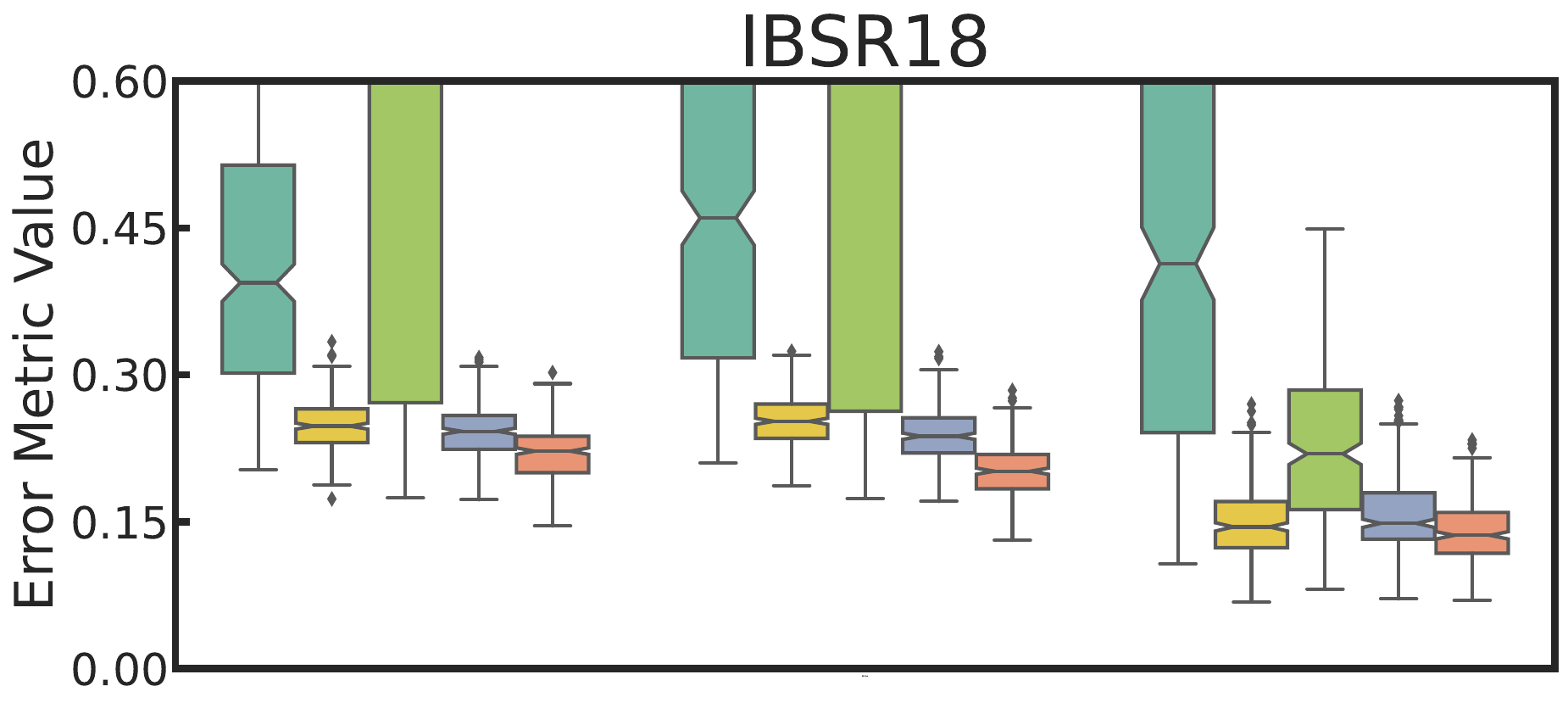}
    \braintablefig{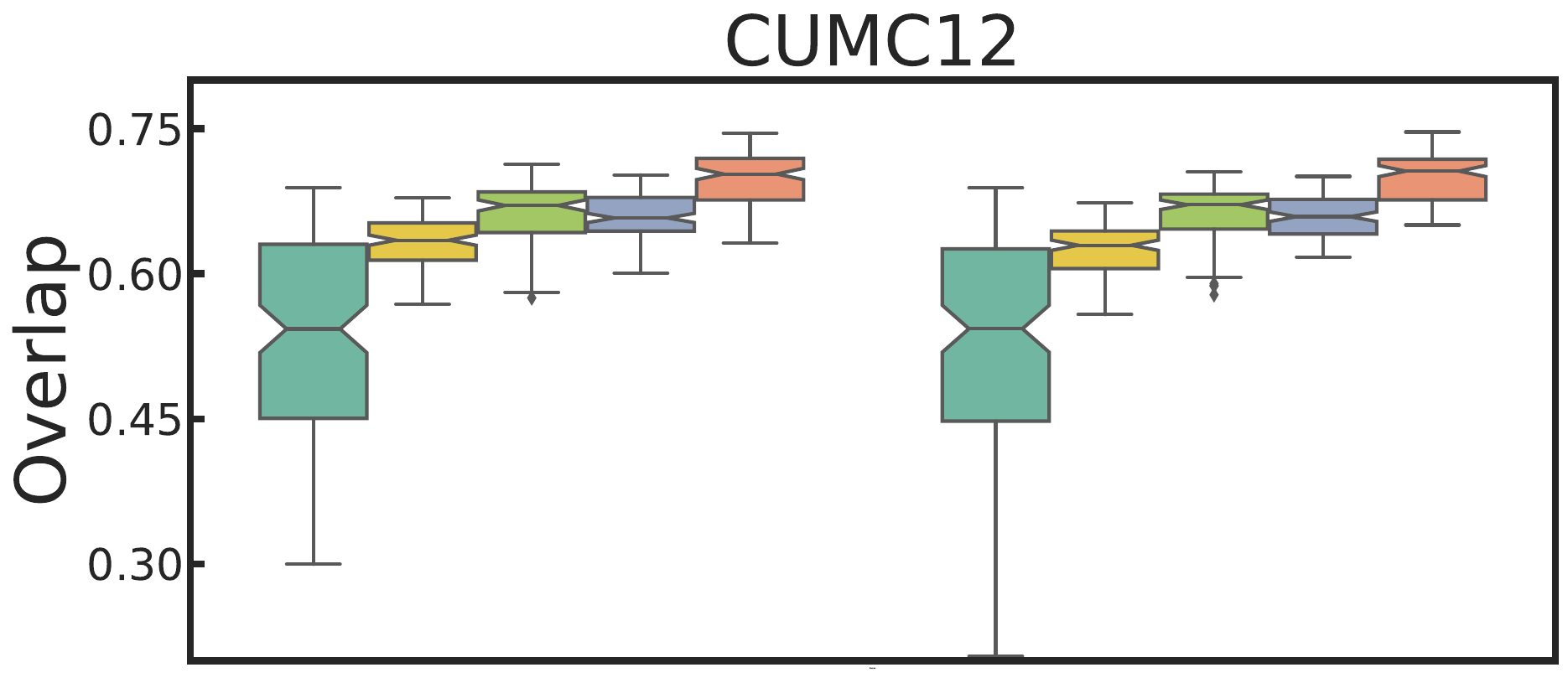}{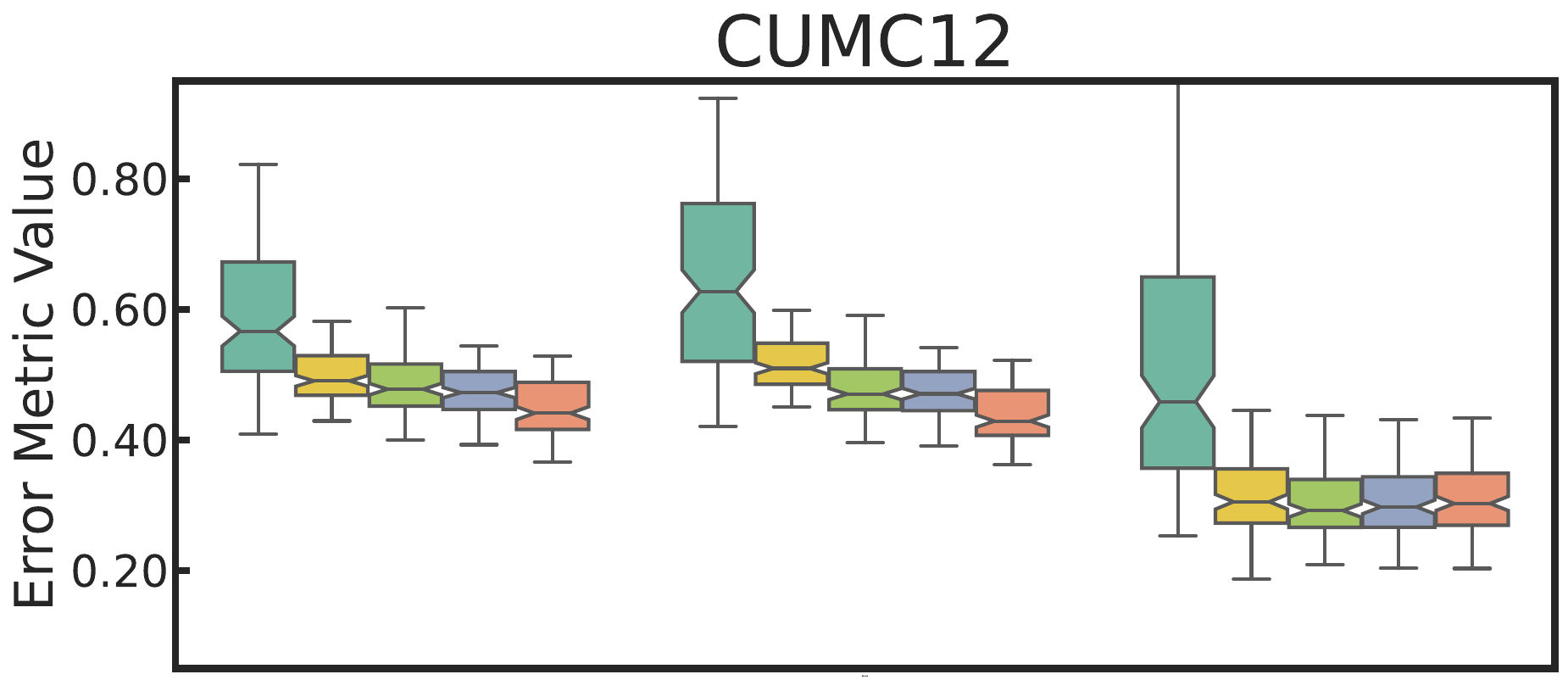}
    \braintablefig{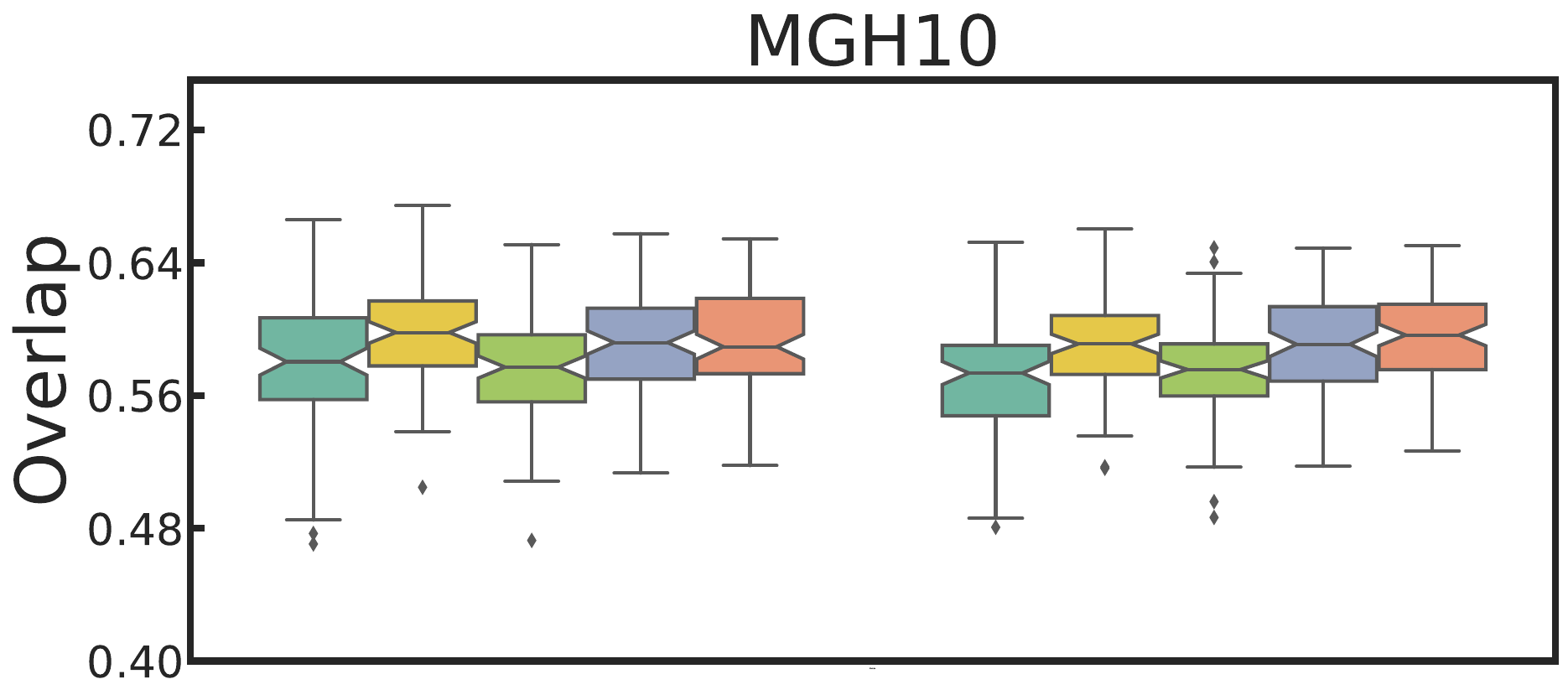}{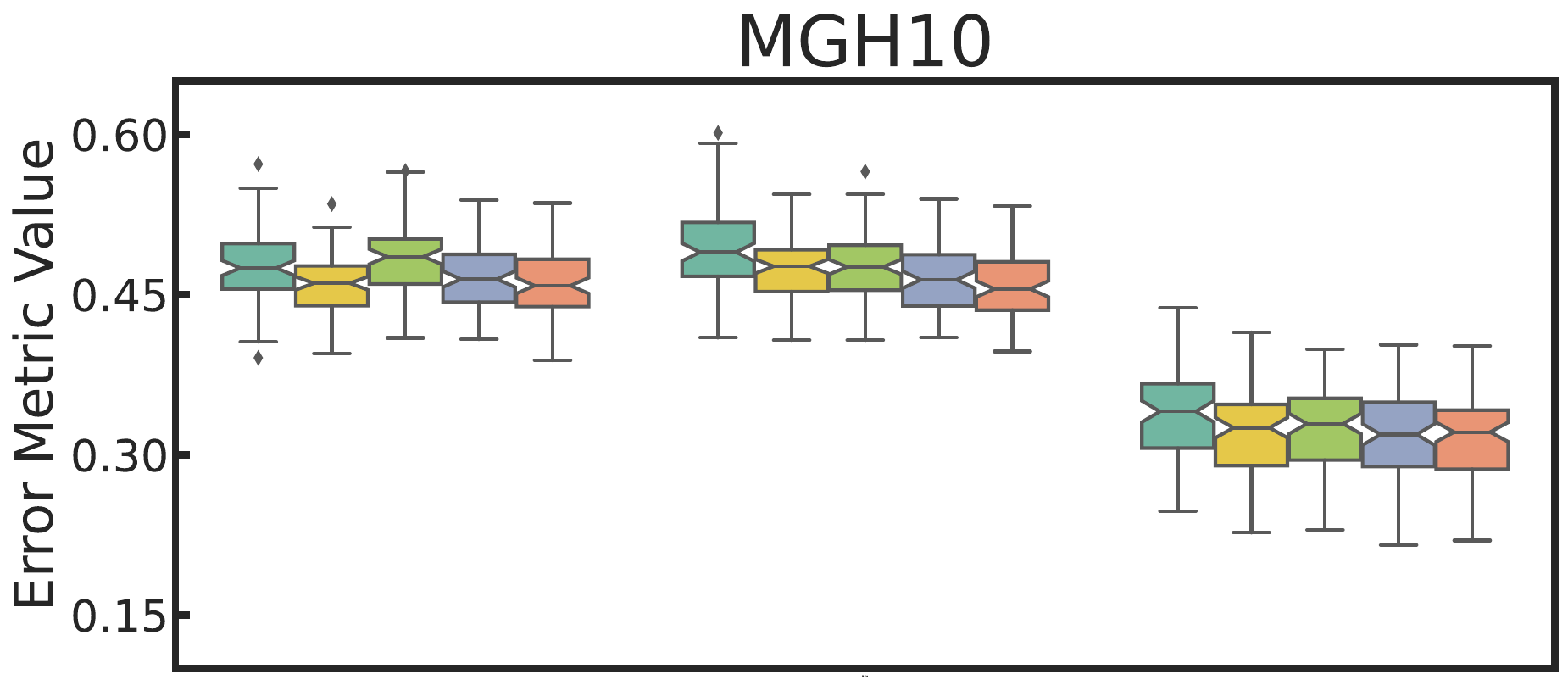}
    \braintablefig{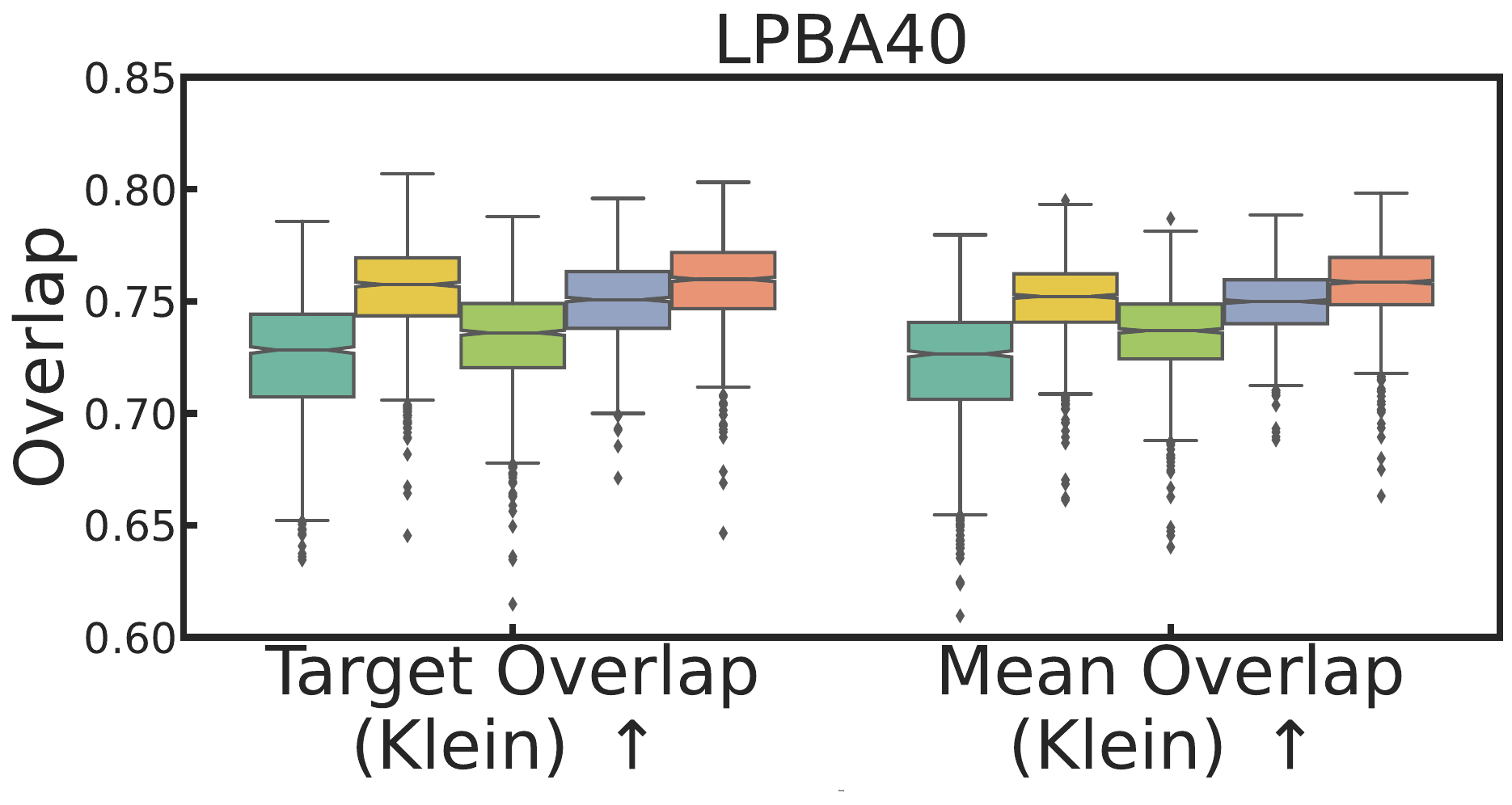}{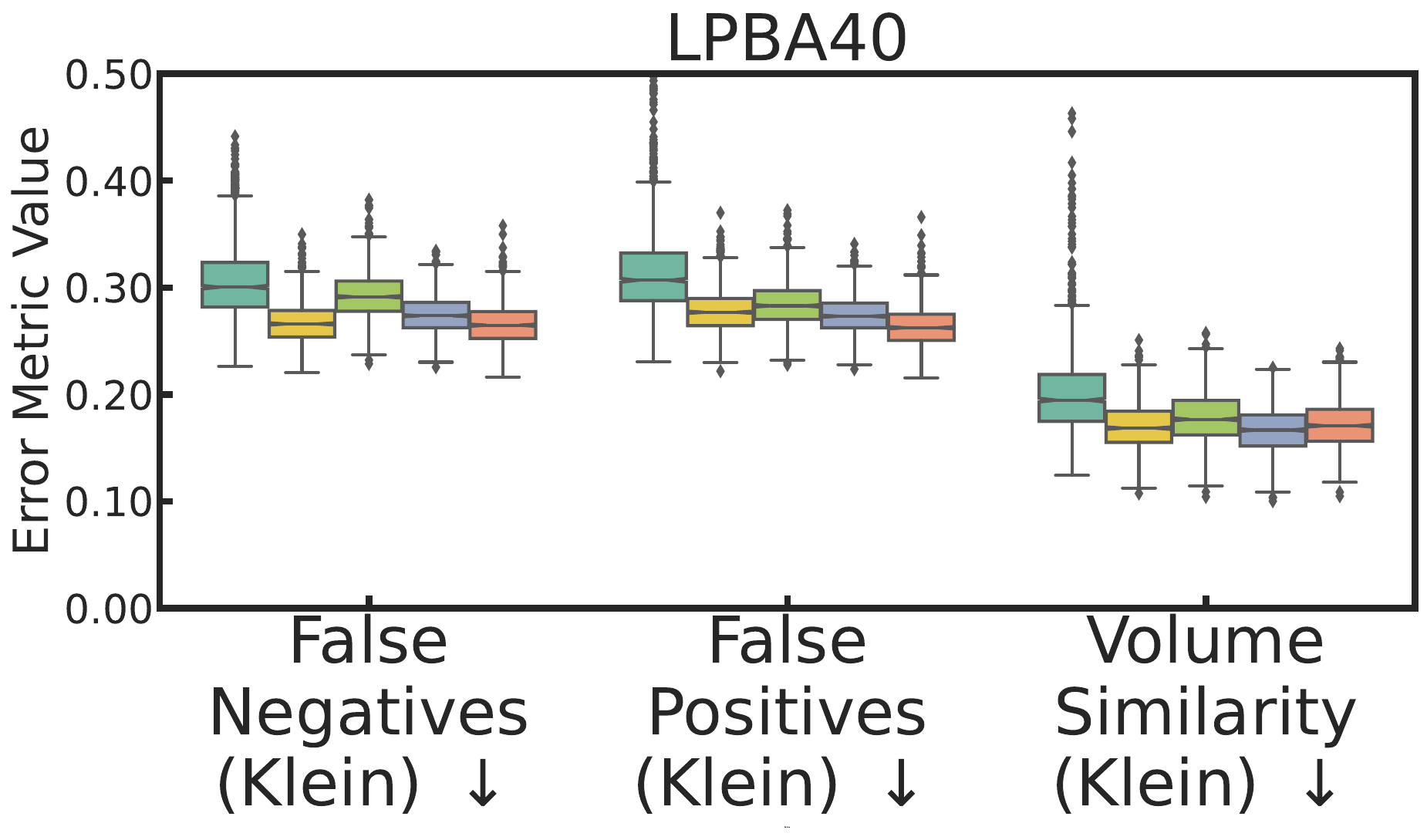}
    \caption{\textbf{Comparison of our method with ANTs on 4 MRI brain datasets}: Registration quality is validated by measuring volume overlap of label maps between the fixed and warped label maps.
    \textbf{(a)}: For anatomical region $r$, warped (binary) label map ${S}_r$ and fixed label map ${T}_r$, target and mean overlap are defined as $|{S}_r \cap {T}_r| / |{T}_r|$ and $2|{S}_r \cap {T}_r| / (|{S}_r| + |{T}_r|)$.
    We define the aggregate target overlap over all anatomical regions as $\sum_r (|{S}_r \cap {T}_r| / |{T}_r|)$ and Klein \etal~\cite{klein2009evaluation} define it as $(\sum_r |{S}_r \cap {T}_r|) / (\sum_r |{T}_r|)$, likewise for other metrics.
    The latter aggregation is denoted with the suffix (Klein) in the figure.
    In all four datasets, the boxplots show a narrower interquartile range and substantially higher median than ANTs (higher is better), underscoring the stability and accuracy of our algorithm.
    \textbf{(b)}: Other measures of anatomical label overlap used in~\cite{klein2009evaluation} are false positives ($|{T}_r\backslash {S}_r|/|{T}_r|$), false negatives ($|{S}_r\backslash{T}_r|/|{S}_r|$), and volume similarity ($2(|{S}_r| - |{T}_r|) / (|{S}_r| + |{T}_r|)$) (lower is better).
    We observe similar trends as in \textit{(a)}, with a narrower interquartile range and substantially lower median values.
    Results of per region overlap metrics are in the ~\cref{fig:regionwise-dice-brain}.%
    }
    \label{fig:braintable_klein}
\end{figure}

A natural way to evaluate whether two images are in a common coordinate frame is to evaluate the accuracy of overlap of gross morphological anatomical structures.
The method considers measures of volume and surface overlap, volume similarity, and distance measures to evaluate the alignment of anatomical regions.
Given a source label map ${S}_r$ and target label map ${T}_r$ and a cardinality operator $|.|$, we consider the following overlap measures.
We consider `target overlap' and `mean overlap' (also known as Dice score) as the primary measures of agreement between the source and target label maps.
\begin{equation}
\small TO_r = \frac{| {S}_r \cap {T}_r |}{| {T}_r |}, MO_r = 2\frac{|{S}_r \cap {T}_r|}{|{S}_r| + |{T}_r|}
\end{equation}
The aggregates over all regions are given by:
\begin{equation}
\small TO = \frac{1}{N_r}\sum_r TO_r \quad,\quad MO = \frac{1}{N_r}\sum_r MO_r
\end{equation}
Klein \etal \cite{klein2009evaluation} also propose a `Union Overlap' metric which is a monotonic function of the Dice score.
Therefore, we do not use this in our evaluation.
To complement the above agreement measures, we also compute false negatives (FN), false positives (FP), and volume similarity (VS) coefficient for anatomical region $r$:
\begin{equation}
    FN_r = \frac{|{T}_r \backslash {S}_r|}{|{T}_r|}, \quad FP_r = \frac{|{S}_r \backslash {T}_r|}{|{S}_r|}, \quad VS_r = 2\frac{|{S}_r| - |{T}_r|}{|{S}_r| + |{T}_r|}
\end{equation}
Comparison on other metrics proposed in ~\cite{klein2009evaluation} and regionwise analysis are shown in ~\cref{fig:braintable_klein,fig:regionwise-dice-brain}.
Similar to the overlap metrics, we compute the aggregates as in the original evaluation denoted by $FN_{Klein}, FP_{Klein}, VS_{Klein}$ and average over regions denoted simply by FN, FP, VS.

\paragraph{OASIS dataset}
On the OASIS dataset, we use same the evaluation criteria as in the Learn2Reg challenge~\cite{hering2022learn2reg}, i.e. Dice score overlap and 95th percentile of the Haussdorf distance computed for 35 subcortical structures.
This leads to a total of 12 evaluation metrics that we use to compare our method with 4 baselines - ANTs, Demons~\cite{vercauteren2007diffeomorphic}, VoxelMorph~\cite{balakrishnan2019voxelmorph} and SynthMorph~\cite{hoffmann2021synthmorph}, representing established classical and deep learning registration algorithms.

In total, we compare with state-of-the-art baselines on over \textit{2000 brain volume pairs}, with varying number of labeled anatomical regions and resolutions.

\subsection{Lung CT mapping challenges (EMPIRE10 and NLST)}

Registration of temporally spaced breathhold scans can help in tracking disease progression, or registration between inspirion and expiration scans can enable improved monitoring of airflow and
pulmonary function.
The EMPIRE10 challenge ~\cite{murphy2011evaluation} aims to provide a platform for in-depth evaluation and fair comparison of available registration algorithms for this application.
The dataset consists of 30 pairs of chest CT scans, with intra-subject registration across a variety of healthy or diseased subjects.
The scan pairs consist of inspiration-expiration scans, breathhold scans over time, scans from 4D data, ovine data, contrast-noncontrast, and artificially warped scan pairs.
The ovine data was acquired where breathing was controlled, and metallic markers were surgically implanted to provide landmark annotations, followed by a hole-filling algorithm to disguise the markers so that registration algorithms cannot use this artificial information.
Artificially warped scan pairs also provide ground truth correspondences for landmarks and lung boundaries.
The challenge provides a broad range of data complexity, voxel sizes and image acquisition differences.
ANTs is, again one of the top performing methods in this challenge.
Unlike the brain datasets, ground truth labels for fissure and landmarks are not provided for validation.
Therefore, we rely on the evaluation metrics computed privately by the challenge organizers in the evaluation server.
We compare our method with two powerful baselines (i) ANTs, which optimizes the diffeomorphism directly, and (ii) the \textit{DARTEL}~\cite{ashburner2007fast} formulation optimizing a stationary velocity field (SVF), where the diffeomorphism is obtained using an exponential map of the SVF.
We first affinely align the binary lung masks of the moving and fixed images using Dice loss~\cite{dice}.
This is followed by a diffeomorphic registration using the intensity images.

We use the Adam optimizer with learning rate of 3e-3, and a multi-scale optimization with downsampling rates of 6,4,2,1 for 200, 100, 50, 20 iterations.
This is followed by a diffeomorphic registration step with the same multi-scale resolutions and 200,150,75,25 iterations and a learning rate of 0.25.
We use a Gaussian kernel for gradient smoothing with $\sigma_{\text{grad}} = 6.0$ and warp smoothing with $\sigma_{\text{warp}} = 0.4$.
The optimal values for $\sigma_{\text{grad}},\sigma_{\text{warp}}$ are found by a hyperparameter grid search, and are strikingly close to the parameters used in the ANTs submission.

We evaluate three criteria: (1) fissure alignment errors (\%)—the fraction of misaligned fissure voxels (\cref{fig:fissureempire,fig:fissureallmethods}), (2) landmark distance in mm (\cref{fig:landmarkempire}), and (3) singularity errors—the fraction of non-diffeomorphic voxels (\cref{fig:singularempire}). ~\cref{fig:lungtable} highlights the impact of representation choice in modeling diffeomorphisms. DARTEL, using an exponential map, performs significantly worse than ANTs across all metrics by three orders of magnitude. In contrast, our method reduces fissure alignment error by 5$\times$ compared to ANTs and outperforms it in 5 out of 6 landmark subregions.
While all methods theoretically ensure diffeomorphism, SVF-based approaches introduce singularity errors due to non-adaptive scaling-and-squaring.
We discuss the limitations of SVF-based approaches in ~\cref{sec:limitations-stationary-velocity-fields}.
ANTs also introduces some singularities, whereas our method computes numerically perfect diffeomorphic transforms.
Finally,~\cref{fig:fissureallmethods} compares fissure alignment errors among EMPIRE10 submissions, showing FireANTs achieves the lowest landmark errors and the fastest runtime among the top 10 methods, setting new benchmarks in computational efficiency and accuracy.
Our method, on the other hand computes numerically {perfect} diffeomorphic transforms. %
Finally, we compare the fissure alignment error of all submissions in the EMPIRE10 challenge, and show the top 10 algorithms in ~\cref{fig:fissureallmethods}.
Results demonstrate that FireANTs attains the lowest landmark alignment errors compared to an array of contemporary state-of-the-art algorithms.

\paragraph{NLST dataset}
For the NLST dataset\cite{nlst}, we compare with representative state-of-the-art optimization and deep-learning baselines.
We use the evaluation criteria provided by the challenge, and measure results on the Robust Target Registration Error (TRE30) in millimeters between the registered keypoints.
Results in \cref{fig:nlst} show that FireANTs outperforms all baselines on the NLST dataset, with improvements of upto 51.6\% in robust target registration error (TRE30) of provided keypoints compared to state-of-the-art deep learning benchmarks including Im2Grid, Vector-Field Attention, RWC-Net, and a 50.8\% improvement in TRE30 over foundation models like unigradICON.
This demonstrates the broad applicability of FireANTs beyond neuroimaging applications.

\subsection{Other Datasets and Metrics}

\paragraph{PRIMatE Data Exchange (PRIME-DE)}
A growing body of research has documented the utility of MRI data to study neuroanatomical organization and function of non-human primates.
The PRIMatE Data Exchange (PRIME-DE) resource~\cite{primate} provides a platform for the neuroimaging community to facilitate the mapping of the non-human primate connectome.
We use a subset of this dataset collected from five different sources: Aix-Marseille Université, Mount Sinai School of Medicine, McGill University, Stem Cell and Brain Research Institute, and the University of California, Davis, resulting in 116 subjects, and subsequently 13340 subject pairs for registration.
We use the nBEST deep learning framework to perform cerebrum extraction, followed by tissue segmentation.
Since the images are markedly different than human brains, we affinely register all of the extracted cerebrum volumes to the first subject sorted by name to bring them to a common coordinate space and metadata.
This is followed by a diffeomorphic registration using the intensity images.
We use the Dice score of the registered tissue segmentations to evaluate the quality of registration.

\paragraph{Ultracortex}
The Ultracortex dataset~\cite{ultracortex} hosts a unique collection of ultra-high field (9.4 Tesla) MRI data of the human brain.
This dataset includes detailed structural images and high-quality manual segmentations, making it an invaluable resource for researchers in neuroimaging and computational neuroscience.
Out of the 86 T1-weighted images with resolutions spanning from 0.6 to 0.8mm, precise manual segmentation of the gray and white matter for each hemisphere is provided for 12 subjects.
We use the dataset and manually provided segmentations to evaluate the quality of registration of cortical surface mapping using Dice score.
Note that all deep learning methods run out of memory at 0.6 to 0.8mm resolutions, therefore we resample the images to 1.0mm isotropic resolution for evaluation of deep learning methods.

\paragraph{Rodent Datasets}
We use four rodent datasets in this study: Waxholm Rat Brain, Allen CCFv3 mouse brain, RnR-ExM mouse isocortex, and BICCN mouse dataset.
The datasets feature high-resolution atlases of the rat and mouse brain with four different modalities (T2$^*$w MRI, STPT, ExM, fMOST) respectively.
The motivation for using the datasets is to provide a benchmark for \textit{cross-species, multimodal} registration (Waxholm $\rightarrow$ Allen CCFv3), perform well on a high-resolution registration task and leaderboard (RnR-ExM), and to faithfully reproduce high-resolution rodent atlases (BICCN).
Similar to the Ultracortex dataset, most deep learning methods run out of memory at 25$\mu$m resolution for these images, therefore we resample the images to 50$\mu$m resolution for evaluation of deep learning methods.
To handle the multimodal nature of the cross-species registration task, we use Anatomix~\cite{anatomix} as a modality-agnostic feature extractor as feature images to perform registration.
Since the cross-species templates have different labelmaps, comparing Dice scores directly is not possible.
The Waxholm template comprises 95 labeled regions, while the Allen CCFv3 template includes over 300 regions defined in the complete ARA ontology.
To enable a comparable level of anatomical granularity, we coarsened the Allen CCFv3 parcellation to 34 regions by collapsing all subregions beyond depth 3 in the ontology hierarchy into their corresponding parent nodes.
We compute the Mutual Information (MI) between the registered label map corresponding to the Waxholm template image to that of the Allen CCFv3 template image to evaluate the quality of registration.
For the fMOST and RnR-ExM datasets, we use the images to perform pairwise registration and atlas generation respectively.
The performance on the RnR-ExM dataset is evaluated using the Dice score of the registered label map corresponding to the ExM image pairs on a private evalaution server.
For the BICCN dataset, we only provide qualitative results due to the lack of an evaluation criteria.

\paragraph{Zebrafish Datasets}
Analysis of the zebrafish is a growing field of research due to its unique advantages as a vertebrate model organism. The zebrafish brain is small yet structurally complex, offering a tractable system for studying whole-brain organization, development, and function at cellular resolution. High-quality atlases such as AZBA and Z-Brains provide detailed anatomical and gene expression reference templates, enabling cross-modality and cross-sample comparison. These datasets present a valuable testbed for registration algorithms, as they involve significant structural variability, diverse imaging modalities (including confocal, light-sheet, and two-photon microscopy), and finely detailed neuroanatomical annotations. Accurate registration in this setting is critical for integrating large-scale imaging data and mapping functional or genetic information onto common anatomical frameworks.
The adult and larval zebrafish brains have very different structural organization, and show very different characteristics than human brains.
This is a challenging registration task to truly access the out-of-distribution generalization capabilities of registration algorithms.
Due to the lack of a consensus on appropriate evaluation criteria beyond qualitative comparison for these datasets, we use the mutual information between the registered image and template image to evaluate the quality of registration.

\paragraph{Learn2Reg Abdomen MRCT registration}
This dataset is used as a testbed to ablate the effect of Jacobian-free optimization on abdominal MRCT registration.
Abdomen CT-MR registration is a conceptually different registration task copmared to neuroanatomical or pulmonary registration with completely unrelated anatomical structures, organization, and biomechanical dynamics.
We use the validation split provided by the challenge to evaluate the performance of FireANTs with and without Jacobian-free optimization.

\section{Modular software implementation to enable effective experimentation}
Registration is a key part of many data processing pipelines in the clinical literature. Our software implementation is designed to be extremely flexible, e.g., it implements a number of existing registration methods using our techniques, modular, e.g., the user can choose different group representations (rigid or affine transforms, diffeomorphisms), objective functions, optimization algorithms, loss functions, and regularizers. Users can also stack the same class of transformations, but with different cost functions. For example, they can fit an affine transform using label maps and Dice loss, and use the resultant affine matrix as initialization to fit another affine transform using the cross-correlation registration objective. This enables seamless tinkering and real-time investigation of the data. Deformations can also be composed in increasing order of complexity (rigid $\to$ affine $\to$ diffeomorphisms), thereby avoiding multiple resampling and subsequent resampling artifacts. We have developed a simple interface to implement custom cost functions, which may be required for different problem domains, with ease; these custom cost functions can be used for any of the registration algorithms out-of-the-box. Our implementation can handle images of different sizes, anisotropic spacing, without the need for resampling into a consistent physical spacing or voxel sizes. All algorithms also support multi-scale optimization (even with fractional scales) and convergence monitors for early-stopping.

Our software is implemented completely using default primitives in PyTorch. All code and example scripts is available at
\href{https://github.com/rohitrango/fireants}{https://github.com/rohitrango/fireants}.

\section{On the Ill-conditioning of Image Registration}
\label{sec:twodim}
\begin{figure*}[!h]
  \centering
  \begin{subfigure}[b]{0.9\linewidth}
    \includegraphics[width=\linewidth]{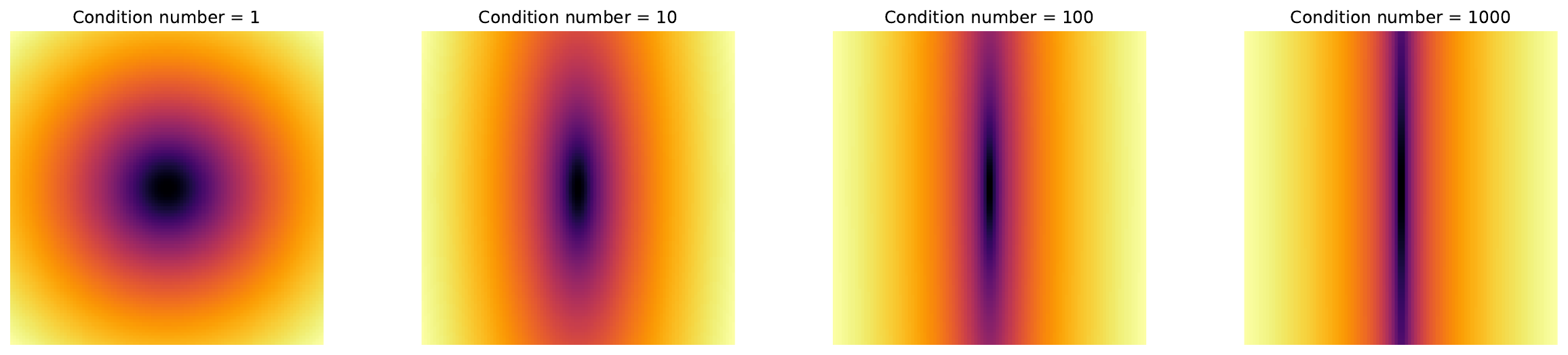}
    \caption{Log-Loss landscape of the toy problem $f_\kappa(x, y) = x^2 + \kappa y^2$ for $\kappa = 1, 10, 100, 1000$. The log-loss becomes increasingly sharp along the y-direction as $\kappa$ increases.}
    \label{fig:toyloss}
  \end{subfigure}
  \begin{subfigure}[b]{0.9\linewidth}
    \includegraphics[width=\linewidth]{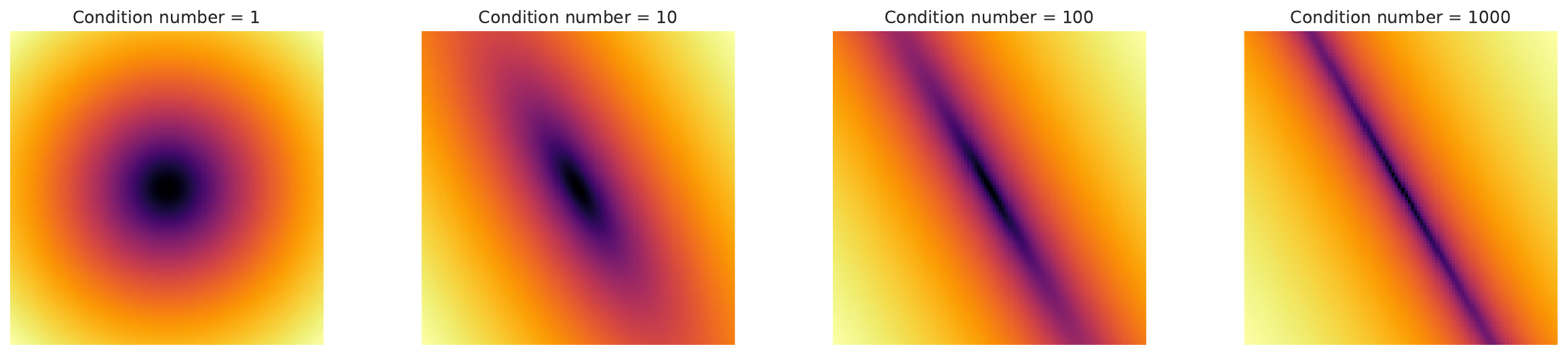}
    \caption{Log-Loss landscape of the toy problem $f_\kappa(x, y) = x_\theta^2 + \kappa y_\theta^2$ for $\kappa = 1, 10, 100, 1000$, where $(x_\theta, y_\theta)$ is the coordinate $(x, y)$ rotated by an angle $\theta$ about the origin.} %
    \label{fig:toylossrotated}
  \end{subfigure}
  \begin{subfigure}[b]{0.47\linewidth}
    \raggedleft
    \includegraphics[width=0.49\linewidth]{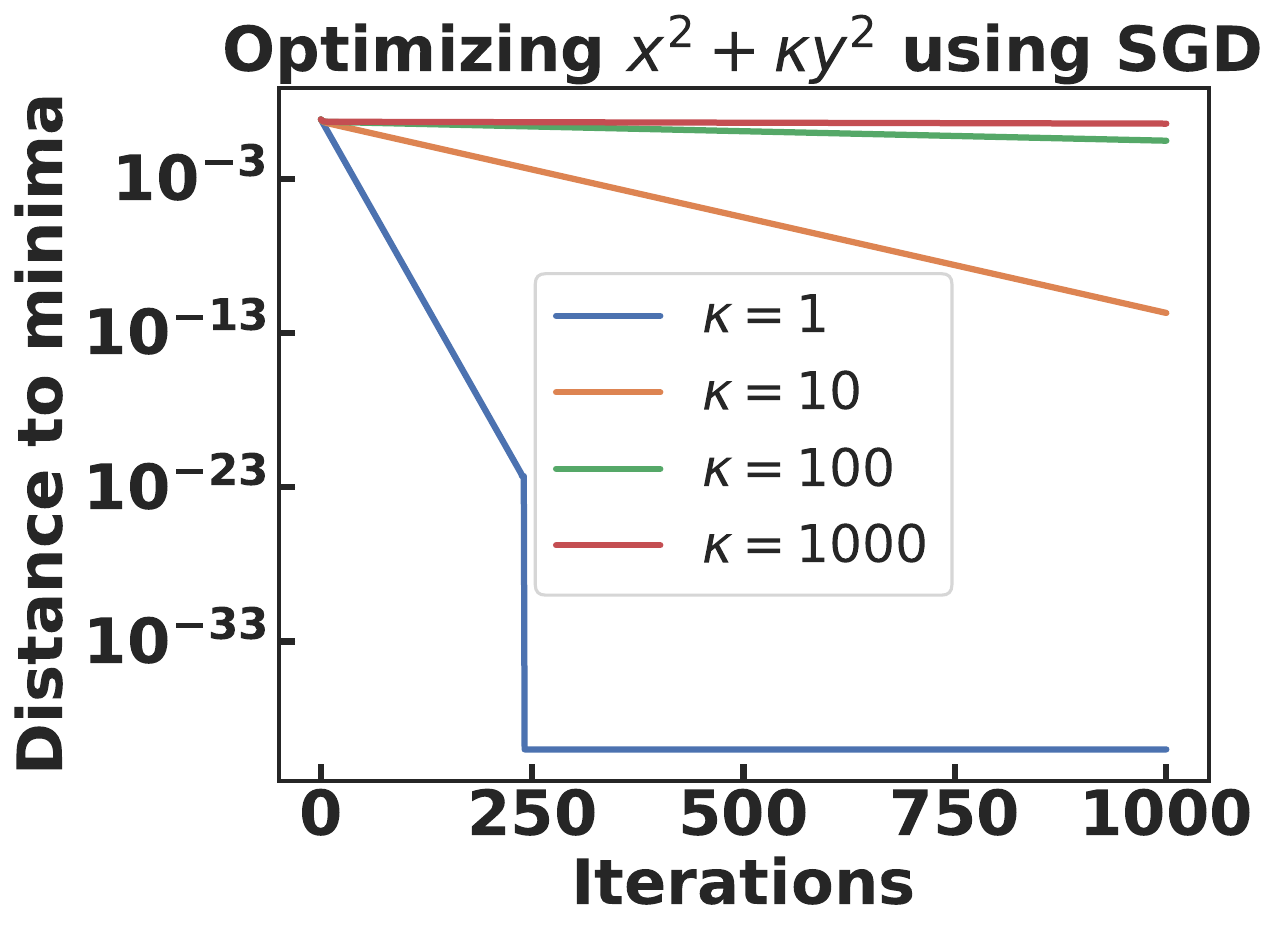}
    \includegraphics[width=0.49\linewidth]{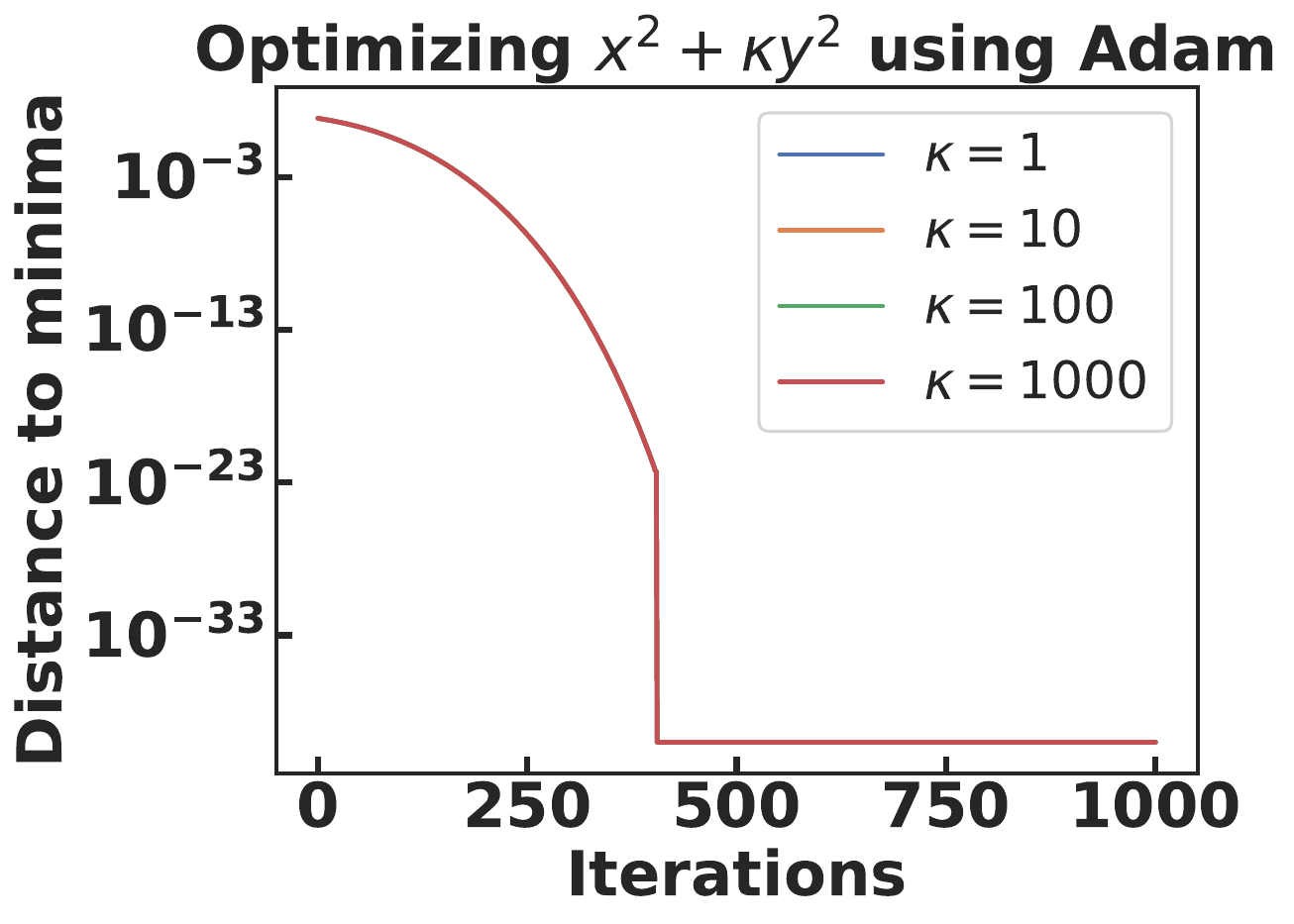}
    \caption{Optimization of \cref{fig:toyloss} using SGD and Adam shows that SGD fails to recover the minima for $\kappa \ge 100$ while Adam is \textit{invariant} to the condition number for diagonal Hessian matrices.
    This is a strong motivation to use first order adaptive optimization for registration where the condition number can exceed $10^{5}$.
    }
    \label{fig:toyoptim}
  \end{subfigure}
  \begin{minipage}{0.02\linewidth}
    \hfill
  \end{minipage}
  \begin{subfigure}[b]{0.47\linewidth}
    \raggedright
    \includegraphics[width=0.49\linewidth]{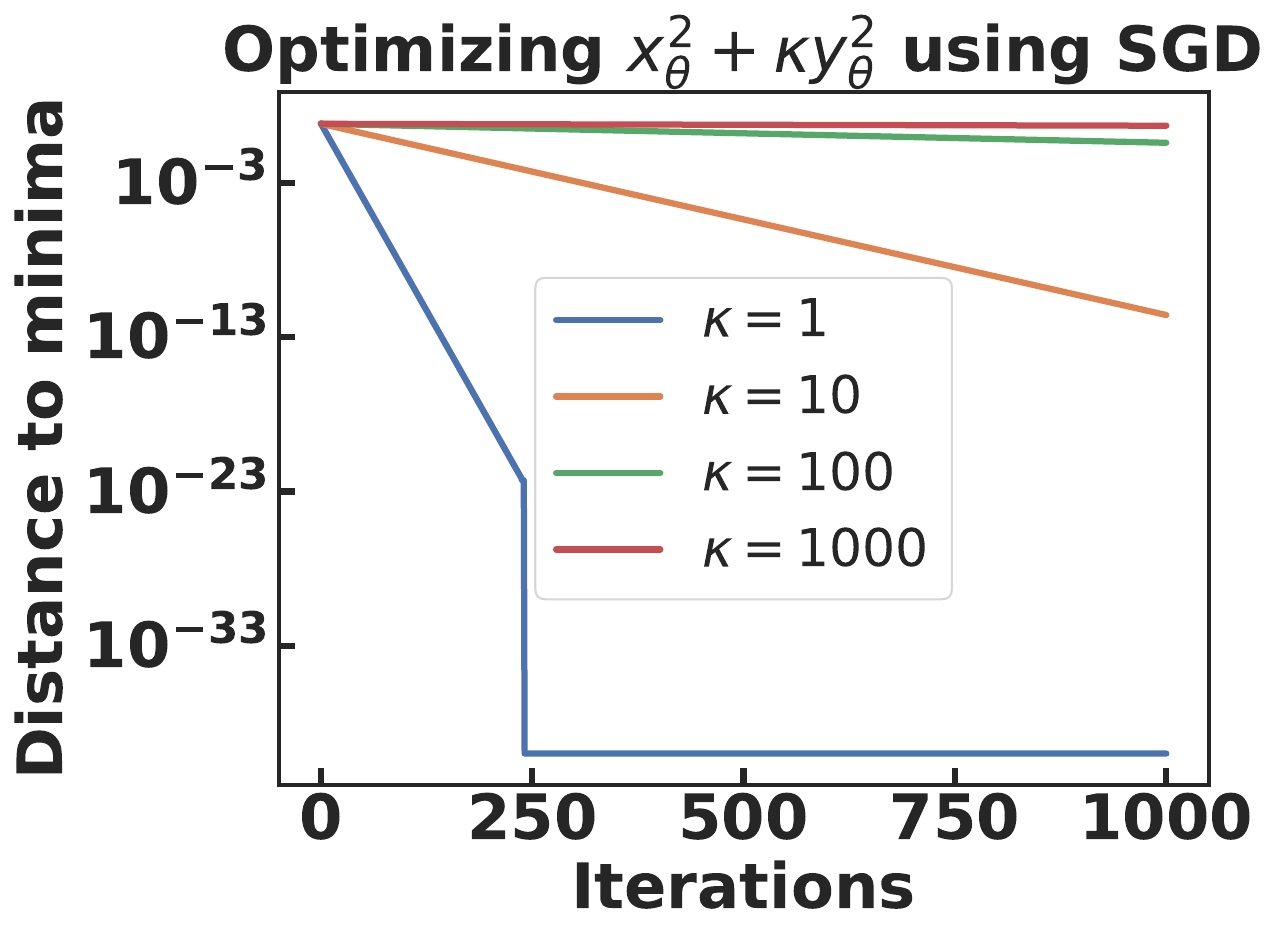}
    \includegraphics[width=0.49\linewidth]{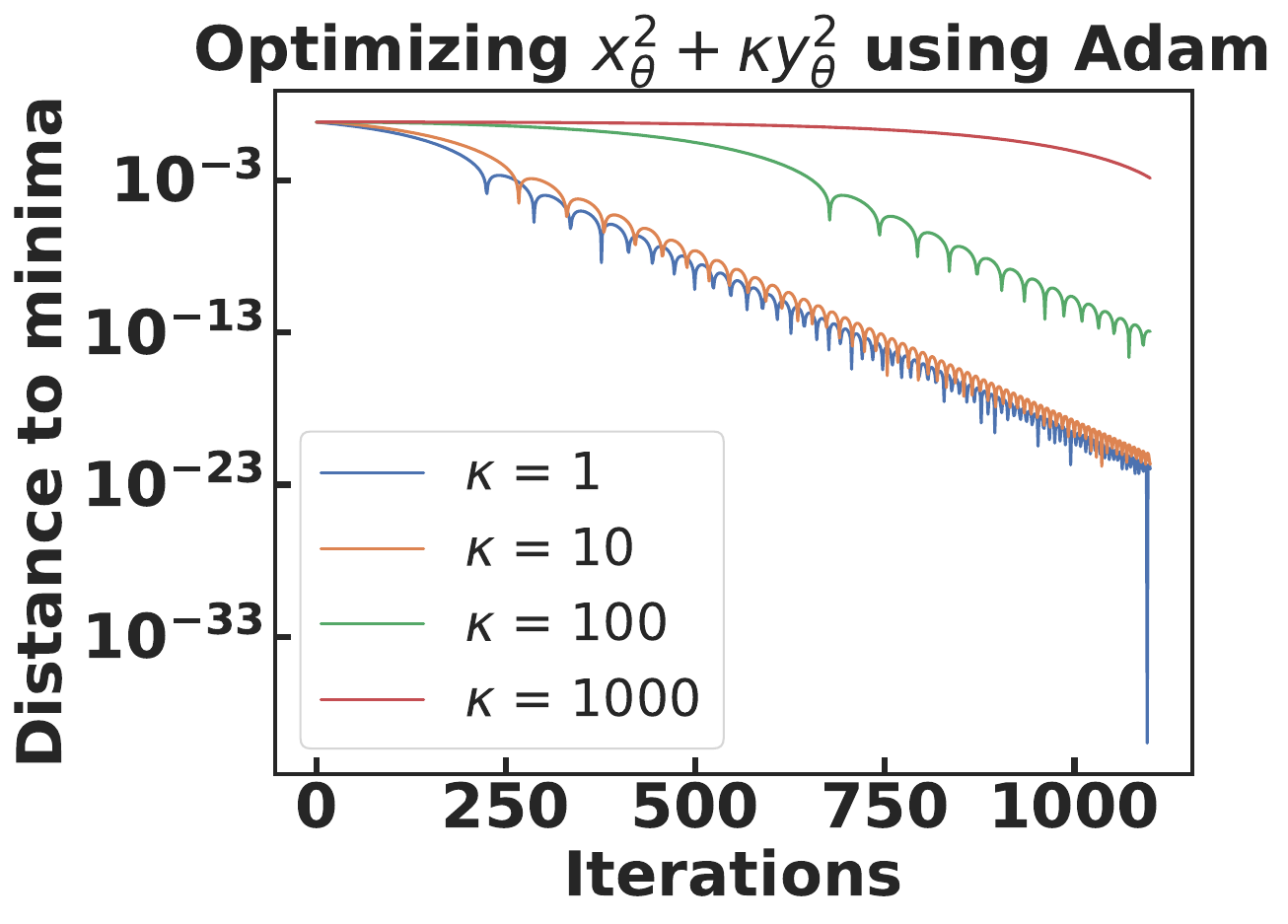}
    \caption{Optimization of \cref{fig:toylossrotated} using SGD shows identical optimization trajectories as \cref{fig:toyoptim}. Adam, however, is not invariant to the condition number because the difference between the true Hessian and its diagonal approximation increases with $\kappa$. Even so, the final point is at a distance of less than $10^{-3}$ units to the minima, showing the mitigating effect of adaptive optimization even for non-diagonal Hessians.}
    \label{fig:toyoptimrotated}
  \end{subfigure}
\end{figure*}

Image registration is a highly ill-conditioned, and non-convex problem necessitating advanced optimization methods for convergence.
To provide more intuition on the effect of $\kappa$ on convergence of the SGD algorithm, we consider a toy example of a 2D optimization problem.
Specfically, we consider a loss function $f_\kappa(x, y) = x^2 + \kappa y^2$ where $\kappa > 1$ becomes the condition number of the problem.
Qualitatively, the effect of the first term diminishes exponentially fast with $\kappa$ (\cref{fig:toyloss}).
Quantitatively, we run both SGD and Adam optimization for a 1000 iterations starting from the point $(x, y) = (5, 5)$.
\cref{fig:toyoptim} shows that SGD works extremely well for $\kappa = 1$ which is the best-conditioned loss function, but quickly gets stuck for $\kappa \ge 100$.
On the contrary, Adam is invariant to the condition number and converges to the minima for all values of $\kappa$.
This is because for a diagonal Hessian (as in this case), the second-order adaptive terms are proportional to the diagonal elements of the Hessian.
These condition numbers are vanishingly small compared to those in typical image registration tasks, which can exceed $10^{5}$, making them extremely ill-conditioned.

\begin{figure}[H]
    \centering
    \includegraphics[width=0.65\linewidth]{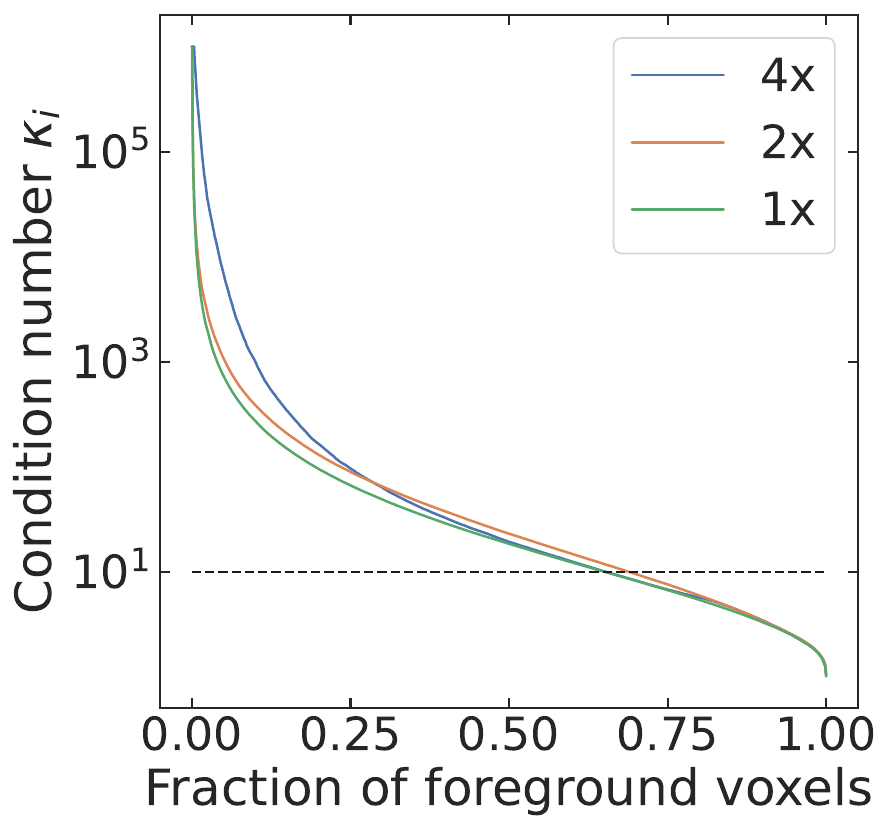}
    \caption{\textbf{Deformable image registration is ill-conditioned}. 
    To quantitatively examine ill-conditioning in registration, we compute the distribution of per-pixel condition number for a MRI registration task, at different image downsampling factors (denoted as 1x, 2x, and 4x). A high condition number signifies exacerbated ill conditioning and requires higher-order optimization.
    A horizontal dashed line denoting $\kappa = 10$ is drawn as a reference for substantial ill conditioning.
    Across all scales, a substantial fraction of foreground voxels are ill-conditioned ($\kappa > 10$), necessitating adaptive first-order optimization for faster convergence and accurate registration.}
    \label{fig:conditioning}
\end{figure}

We also consider a more realistic, but tractable scenario of the convex loss function $f_{\kappa,\theta}(x, y) = x_\theta^2 + \kappa y_\theta^2$, where
$$ \begin{bmatrix}
  x_\theta \\
  y_\theta
\end{bmatrix} = \begin{bmatrix}
  \cos(\theta)  & \sin(\theta) \\
  -\sin(\theta)  & \cos(\theta)
\end{bmatrix} \begin{bmatrix}
  x \\
  y
\end{bmatrix}$$
We choose $\theta = \pi/3$ for this experiment.
This is simply a rotated version of the previous family of loss functions, as shown in \cref{fig:toylossrotated}.
The trajectories obtained from optimization using SGD (\cref{fig:toyoptimrotated}) are virtually identical to that in \cref{fig:toyoptim} since the new gradients are simply rotated versions of the previous gradients, and the distance from the minima is invariant to the rotation.
However, the trajectories from Adam optimization are qualitatively very different, owing to the increasing difference between the true Hessian and its diagonal approximation.
Even so, the final point is at a distance of less than $10^{-3}$ units to the minima for $\kappa = 1000$, showing the effectiveness of adaptive optimization even for ill-conditioned, non-diagonal Hessians.
This is a strong motivation to extend adaptive optimization for non-Euclidean diffeomorphic registration, which is very high-dimensional and ill-conditioned.

\section{Limitations of Stationary Velocity Fields}
\label{sec:limitations-stationary-velocity-fields}

A common approach in diffeomorphic registration is to use stationary velocity fields, i.e. velocity fields that are constant in time.
This velocity field is also an element of the Lie algebra that can be used to generate a diffeomorphism using the exponential map.
Since the velocity field itself resides in Euclidean space, adaptive optimization algorithm like Adam can be applied to optimize the velocity field.
Many deep learning methods employ this approach since it is hard to produce valid diffeomorphic transforms using a network but it is easy to produce a valid stationary velocity field.
CLAIRE\cite{mang2024claire} mentions the limitations of this approach without further elaboration.
We discuss three limitations of the SVF based optimization approach:

\paragraph{Computational Cost}
Optimizing the velocity field requires computing the exponential map using the scaling-and-squaring approach (\cref{eq:scalingandsquaring}) to obtain the diffeomorphism.
Typical registration pipelines run scaling-and-squaring 6-8 times to obtain the diffeomorphism, and the backprop requires another 6-8 steps of iterative backward calls to compute the gradient of the velocity field.
This is a significantly expensive operation performed \textit{every iteration} of the optimization, leading to a substantial slowdown in runtime.
In contrast, direct optimization requires only one warp composition to perform the diffeomorphic update $\varphi_{t+1} = \varphi_t \circ (id + \eta_t v_t)$.
The significant difference in runtime is observed for ANTs and DARTEL in \cref{fig:timinglung}.
SVFs also cannot represent diffeomorphisms that are integrals of time-dependent velocity fields, which are more flexible and can represent a wider range of deformations \cite{wu2022nodeo,wu2024neural,mang2024claire}.

\paragraph{Tradeoff between Numerical Accuracy and Computational Cost}
Exponential maps of SVFs are mathematically diffeomorphic in nature, but it is observed that numerically SVFs may not be diffeomorphic.
Other works \cite{wu2022nodeo} show that SVF based baselines like Log Demons have significantly more non-diffeomorphic voxels than time-dependent velocity fields like SyN and NODEO.
We see similar trends in \cref{fig:singularempire}, where the SVF based DARTEL has significantly more non-diffeomorphic voxels than the direct optimization in ANTs and FireANTs.
This numerical inaccuracy traces back to the discretization error in the scaling and squaring approach, which is essentially Euler integration of the velocity field.
In the base case of the scaling-and-squaring approach (\cref{eq:ssbasecase}), $\varphi^{(1/2^M)} = id + v_0 / 2^M$ is not guaranteed to be a diffeomorphism unless the Lipschitz constant of $v_0$, denoted as $LP(v_0)$ is less than $2^M$.
The inductive recursion step (\cref{eq:scalingandsquaring}) only preserves the diffeomorphic property if the base case is a diffeomorphism, otherwise the non-diffeomorphism can propagate throughout the subsequent steps to the final diffeomorphism.
Here, we provide a proof that the base case is a diffeomorphism only for $v_0$ with Lipschitz constant less than $2^M$, showing that velocity fields with large deformations or highly variable deformations require finer Euler integration steps (i.e. larger $M$) to ensure numerical diffeomorphisms.

\begin{theorem}
  For a $C^\infty(\Omega, \bR^d)$ velocity field $v_0$ with compact support on $\Omega$ such that $v_0(x) = 0$ on $x \in \partial \Omega$, the transform $\varphi = id + \epsilon v_0$ is a diffeomorphism for $|\epsilon| < 1/LP(v_0)$, where $LP(v_0)$ is the Lipschitz constant of $v_0$.
\end{theorem}

\begin{proof}
  Since $v_0$ is a $C^\infty(\Omega, \bR^d)$ (is continuously differentiable and is compact on $\Omega$) velocity field, the Jacobian of the velocity exists, and is denoted as $J(v_0)$.
  We invoke the Hadamard's global inverse function theorem~\cite{Hadamard1906} (HGIF theorem) to show that $\varphi$ is a diffeomorphism for $|\epsilon| < 1/LP(v_0)$.

  The HGIF theorem requires that $\varphi$ is smooth (true by our definition), and the Jacobian of the transformation is non-singular for all $x \in \Omega$, and that $\| J\varphi(x) ^{-1} \|$ is bounded for all $x \in \Omega$.
  Since $v_0$ is defined only on a compact domain $\Omega$, we use the Whitney extension theorem~\cite{whitney1992analytic} to extend $v_0$ to a $C^\infty(\bR^d)$ velocity field by simply setting $v_0(x) = 0$ for $x \in \bR^d \setminus \Omega$.

  For $x \in \bR^d \setminus \Omega$, we have $\varphi(x) = x$, and therefore $J\varphi(x) = I$ which is invertible, and $\| J\varphi(x) ^{-1} \| = 1$ which is bounded.

  For $x \in \Omega$, we have
  \begin{align}
    J\varphi(x) &= I + \epsilon Jv_0(x) \\
    \implies \| J\varphi(x) - I \| &= |\epsilon| \| Jv_0(x) \| \le |\epsilon| LP(v_0) < 1
  \end{align}
  since $|\epsilon| < 1/LP(v_0)$.
  Since $\| J\varphi(x) - I \| < 1$, $J\varphi(x)$ is non-singular for all $x \in \Omega$ from the Neumann convergent series of matrix $(I - A)$ (i.e. $A^{-1} = \sum_{k=0}^{\infty} (I-A)^k$).

  For $A = J\varphi(x)$, we have $\| I - A \| < 1$ from the above inequality. Let $\| I - A \| \le \delta$ for some $\delta < 1$.
  Using the Neumann convergent series of matrix $(I - A)$ (i.e. $A^{-1} = \sum_{k=0}^{\infty} (I-A)^k$ for $\| I - A \| < 1$), we have
  \begin{align}
    \| A^{-1} \| &\le \sum_{k=0}^{\infty} \| (I-A)^k \| \\
    &\le \sum_{k=0}^{\infty} \delta^k  = \frac{1}{1 - \delta}
  \end{align}
  This shows that $\| J\varphi(x) ^{-1} \| \le \frac{1}{1 - \delta}$ for all $x \in \Omega$ and is bounded.

  Since $\varphi$ is $C^\infty(\bR^d, \bR^d)$, and the Jacobian is non-singular and its inverse is bounded for all $x \in \bR^d$, we have that $\varphi$ is a diffeomorphism for all $x \in \bR^d$.
\end{proof}
If $|\epsilon| \ge 1/LP(v_0)$, then $\| J\varphi(x) - I \| < 1$ may not hold and the Jacobian may be singular for some $x \in \Omega$, breaking local invertibility.
When scaling-and-squaring is employed, the velocity field might have large magnitudes to capture large deformations during optimization, leading to a large Lipschitz constant.
Fixing the number of integration steps $M$ can lead to numerical non-diffeomorphisms if the Lipschitz constant exceeds $2^M$.
In principle, $M$ should adaptively chosen to the lowest value such that $2^M > LP(v_0)$.

\begin{figure}[t]
\centering
    \includegraphics[width=\linewidth]{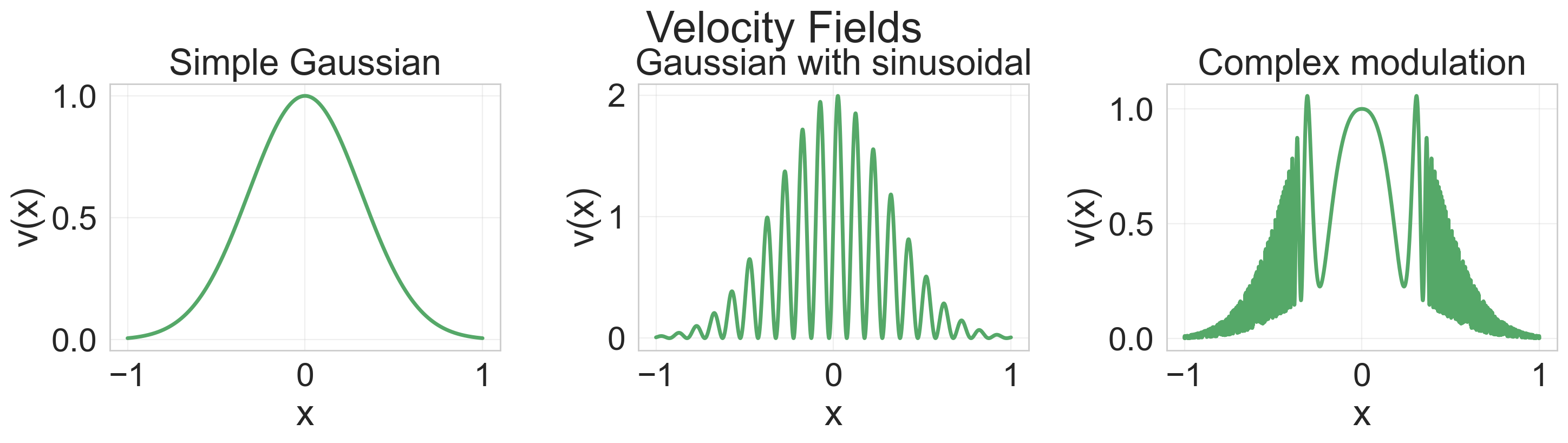}
    \caption{Three 1D velocity fields with increasing Lipschitz constants to illustrate the dependence of the number of integration steps $M$ on ensuring numerically accurate diffeomorphisms.}
    \label{fig:scaling-and-squaring-functions}
\end{figure}

\paragraph{Empirical Verification}
To demonstrate this empirically, we choose three 1D velocity fields over the interval $\Omega = [-1, 1]$ with increasing Lipschitz constants (illustrated in \cref{fig:scaling-and-squaring-functions}), and plot the amount of non-diffeomorphic voxels as a function of $M$. %
We choose 1D velocity fields for simplicity and easy visualization but all results generalize to higher dimensions.
The three velocity fields are named and defined as follows:
\begin{itemize}
  \item Simple gaussian: $v(x) = \exp(-5x^2)$
  \item Gaussian with sinusoidal: $v(x) = \exp(-5x^2) (1 + \sin(20 \pi x))$
  \item Complex modulation: $v(x) = \exp(-5x^2) (1 + 0.7 \sin(\pi \exp(|\pi x|^3)))$
\end{itemize}

\begin{figure}[t]
  \centering
  \includegraphics[width=\linewidth]{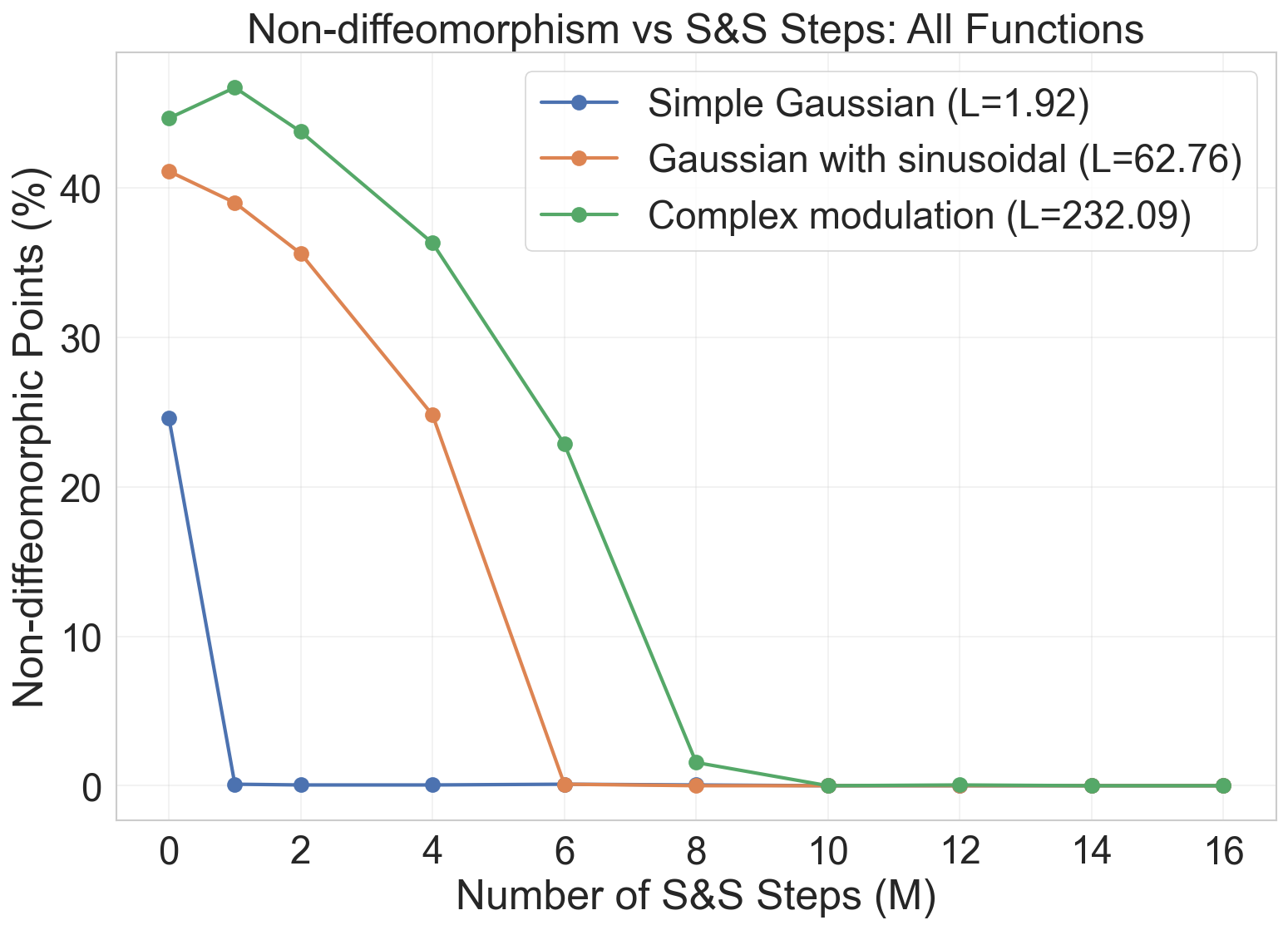}
  \caption{Fraction of non-diffeomorphic voxels as a function of $M$ for the three velocity fields.}
  \label{fig:non-diffeo-analysis}
\end{figure}

\begin{figure}[t]
  \centering
  \includegraphics[width=0.9\linewidth]{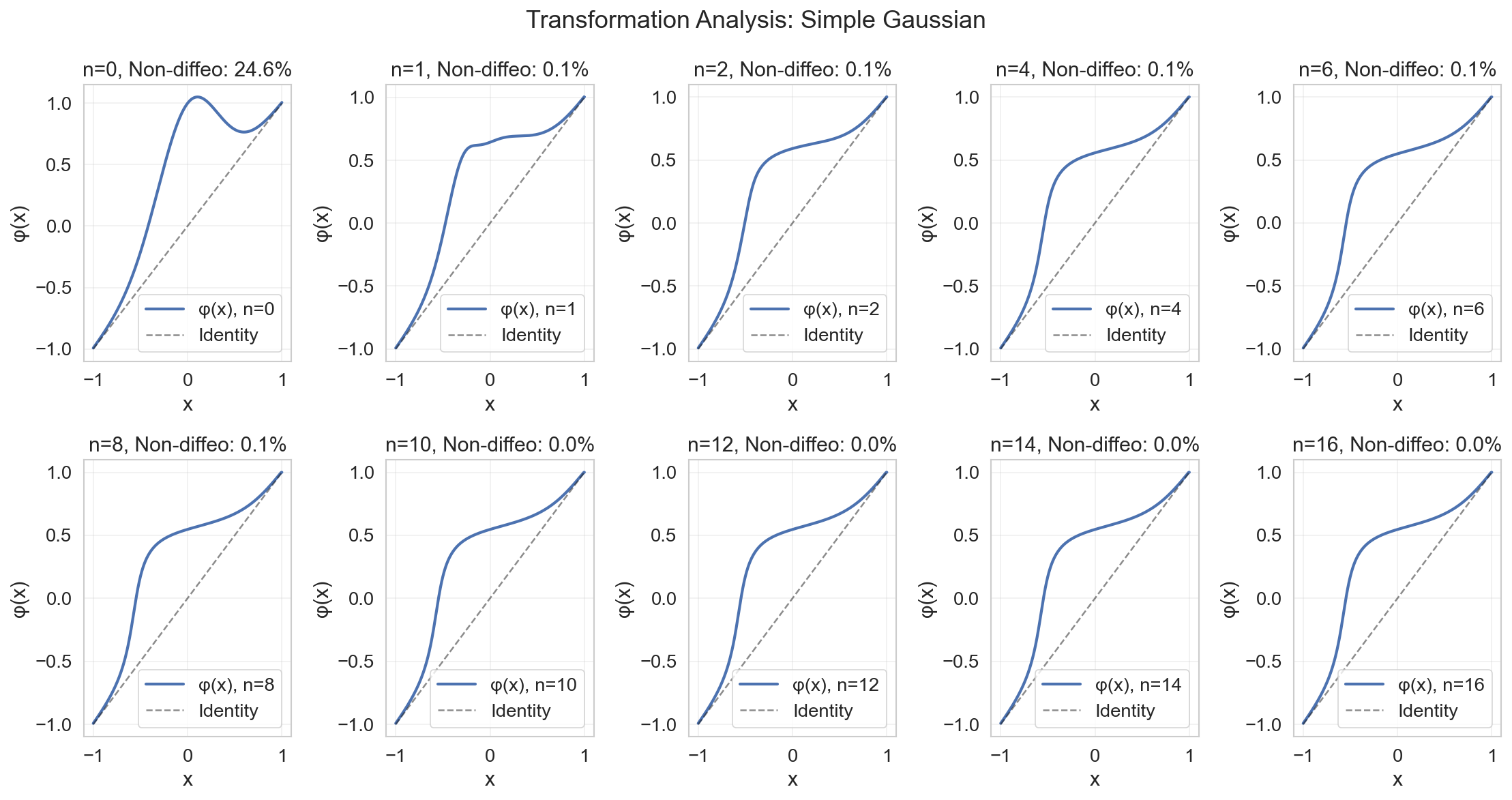}
  \includegraphics[width=0.9\linewidth]{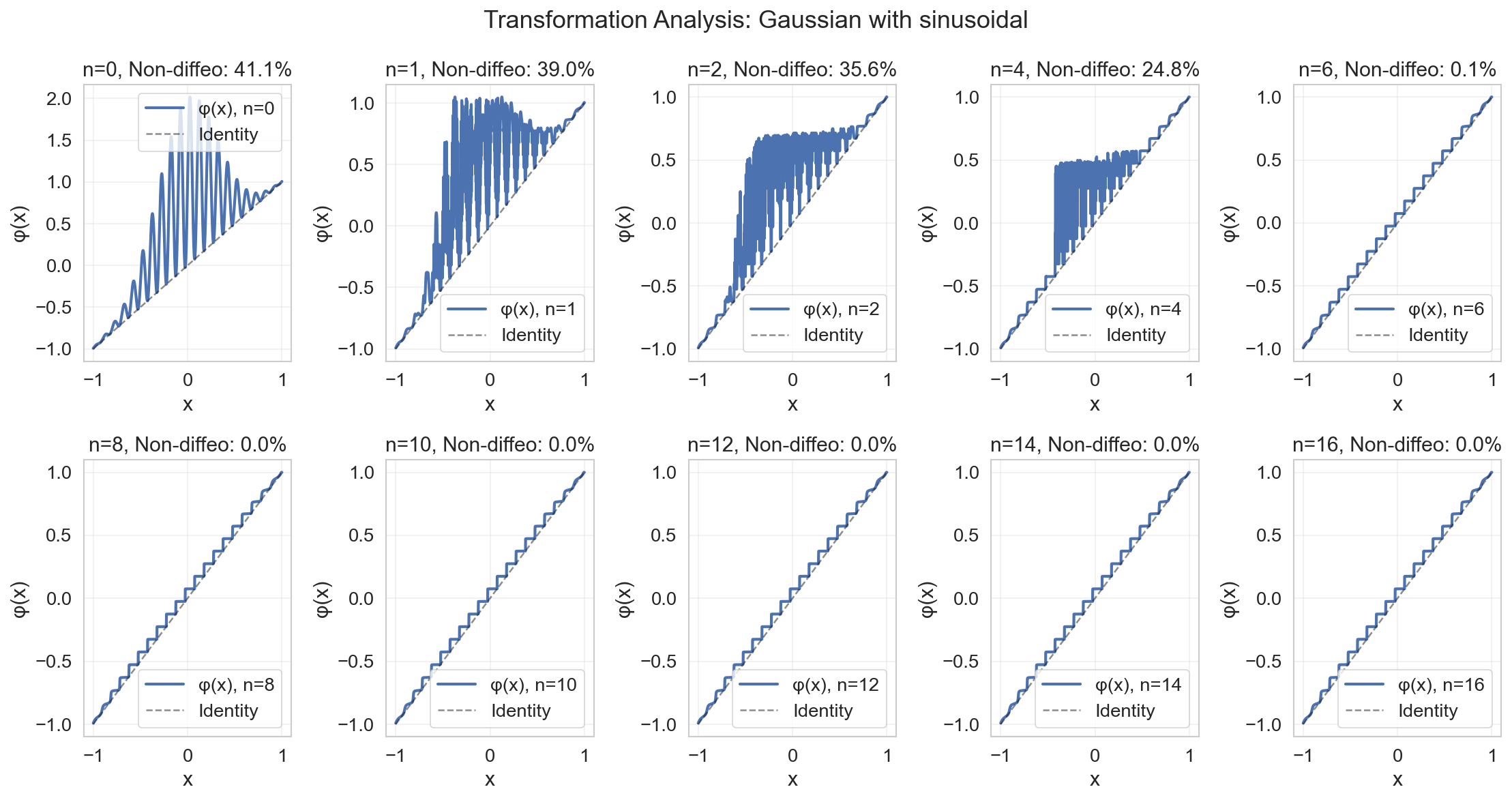}
  \includegraphics[width=0.9\linewidth]{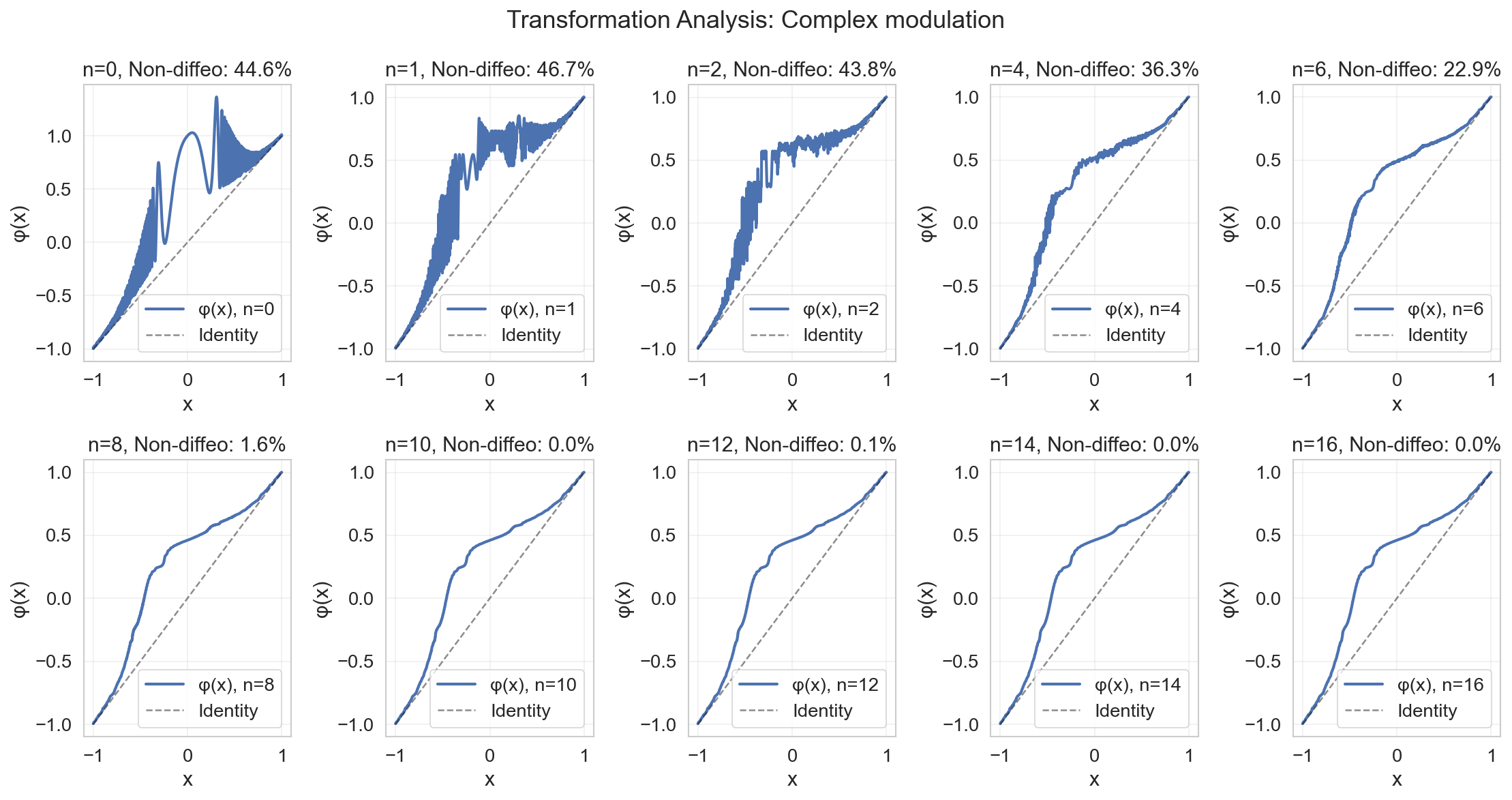}
  \caption{Illustrative example of the effect of the number of integration steps $M$ for scaling-and-squaring and the final deformation obtained for the three velocity fields.
  Larger Lipschitz constants require a larger number of integration steps to ensure numerical diffeomorphisms with the scaling-and-squaring approach.
  }
  \label{fig:scaling-and-squaring-results-ablation}
\end{figure}

The fraction of non-diffeomorphic voxels as a function of $M$ is shown in \cref{fig:non-diffeo-analysis}, along with the Lipschitz constant of the velocity fields.
This plot shows that the Simple Gaussian velocity field required only 1 integration step to ensure a diffeomorphism over $\Omega$ but the sinusoidal velocity field required 6 steps and the complex modulation velocity field required 10 steps.
Note that all three velocity fields are bounded by 1, but their Lipschitz constants are different by orders of magnitude.
The result of the exponential map computed with increasing number of integration steps $M$ for each velocity field is also visualized in \cref{fig:scaling-and-squaring-results-ablation}.
Velocity fields with larger Lipschitz constants require a larger number of integration steps regardless of the actual magnitude of the velocity field.
Since the Lipschitz constant of the velocity field is not known a priori, this creates a tradeoff between numerical accuracy and computational cost.

In contrast, when optimizing diffeomorphisms directly using the update rule $\varphi_{t+1} = \varphi_t \circ (id + \epsilon_t v_t)$, the scaling factor $\epsilon_t$ is chosen to be $\eta / LP(v_t)$ where $\eta$ is the learning rate.
$LP(v_t)$ can be bounded by the half of the norm of the velocity field divided by the resolution of the image, since we have
\begin{align}
  LP(v_t) = \sup_{x, y \in \Omega} \frac{\| v_t(x) - v_t(y) \|}{\| x - y \|} \le \sup_{x' \in \Omega} 2\frac{\| v_t(x') \|}{\| \delta x' \|}
\end{align}
avoiding the need to compute the Lipschitz constant directly.

\begin{figure}[h!]
    \centering
    \includegraphics[width=0.9\linewidth]{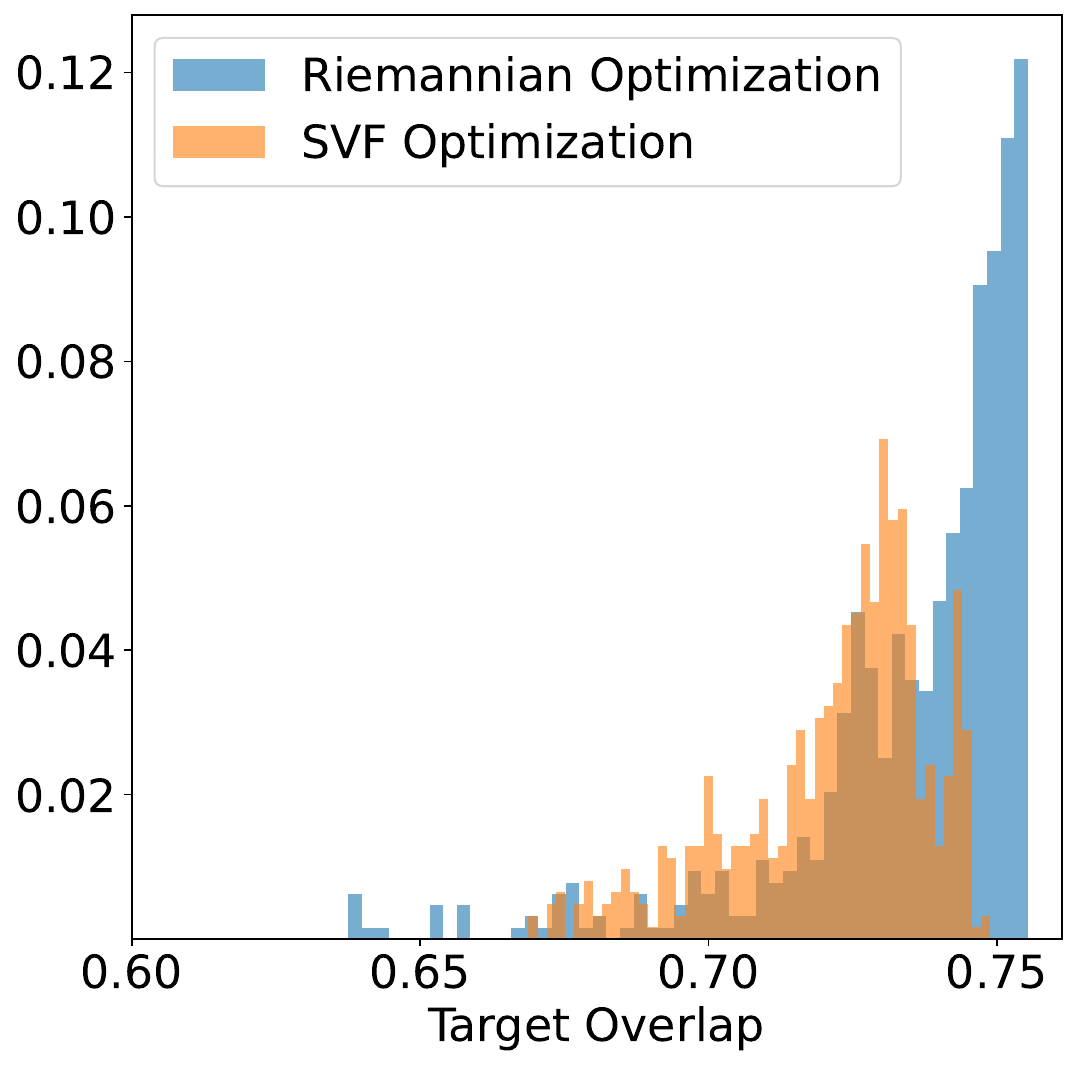}
    \caption{ %
    \textbf{Comparison of exponential versus direct optimization on LPBA40 dataset}: We run the hyperparameter grid search on the LPBA40 dataset using direct Riemannian gradient updates with Adam optimizer (denoted as \textit{rgd}), and optimizing the velocity field by computing the exponential map to represent the diffeomorphism (denoted as \textit{exp}) across all the configurations shown in Fig.~\ref{fig:hyperparam}(a).
    The average target overlap for each configuration is then stored, and a histogram of target overlap values of the dataset is constructed.
    Note that the \textit{rgd} variant has a significantly more number of configurations near the optimal value, and the average performance and the overall distribution of our optimization is better for the \textit{rgd} variant than \textit{exp}.
    Similar trends can be observed for the EMPIRE10 lung challenge in Fig.~\ref{fig:lungtable}, where the \textit{exp} representation underperforms for the same cost function, data, etc.
    Therefore, we recommend direct RGD optimization for diffeomorphisms.
    }
    \label{fig:rgd-vs-exp}
\end{figure}

\paragraph{Sensitivity to Perturbations and limits of Expressivity}
Another potential source of numerical instability arises from the sensitivity of diffeomorphisms to perturbations in their underlying velocity fields.
While perturbation and sensitivity analyses are well established for matrix exponentials, often showing that output deviations grow exponentially with the norm of input perturbations \cite{van1977sensitivity,zhu2008sensitivity}. %
Several works have also investigated the singularities of the Euler equation ~\cite{singularityeuler,preston2004ideal,lee2018geometry} that leads to blowups in geodesic flows that represent diffeomorphisms.
Unlike finite-dimensional Lie groups, the derivative of the exponential can fail to be surjective, possibly producing ill-conditioning and numerical instability near certain vector fields representing conjugate directions \cite{ebin2006singularities}.
SVFs have a few limitations in regards to expressivity.
For example, there are diffeomorphisms arbitrarily close to the identity that are not contained in flows (1-parameter subgroups or SVFs) \cite{milnor1984}, showing that the exponential map is not surjective to the group of diffeomorphisms even locally.
The strong dependence on initial conditions and parameters observed in such systems suggests that analogous sensitivities may contribute to numerical instability and inexpressivity in SVF-based optimization methods.
This is empirically observed in \cref{fig:rgd-vs-exp}, where the \textit{exp} representation underperforms for the same cost function and dataset, potentially due to instability and inexpressivity being factors in the small performance degradation since the other parameters of the optimization (loss function, regularization) are kept constant or determined using cross-validation (optimal learning rate, for example).

Direct optimization of diffeomorphisms do not suffer from these limitations, since the perturbations of the diffeomorphism (outputs) are controlled directly by the magnitude of the velocity field in the update rule $\varphi_{t+1} = \varphi_t \circ (id + \epsilon_t v_t)$, and any diffeomorphism close to the identity can be obtained trivially by controlling $v_t$ and $\epsilon_t$ appropriately.

\begin{figure}[h!]
    \centering
    \includegraphics[width=\linewidth]{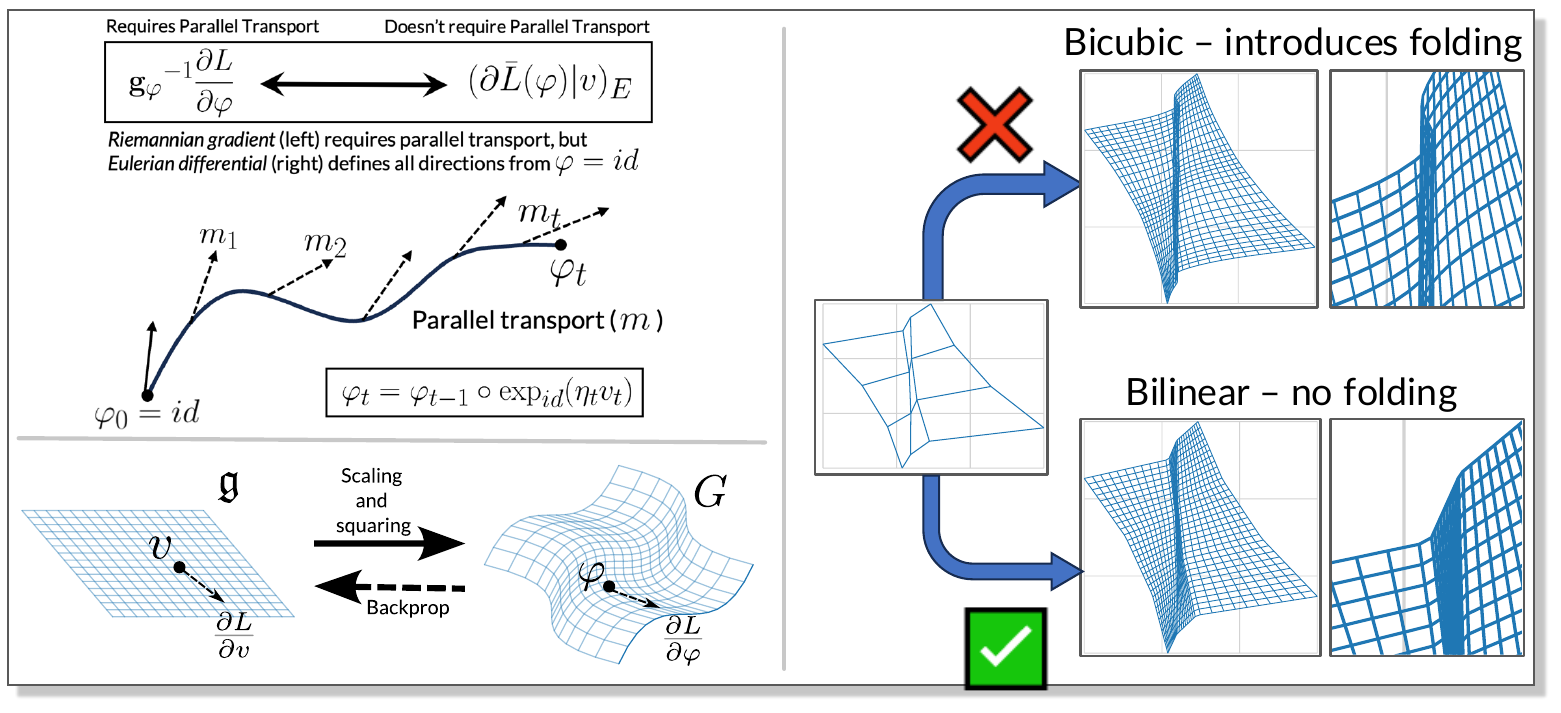}
    \caption{\textbf{Technical framework of FireANTs}: 
    \textbf{Left} shows the technical contributions of FireANTs. We extend Adaptive Optimization to multi-scale Diffeomorphisms by first writing the Riemannian gradient update, and then avoiding parallel transport of the optimization state by leverging the interchangability of the Riemannian gradient at arbitrary transform $\varphi_t$ with the Riemannian gradient at $\varphi = \textbf{Id}$.
    For the Lie-algebra representation, the Gateaux derivative $\frac{\partial L}{\partial \varphi}$ is projected to $\frac{\partial L}{\partial v}$ using analytical backprop. Since the Lie algebra is a vector space, we use standard adaptive optimizers (see~\cref{sec:rgd} for more details).
    \textbf{Right} takes a closer look at multi-scale interpolation for diffeomorphisms represented as a warp field. Bicubic interpolation can introduce folding of the warp field at a finer resolution due to overshooting, but bilinear interpolation does not. Therefore, we use this for interpolating the warp field and the optimizer state.
    }
    \label{fig:technical-framework}
\end{figure}

\begin{figure}[!htpb]

\end{figure}

\begin{table}[!ht]
  \centering
  \caption{\textbf{Quantitative performance on OASIS validation set.}FireANTs performs competitively with state-of-the-art registration methods on the OASIS dataset on both Dice Overlap and Haussdorf distance.}
  \label{fig:oasis}
  \resizebox{0.8\linewidth}{!}{%
  \begin{tabular}{lcc}
  \hline
  \multicolumn{3}{c}{\textbf{Validation metrics on OASIS}} \\ \hline
  \textbf{Method} & \textbf{Dice} & \textbf{HD95} \\ \hline
  Affine (Baseline)  & 0.572 $\pm$ 0.051 & 3.831 $\pm$ 0.718 \\
  \hline
  ANTs~\cite{ants} & 0.786 $\pm$ 0.033 & 2.209 $\pm$ 0.534 \\
  VoxelMorph~\cite{balakrishnan2019voxelmorph} & 0.753 $\pm$ 0.145 & - \\
  LogDemons~\cite{vercauteren2007diffeomorphic} & 0.804 $\pm$ 0.022 & 2.068 $\pm$ 0.448\\
  SynthMorph~\cite{hoffmann2021synthmorph} & 0.785 $\pm$ 0.023 & 2.311 $\pm$ 0.452 \\ \hline
  FireANTs & 0.791 $\pm$ 0.028 & 2.793 $\pm$ 0.602 \\ \hline
  \end{tabular}
  }
\end{table}

\begin{figure}[!htpb]
  \begin{minipage}{\linewidth}
    \centering
    \subcaption{Trick to avoid parallel transport in Riemannian Adaptive Optimization using Eulerian differentials}
    \includegraphics[width=0.75\linewidth]{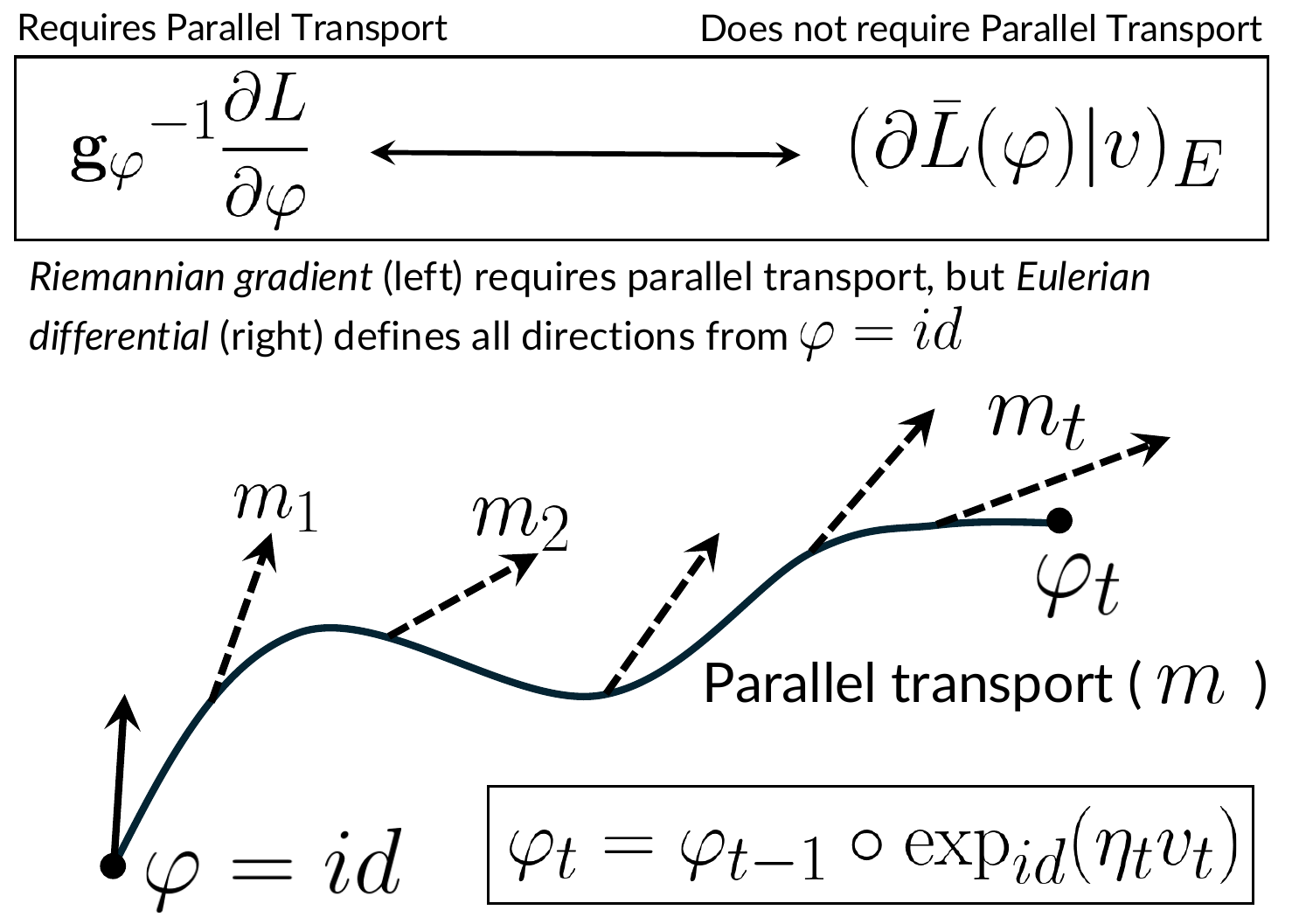}
  \end{minipage}
  \begin{minipage}{\linewidth}
  \subcaption{Bicubic interpolation of diffeomorphic map does not preserve diffeomorphism}
  \begin{minipage}{0.3\linewidth}
    \includegraphics[width=\linewidth]{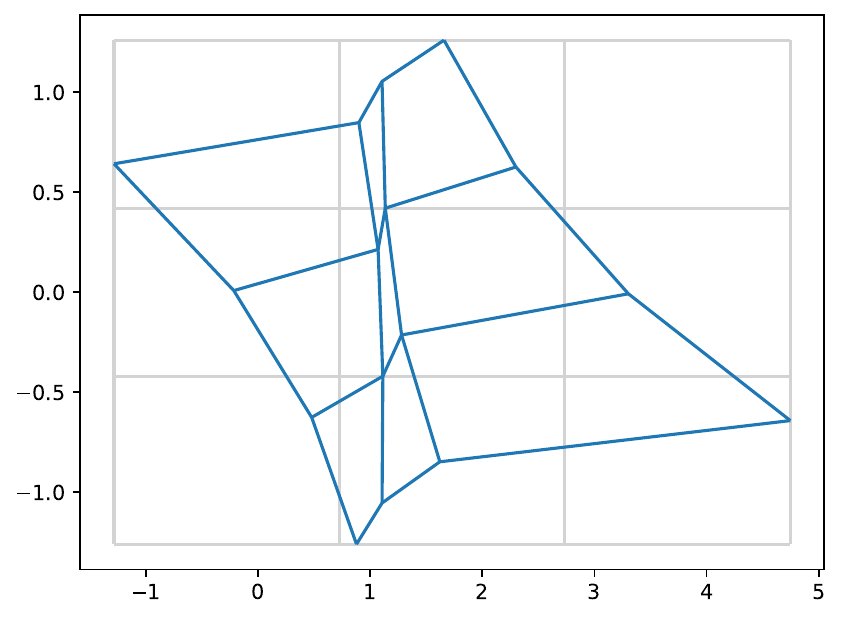}
  \end{minipage}
  \begin{minipage}{0.65\linewidth}
    \centering
    \includegraphics[width=0.3\linewidth]{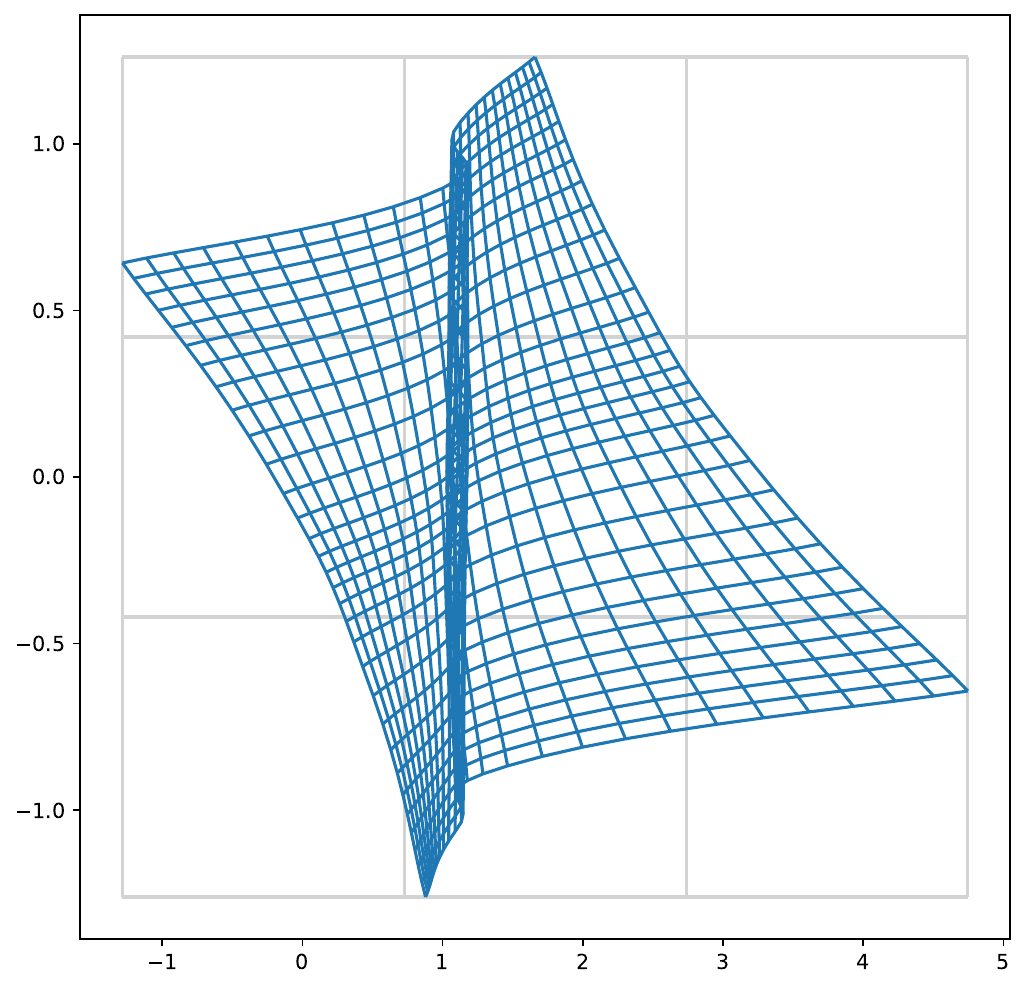}
    \adjincludegraphics[width=0.3\linewidth, trim={{.4\width} {0.78\height} {.5\width} {0.12\height}}, clip]{figures/supp/warp_cubic.pdf}
    \includegraphics[width=0.37\linewidth]{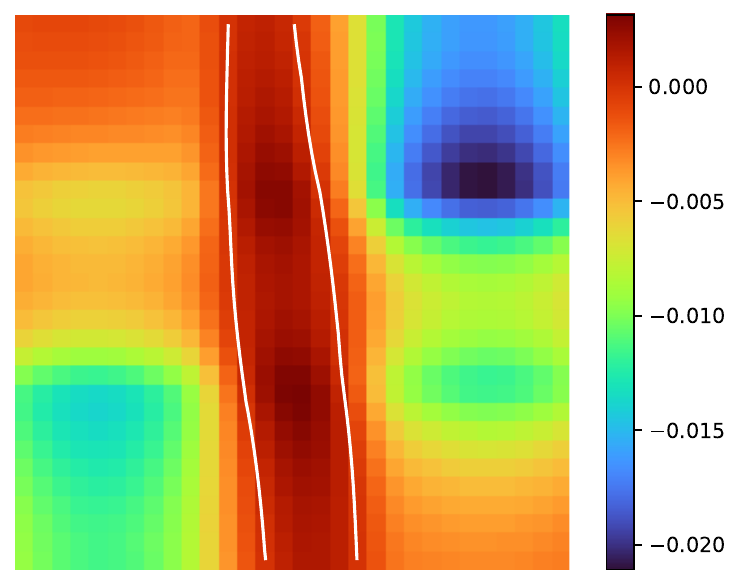}
    \includegraphics[width=0.3\linewidth]{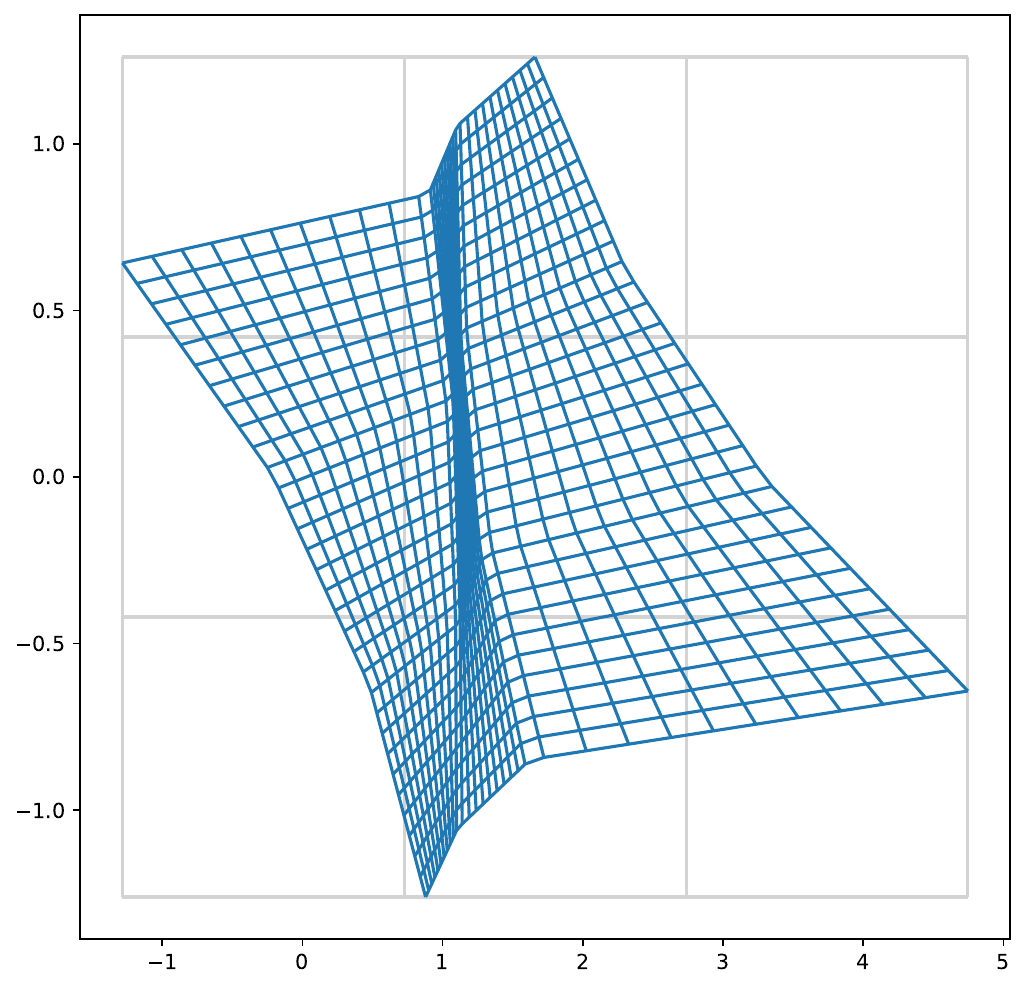}
    \adjincludegraphics[width=0.3\linewidth, trim={{.4\width} {0.78\height} {.5\width} {0.12\height}}, clip]{figures/supp/warp_bilinear.pdf}
    \includegraphics[width=0.37\linewidth]{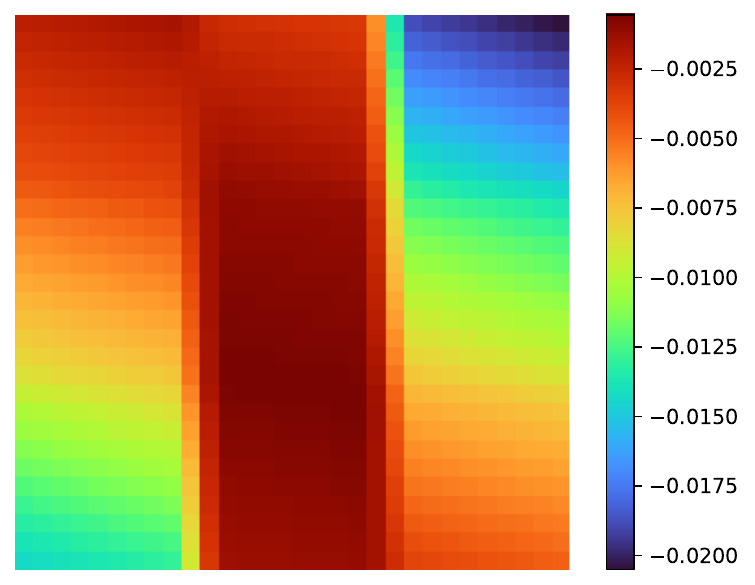}
  \end{minipage}
  \end{minipage}
  \caption{\textbf{Overview of tricks for multi-scale adaptive optimization for diffeomorphisms}: (a) We exploit the group structure of diffeomorphisms to define an Eulerian differential that avoids the need for parallel transport in adaptive optimization algorithms.
  (b) We show the effect of downsampling on the warp and determinant of the Jacobian for a single image pair. The first column shows the initial warp, and the second and third columns show the warp and determinant of the Jacobian for the cubic and bilinear interpolations, respectively.}
  \label{fig:rie-opt-tricks}
\end{figure}

\begin{figure}[ht!]
    \centering
    \includegraphics[width=0.9\linewidth]{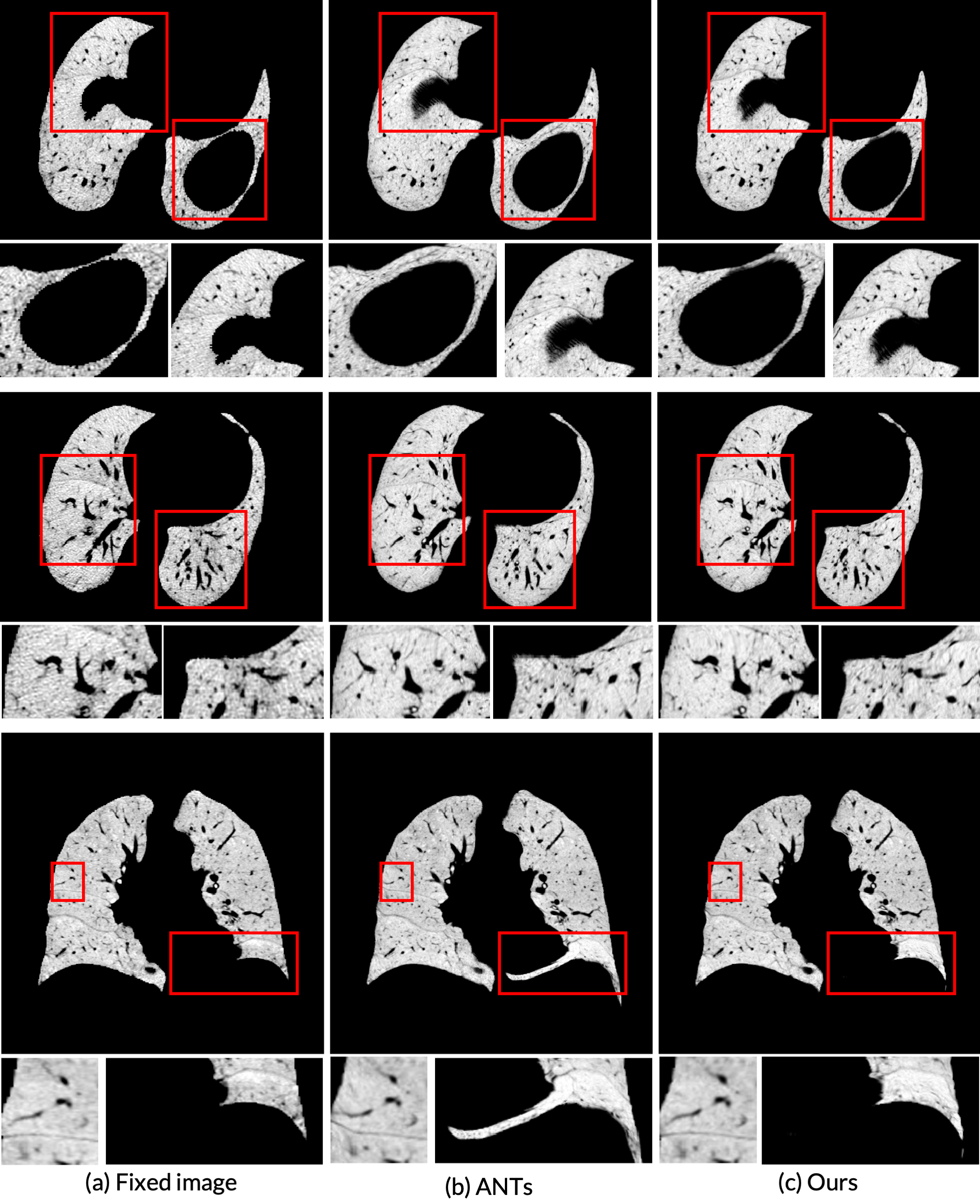}
    \caption{\textbf{Qualitative results on EMPIRE10 challenge}: (a) shows the fixed image, (b) shows the registration performed by ANTs, and (c) our method, all with zoomed in regions.
    ANTs performs a coarse registration with ease, but still leaves out critical alignment of lung boundary and airways by not utilizing adaptive optimization.
    Our method performs \textit{perfectly} diffeomorphic registration by construction, and does not lead to any registration errors, both in the lung boundaries or internal features.
    }
    \label{fig:lung-images-1}
\end{figure}

\begin{figure}[ht!]
    \centering
    \includegraphics[width=0.9\linewidth]{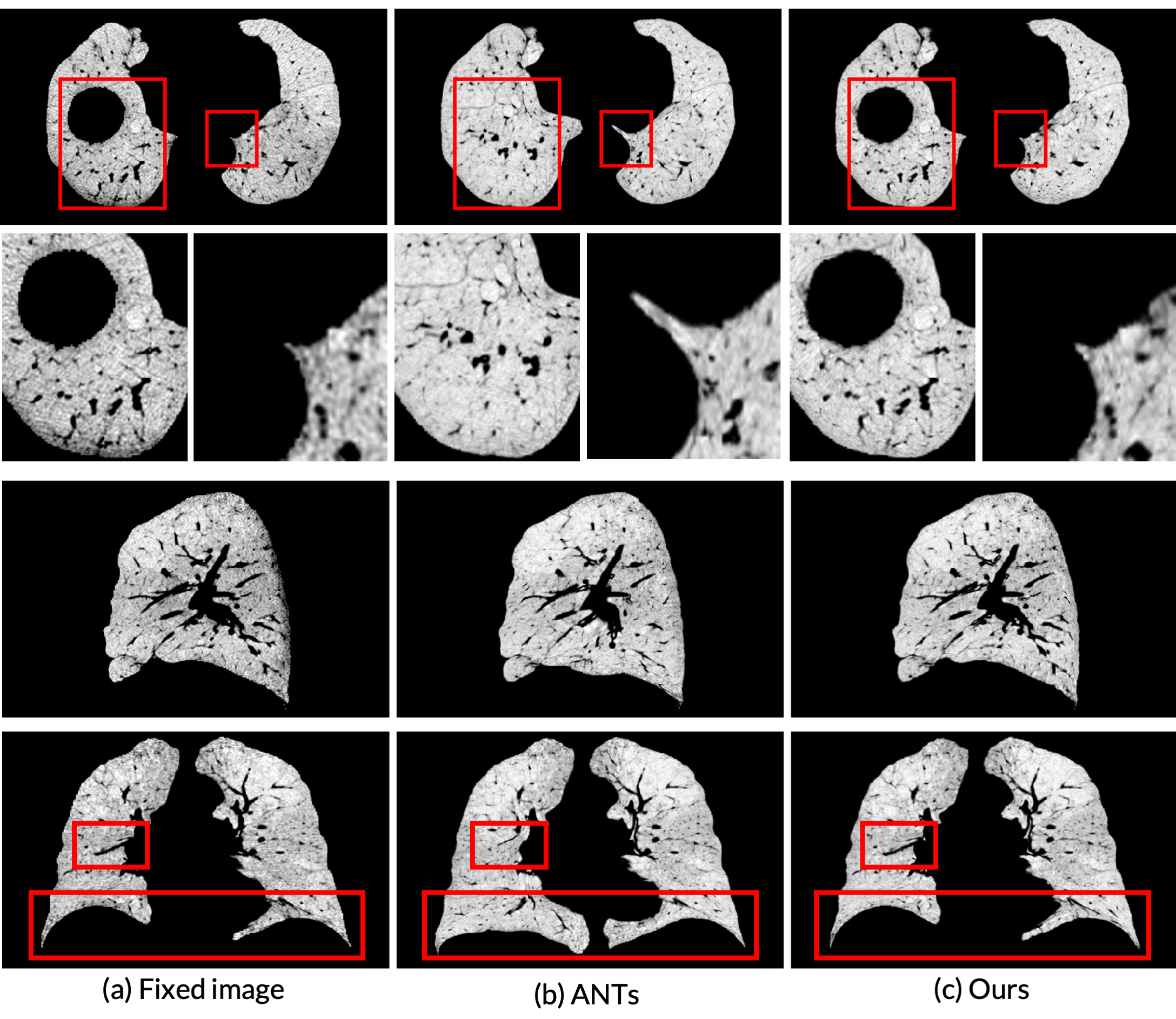}
    \caption{\textbf{More Qualitative results on EMPIRE10 challenge}: \textbf{(a)} shows the fixed image, \textbf{(b)} shows the registration performed by ANTs, and \textbf{(c)} our method, all with zoomed in regions.
    ANTs performs a coarse registration with ease, but still leaves out critical alignment of lung boundary and airways by not utilizing adaptive optimization.
    Our method performs \textit{perfectly} diffeomorphic registration by construction, and does not lead to any registration errors, both in the lung boundaries or internal features.
    }
    \label{fig:lung-images-2}
\end{figure}

\end{document}